%% file: main.tex
\documentclass{article}

\usepackage[accepted]{icml2026}

\usepackage[utf8]{inputenc} 
\usepackage[T1]{fontenc}    
\usepackage{hyperref}
\hypersetup{breaklinks}

\usepackage{url}            
\usepackage{booktabs}       
\usepackage{amsfonts}       
\usepackage{nicefrac}       
\usepackage{microtype}      
\usepackage{xcolor}         
\usepackage{microtype}
\usepackage{graphicx}

\usepackage{amsmath}
\usepackage{amssymb}
\usepackage{mathtools}
\usepackage{amsthm}

\usepackage{multicol}
\usepackage{xspace}
\usepackage{subfigure}
\usepackage{multirow}
\usepackage{wrapfig}
\usepackage{float}

\theoremstyle{plain}
\newtheorem{theorem}{Theorem}[section]

\newtheorem{lemma}[theorem]{Lemma}
\newtheorem{corollary}[theorem]{Corollary}
\theoremstyle{definition}

\newtheorem{assumption}[theorem]{Assumption}
\theoremstyle{remark}
\newtheorem{remark}[theorem]{Remark}

\newcommand{\clove}{\textit{CLoVE}\xspace}
\newcommand{\IFCA}{\textit{IFCA}\xspace}
\newcommand{\CFL}{\textit{CFL-S}\xspace}
\newcommand{\local}{\textit{Local-only}\xspace}
\newcommand{\fedavg}{\textit{FedAvg}\xspace}
\newcommand{\kfed}{\textit{k-FED}\xspace}

\icmltitlerunning{CLoVE: Personalized Federated Learning through Clustering of Loss Vector Embeddings}

\begin{document}

\twocolumn[
  \icmltitle{CLoVE: Personalized Federated Learning\\ through Clustering of Loss Vector Embeddings}



  \icmlsetsymbol{equal}{*}

  \begin{icmlauthorlist}
    \icmlauthor{Randeep Bhatia}{equal,NBL}
    \icmlauthor{Nikos Papadis}{equal,NBL}
    \icmlauthor{Murali Kodialam}{NBL}
    \icmlauthor{TV Lakshman}{NBL}
    \icmlauthor{Sayak Chakrabarty}{NW}
  \end{icmlauthorlist}

  \icmlaffiliation{NBL}{Nokia Bell Labs, Murray Hill, NJ, USA}
  \icmlaffiliation{NW}{Northwestern University, IL, USA}

  \icmlcorrespondingauthor{Randeep Bhatia}{randeep.bhatia@nokia-bell-labs.com}
  \icmlcorrespondingauthor{Nikos Papadis}{nikos.papadis@nokia-bell-labs.com}

  \icmlkeywords{federated learning, personalized federated learning, clustering, embeddings}

  \vskip 0.3in
]


\printAffiliationsAndNotice{\icmlEqualContribution}

\begin{abstract}
\input{sections/0-abstract}
\end{abstract}

\input{sections/1-intro}

\input{sections/2-related-work}

\input{sections/3-our-algorithms}

\input{sections/4-theory}

\input{sections/5-perf-eval}

\input{sections/6-conclusion}



\section*{Impact Statement}
This paper presents work whose goal is to advance the field of machine learning. There are many potential societal consequences of our work, none of which we feel must be specifically highlighted here.

\bibliography{references}
\bibliographystyle{icml2026}


\newpage
\appendix
\onecolumn

\input{sections/A1-proofs}

\input{sections/A2-discussion}

\input{sections/A3-hyperparameters-and-datasets}

\input{sections/A4-more-experiments}


\end{document}

%% file: sections/0-abstract.tex
We propose \clove (Clustering of Loss Vector Embeddings), a novel algorithm for Clustered Federated Learning (CFL). 
In CFL, clients are naturally grouped into clusters based on their data distribution. 
However, identifying these clusters is challenging, as client assignments are unknown. 
\clove utilizes client embeddings derived from model losses on client data, and leverages the insight that clients in the same cluster share similar loss values, while those in different clusters exhibit distinct loss patterns. 
Based on these embeddings, \clove is able to iteratively identify and separate clients from different clusters and optimize cluster-specific models through federated aggregation. 
Key advantages of \clove over existing CFL algorithms are (1) its simplicity, (2) its applicability to both supervised and unsupervised settings, and (3) the fact that it eliminates the need for near-optimal model initialization, which makes it more robust and better suited for real-world applications. 
We establish theoretical convergence bounds, showing that 
\clove can recover clusters accurately with high probability in a single round and converges exponentially fast to optimal models in a linear setting. 
Our comprehensive experiments comparing with a variety of both CFL and generic Personalized Federated Learning (PFL) algorithms on different types of datasets and an extensive array of non-IID settings demonstrate that \clove achieves highly accurate cluster recovery in just a few rounds of training, along with state-of-the-art model accuracy, across a variety of both supervised and unsupervised PFL tasks.

%% file: sections/1-intro.tex
\section{Introduction}
\label{sec:introduction}

Federated learning (FL) has emerged as a pivotal framework for training models on decentralized data, preserving privacy and reducing communication overhead \citep{mcmahan2017communication, konecny2016federated, bonawitz2019towards}. 
However, a significant challenge in FL is posed by the heterogeneity of data distributions across clients (non-IID data) \citep{zhao2018federated}, as this can adversely impact the accuracy and convergence of FL models. 
To address this issue, Personalized Federated Learning (PFL) \citep{tan2023towards, hanzely2020federated} has developed as an active research area, focusing on tailoring trained models to each client's local data, thereby improving model performance.

Our research focuses on a variant of PFL known as {\it Clustered Federated Learning (CFL)}. 
In CFL, \citep{IFCA_paper_2020, Werner_2023, Sattler_2021_CFL, Mansour_2020, Chung_2022, long_multicenter_2023_FeSEM, duan_flexible_2022_FlexCFL, vahidian_efficient_2023_PACFL}, clients are naturally grouped into $K$ \textit{true clusters} based on inherent similarities in their local data distributions. 
However, these cluster assignments are not known in advance. 
The objective of CFL is to develop distinct models tailored to each cluster, utilizing only the local data from clients within that cluster. 
To achieve this goal, identifying the cluster assignments for each client is crucial, as it enables the construction of accurate, cluster-specific models that capture the unique characteristics of each group.

Many existing approaches \citep{awasthi2012improved, kumar2010clustering} that provide guarantees on cluster assignment recovery often rely on the assumption that data of different clusters are well-separated. 
However, this assumption is often violated in real-world datasets, as evidenced by the modest performance of k-means on MNIST clustering even in a centralized setting. 
Its low Adjusted Rand Index (ARI)\footnote{
ARI is a measure of the level of agreement between two clusterings, calculated as the fraction of pairs of cluster assignments that agree between the clusterings, adjusted for chance.
} of only approximately $50\%$ \citep{MNIST_kmeans_blogpost} can be attributed to the high variance in the MNIST image data, resulting in many digits being closer to other clusters than to their own digit's cluster. 
Furthermore, in the federated setting, clustering approaches like \textit{k-FED} \citep{Dennis_2021} can recover clustering in one shot, but they also rely on the inter-cluster separation conditions of \cite{awasthi2012improved}.
However, as we demonstrate (Appendix~\ref{app:mixed_linear_regression_experiments}), even \textit{k-FED} achieves low ARI on clustering common mixed linear regression distributions \citep{yi2014alternating}, 
as the cluster centers of those distributions can also be very close, highlighting the limitations of these approaches.

\begin{figure}[ht]
    \centering
    \includegraphics[width=\columnwidth]{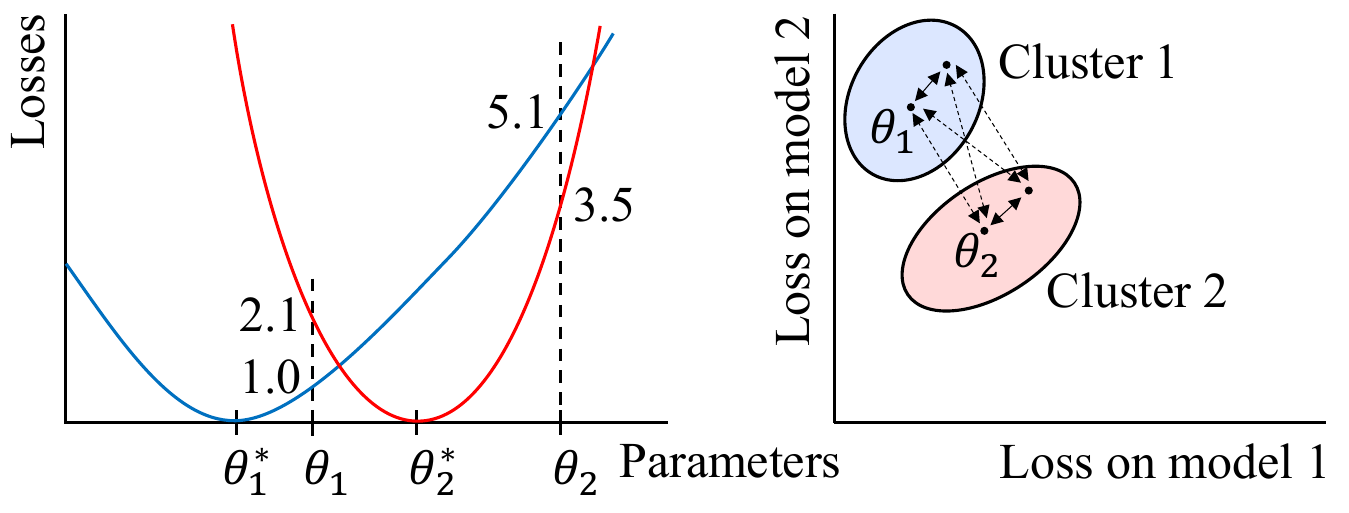}
    \caption{
    \textit{Left: Model parameters are on the x-axis and model losses on the y-axis.     
    Blue and red curves depict approximate losses for clients of cluster $1$ and $2$ respectively.
    $\theta^*_1$ and $\theta^*_2$ are optimal models for clients of clusters $1$ and $2$  respectively.     
    Initial models are $\theta_1$ and $\theta_2$. Loss on model $\theta_1$ ($\theta_2$) is $1.0$ ($5.1$) and $2.1$ ($3.5$) for clients of clusters $1$ and $2$, respectively.     
    Loss vectors are centered around $[1.0, 5.1]$ for clients of cluster $1$ and around $[2.1, 3.5]$ for clients of cluster $2$.     
    In \IFCA, all clients select model $\theta_1$, as it minimizes their loss.     
    Consequently they all get assigned to cluster $1$.     
    Moreover, because of this assignment, model $\theta_2$ receives no updates, and therefore its loss stays high; as a result, no clients ever select it in future rounds.    
    Ultimately, this leads to all clients converging to the same cluster, failing to recover the true client clusters.     
    In contrast, clustering based on loss vectors (right) achieves accurate client partitioning, as the loss vectors of clients from different clusters are highly distinct.}
    }
    \label{fig:IFCA_vs_ours_intro_fig}
\end{figure}

An alternative way to recover cluster assignments is to leverage the separation of cluster-specific models. For instance, the Iterative Federated Clustering Algorithm (\IFCA) \citep{IFCA_paper_2020} assigns clients to clusters based on the model that yields the lowest empirical loss.
However, \IFCA has a significant limitation: it requires careful initialization of cluster models, ensuring each model is sufficiently close to its optimal counterpart. 
Specifically, the distance between an initial model and the cluster's optimal model must be strictly less than half the minimum separation distance between any two optimal models. 
In practice, achieving this distance requirement can be challenging. 
For example, in our experiments with the MNIST dataset, \IFCA often failed to recover accurate cluster assignments. 
As shown in Fig.~\ref{fig:IFCA_vs_ours_intro_fig}, this issue typically occurs when clients from two or more clusters achieve the lowest loss on the same model, and thus get assigned to the same cluster. 
This, in turn, causes the algorithm to get stuck and fail to recover accurate clusters.

In this paper, we propose the Clustering of Loss Vector Embeddings (\clove) PFL algorithm, a novel approach to tackle both the challenges of constructing accurate cluster-specific models and recovering the underlying cluster assignments in a FL setting. 
Our solution employs an iterative process that simultaneously builds multiple personalized models and generates client embeddings, represented as vectors of losses achieved by these models on the clients' local data. 
Specifically, we leverage the clustering of client embeddings to refine the models, while also using the models to inform the clustering process. 
The underlying principle of our approach is that clients within the same cluster will exhibit similar loss patterns, whereas those from different clusters will display distinct loss patterns across models. 
By analyzing and comparing these loss patterns through the clustering of loss vectors, our approach effectively separates clients from different clusters while grouping those from the same cluster together. 
This, in turn, leads to improved models for each cluster, as each model predominantly receives updates from clients within its corresponding cluster, thereby enhancing accuracy and robustness. 
Note that even if there is low separability in the clients' data points, the loss vectors can still be well-separated, as demonstrated in our experiments with various types of data mixings with significant raw-data overlap across clusters.

\clove exhibits a distinct advantage over existing CFL methods, 
including \IFCA, \CFL~\citep{Sattler_2021_CFL}, \kfed \cite{Dennis_2021}, \textit{FlexCFL}~\citep{duan_flexible_2022_FlexCFL}, and \textit{FeSEM}~\citep{long_multicenter_2023_FeSEM}. 
Notably, \clove achieves robust clustering results even from poor initializations (where ``poor''  refers to a random/non-specific initialization of the model weights), a significant advantage over \IFCA, which requires careful model initialization within some limits.
Compared to \CFL, which often necessitates hundreds of communication rounds to reach convergence, 
\clove achieves clustering in fewer than ten rounds while reliably handling sparse client participation and stragglers. 
This accelerated convergence results in substantially lower computation and communication overhead compared to both \IFCA and \CFL.
Unlike \textit{FlexCFL}, which produces a static clustering in a single round, 
\clove dynamically refines its cluster assignments, offering the dual benefits of low-cost operation and the capacity to 
improve clustering quality over time. 
Moreover, \clove's loss vector-based cluster recovery approach avoids the local optima issues inherent 
in expectation maximization approaches such as \textit{FeSEM}, thereby ensuring optimal cluster recovery with high probability.

We conduct a comprehensive evaluation of our approach, demonstrating its ability to recover clusters with high accuracy while constructing per-cluster models in just a few rounds of FL.
Furthermore, our analytical results show that our approach can achieve single-shot, accurate cluster recovery with high probability in the case of linear regression, thereby establishing its efficacy and robustness. 

Our technical \textbf{contributions} can be summarized as follows: 
\begin{itemize}
    \setlength{\itemsep}{0pt} 
    \item \textbf{Introduction of the \clove PFL algorithm}: We propose a novel approach that simultaneously builds multiple personalized models and generates client embeddings based on model losses, enabling effective clustering of clients with unknown cluster labels.

    \item \textbf{Simplicity and wide applicability}: 
    Our method, unlike most others, extends to both supervised and unsupervised settings, handling a variety of data distributions,  sparse client participation, and stragglers through a simple, loss-based embedding approach.
        
    \item \textbf{Robustness against initialization}: Unlike existing methods, \clove does not require careful initialization of cluster models, thereby mitigating the risk of inaccurate cluster assignments due to initialization sensitivities.

    \item \textbf{Theoretical guarantees for cluster recovery}: We provide analytical results showing that, in the context of linear regression, \clove can achieve single-shot, accurate cluster recovery with high probability.

    \item \textbf{Efficient cluster recovery and model construction}: We demonstrate through comprehensive evaluations that \clove can recover the true clusters with high accuracy and construct per-cluster models within a few rounds of federated learning in a variety of settings.

\end{itemize}

%% file: sections/2-related-work.tex
\section{Related Work}

A primary challenge in FL is handling heterogeneous data distributions (non-IID data) and varying device capabilities. 
Early optimization methods like \fedavg \citep{mcmahan2017communication} and \textit{FedProx} \citep{li2019convergence} mitigate some issues by allowing multiple local computations before aggregation, yet they typically yield a single global model that may not capture the diversity across user populations \citep{zhao2018federated}.

Personalized Federated Learning (PFL) \citep{hanzely2020federated, tan2023towards} aims to deliver client-specific models that outperform both global models and na\"ive local baselines. 
Various strategies have been proposed, including fine-tuning global models locally \citep{fallah_personalized_2020_Per-FedAvg} or performing adaptive local aggregation \citep{zhang_FedALA_2023}, meta-learning \citep{jiang2023improving}, multi-task learning \citep{smith2017federated, xu_personalized_2023_FedPAC}, and decomposing models into global and personalized components \citep{deng2020adaptive, Mansour_2020}. 
However, these methods operate in a supervised setting and require labeled data.

Clustered Federated Learning (CFL) is an emerging approach in PFL that involves clustering clients based on similarity measures to train specialized models for each cluster (e.g. \textit{IFCA} based on model loss values, \citep{Werner_2023, kim2024} and \textit{FlexCFL} based on gradients). 
The \CFL algorithm utilizes federated multitask learning and gradient embeddings in order to iteratively form and refine clusters.
\cite{vahidian_efficient_2023_PACFL}'s \textit{PACFL} algorithm derives a set of principal vectors from each cluster's data, and \textit{FedProto} \citep{tan_fedproto_2022} constructs class prototypes, which allow the server to identify distribution similarities.

Traditional clustering methods, such as spectral clustering and k-means, have also been adapted for federated scenarios. 
In \textit{FeSEM}, personalization is achieved by deriving the optimal matching of users and multiple model centers, and in \cite{chen_spectral_2023} through spectral co-distillation. 
\cite{FedGWC} is a hierarchical approach like \cite{Sattler_2021_CFL}, and as such the number of rounds for clustering can be very large.
Despite their effectiveness, these methods often require complex similarity computations, additional communication steps, or additional amount of state kept at the server.

Recent advancements in CFL have focused on enhancing convergence and reducing reliance on initial cluster assignments. 
Improved algorithms \citep{improved_clustered_fed2024} address some limitations of earlier approaches that require careful initialization like \IFCA, but their methods are complex.
Our \clove algorithm overcomes these drawbacks using a simple method that utilizes robust loss-based embeddings, converges quickly to the correct clusters, eliminates the need for precise initialization, and operates in a variety of both supervised and unsupervised settings.

%% file: sections/3-our-algorithms.tex
\section{Embeddings-driven Federated Clustering}
\label{sec:our_algorithms}

The typical FL architecture consists of a server and a number of clients. 
Let $M$ be the number of clients, which belong to $K$ clusters. 
Data for clients of cluster $j \in [K]$  follow a distinct data distribution $D_j$  (where the notation $[K]$ denotes the set $\{1,2,...,K\}$). 
Each client $i \in [M]$ has $n_i$ data points. 
Let $f(\theta; z): \Theta \rightarrow \mathbb{R}$ be the loss function associated with data point $z$, where $\Theta \subseteq \mathbb{R}^d$ is the parameter space. 
The goal in CFL is to simultaneously cluster the clients into $K$ clusters and train $K$ models to minimize the population loss function $\mathcal{L}_j(\theta) \triangleq \mathbb{E}_{z \sim D_j} [f(\theta;z)]$ for each cluster $j \in [K]$.
The empirical loss for the data $Z_i$ of client $i$ is defined as $\mathcal{L}^i (\theta; Z_i) \triangleq (1/|Z_i|)  \sum_{z \in Z_i} f(\theta; z)$.

\begin{algorithm}[ht!]
\fontsize{9pt}{9pt}\selectfont
    \caption[short]{Clustering of Loss Vector Embeddings (\clove) algorithm for PFL}
    \label{alg:CLoVE}

\begin{algorithmic}[1]  
    \renewcommand{\algorithmicrequire}{\textbf{Input:}}    
    \REQUIRE \# of clients $M$, upper bound on \# of models $\hat{K}$, \# of data points $n_i$ for client $i$, $i \in [M]$, participation rate $\rho$
    \renewcommand{\algorithmicrequire}{\textbf{Hyperparameters:}}
    \REQUIRE learning rate $\gamma$, \# of rounds $T$, \# of local epochs $\tau$ (for model averaging)
    \renewcommand{\algorithmicensure}{\textbf{Output:}}
    \ENSURE number of clusters $K^{(T)}$, final model parameters $\left\{ \theta_j^{(T)}, j \in [K^{(T)}] \right\}$

    \STATE Server: initialize model parameters $\theta_j^{(0)}$ for each of the $K^{(0)}$ models $j \in [K^{(0)}]$, where $K^{(0)} = \hat{K}$ \\
    
    \FOR{round $t=0, 1, ..., T-1$}

        \STATE $M^{(t)} \leftarrow$ random subset of $\rho$ fraction of the clients (deals with partial participation/stragglers) \\
        
        \IF{clustering has not stabilized}
            
            \STATE Server: broadcast model parameters $\theta_j^{(t)}$ of all $\hat{K}$ models to all clients in $M^{(t)}$\\
            
            \FOR{each client $i \in M^{(t)}$ in parallel}
                \STATE Run client $i$'s data through all $\hat{K}$ models and get the losses $\mathcal{L}^{i,{(t)}} \triangleq \left( \mathcal{L}^i(\theta_j^{(t)}) \right), j \in [\hat{K}]$
    
                \STATE Send loss vector $\mathcal{L}^{i,{(t)}}$ to the server
            \ENDFOR
                
            \STATE Server (clustering step):\\    
            \STATE $K^{(t+1)} = \text{\fontfamily{qcr}\selectfont selectBestK}(\mathcal{L}^{(t)}, \hat{K})$, where $\mathcal{L}^{(t)} \triangleq \{ \mathcal{L}^{i, (t)}, i \in [M^{(t)}] \}$ is the loss matrix \\
             
            \STATE Run clustering algorithm to group the set of vectors $\{ \mathcal{L}^{i,{(t)}}, i \in [M^{(t)}] \}$ to form $K^{(t+1)}$ client clusters \\
            
            \STATE Map client clusters to models using min-cost matching and assign each client $i \in M^{(t)}$ to the corresponding model $\kappa_i \equiv \kappa_i^{(t)} \in [K^{(t+1)}]$ of its client cluster

        \ELSE
            \STATE Server: \\
            \STATE For new clients: assign them a model using  models and centroids saved at stability \\ 
            
            \STATE Send updated model parameters of its assigned model $\kappa_i$ to each client $i \in M^{(t)}$\\
        \ENDIF
        
        \FOR{each client $i \in M^{(t)}$ in parallel}
            \STATE \textbf{-- Option I} (model averaging): 
            Compute new model parameters $\widetilde{\theta}_i = \text{\fontfamily{qcr}\selectfont LocalUpdate}(i, \theta_{\kappa_i}^{(t)}, \gamma, \tau) $ for model $\kappa_i$
            and send them to the server

            \STATE \textbf{-- Option II} (gradient averaging): 
            Compute stochastic gradient $g_i = \widehat{\nabla} \mathcal{L}^i (\theta_{\kappa_i}^{(t)})$ using model $\kappa_i$
            and send it to the server
        \ENDFOR

        \STATE Server (averaging step):\\
        \STATE Clients that were assigned to model $j$ at round $t$: $C_j^{(t)} \triangleq \{ i \in M^{(t)} \text{ s.t. } \kappa_i^{(t)} = j\}$ \\
        
        \STATE \textbf{-- Option I} (model averaging): 
            $\forall~j \in [K^{(t+1)}] \text{ s.t. } C_j^{(t)} \neq \emptyset$, $\theta_{j}^{(t+1)} = \dfrac{\sum_{i \in C_j^{(t)}} n_i \widetilde{\theta}_{i}} {\sum_{i \in C_j^{(t)}} n_i}$
            
        \STATE \textbf{-- Option II} (gradient averaging): 
            $\forall~j \in [K^{(t+1)}] \text{ s.t. } C_j^{(t)} \neq \emptyset$, $\theta_{j}^{(t+1)} = \theta_{j}^{(t)} - \gamma \dfrac{\sum_{i \in C_j^{(t)}} n_i g_i} {\sum_{i \in C_j^{(t)}} n_i} $\\
    \ENDFOR

    \STATE return $K^{(T)}$, $\left\{ \theta_j^{(T)}, j \in [K^{(T)}] \right\}$ \\
\end{algorithmic}
\end{algorithm}

The pseudocode of our \textit{Clustering of Loss Vector Embeddings} (\clove) PFL algorithm is shown in Alg. \ref{alg:CLoVE}.
The algorithm operates as follows.
Initially, the server creates $\hat{K}$ models, where $\hat{K}$ is the upper bound on the number of models, given as an input (line 1). 
(If the true $K$ is known upfront, then $\hat{K} = K$).
Until cluster stability is reached, it sends all $\hat{K}$ models to a randomly selected fraction $\rho$ of the participating clients in round $t$ (lines 3, 5). 
Each of these clients runs its data through all models and gets the losses, which form one loss vector of length $\hat{K}$ per client. 
These loss vectors are then sent to the server (lines 6-8). The server, if needed, determines the number of clusters $K^{(t+1)}$ for the next round by calling the function $\text{\fontfamily{qcr}\selectfont selectBestK}(\cdot)$ (e.g. via searching over a range of values and choosing the one that yields the highest silhouette score, as described in Alg. \ref{alg:selection_of_K}, or by using the elbow method). 
It then groups clients into $K^{(t+1)}$  clusters (line 12) by clustering  their loss vectors (e.g. using k-means). 
It then assigns clients to those models to which their clusters get matched via a minimum-cost matching (line 13). 
For this, a bipartite graph is created, with the nodes being the client clusters on the left and the models on the right, and the weight of each edge being the sum of the losses of the clients in the cluster on the left on the model on the right. Min-cost bipartite matching is found using well known methods (e.g. max-flow-based \citep{munkres1957algorithms, lovasz2009matching, jonker1987shortest}). 
Subsequently, each client trains its assigned model $\kappa_i$ locally on its private data (lines 19-22) by running gradient descent denoted by the function $\text{\fontfamily{qcr}\selectfont LocalUpdate}(\cdot)$ for $\tau$ epochs, starting with model $\kappa_i$'s parameters, and sends its updates (model parameters in Option I and gradients in Option II) to the server. 
Then, the averaging step follows: the server aggregates and applies all the received updates for each model $j$ from the set $C_j^{(t)}$ of clients that updated it (lines 23-26).
In the next iteration, the server sends the computed updated parameters once again to the clients (line 5), and the process continues, until the clustering (i.e. assignment of clients to clusters/models) stabilizes.

Checking for cluster stability can be done as follows, in a way that the server does not need to keep any client-specific state.
Each client can keep the history of its assignment to models over the different rounds.
Once a client determines that its historical assignments have been stable for a set number of rounds required for stability, the client declares to the server that it has reached stability.
The server at each round checks whether all (or a desired percentage) of the participating clients have reached stability, and if so, it decides that universal stability has been achieved.

After the algorithm reaches a stable state, the server no longer needs to compute loss vectors or perform clustering. 
Instead, it can simply send the already assigned (single) model  to each client (line 17).
For any client that does not yet have an assigned model, the server assigns it the model whose cluster centroid is closest to the client's loss vector using the models and centroids saved at the point of stability (line 16). 
At this stage, the algorithm behaves like standard \fedavg, so both the communication and the computational overhead are minimal.

If the number of clusters is not known a priori, the initial upper bound $\hat{K}$ can for example be set to a multiple (e.g. $2 \times$) of the number of clusters obtained using a one-shot clustering method as in \textit{FlexCFL} or \textit{PACFL}, which can be run before training begins.
Then the choice of $K^{(t+1)}$ at each round happens as follows, as per Alg. \ref{alg:selection_of_K} (found in Appendix~\ref{app:alg-for-selecting-K}).
A range of candidate $K$ values is examined, up to the predetermined upper bound $\hat{K}$.
For each candidate $K$, the loss vectors are clustered and the silhouette score of the resulting clustering is computed.
Eventually, the value of $K$ with the highest silhouette score is returned.
Note that this procedure is not necessary when the number of clusters $K$ is known a priori: then, $\hat{K} = K$ and $K^{(t)} = K$ for all rounds $t$, and thus line 11 of Alg.~\ref{alg:CLoVE} can be skipped.

%% file: sections/4-theory.tex
\section{Theoretical analysis of \clove's performance}
\label{sec:theory}

We now provide theoretical evidence supporting the superior performance of \clove. We focus on the case where the number of clusters $K$ is fixed and known a priori. In the following we omit all proofs, which can be found in Appendix \ref{app:proofs}. To analyze the efficacy of \clove, we consider a mixed linear regression problem~\citep{yi2014alternating} in a federated learning setting~\citep{IFCA_paper_journal_version_2022}. 
In this context, each client's data originates from one of \(K\) distinct clusters, denoted as \(\mathcal{C}_1, \mathcal C_2, \dots, \mathcal C_K\). Within each cluster $\mathcal{C}_k$, the data pairs $z_i = (x_i, y_i)$ are drawn from the same underlying distribution $\mathcal{D}_k$, which is generated by the linear model: $y_i = \langle \theta^*_k, x_i \rangle + \epsilon_i$.
Here, the feature $x_i$ follows a standard normal distribution $\mathcal{N}(0, I_d)$, and the additive noise $\epsilon_i$ follows a normal distribution $\mathcal{N}(0, \sigma^2)$, both independently drawn. The loss function is the squared error: $f(\theta; x, y) = \left( \langle \theta, x \rangle - y \right)^2.$
Under this setting, the parameters $\theta^*_k \in \mathbb{R}^d$ are the minimizers of the population loss for cluster $\mathcal{C}_k$.

At a high level, our analysis unfolds as follows. We demonstrate that by setting the entries of a client's loss vector (SLV) to the square root of model losses, \clove can accurately recover client clustering with high probability in each round. This, in turn, enables \clove to accurately match clients to models with high probability by utilizing the minimum-cost bipartite matching between clusters and models. Consequently, we show that the distance between the models constructed by \clove and their optimal counterparts decreases exponentially with the number of rounds, following the application of client gradient updates, thus implying fast convergence of \clove to optimal models.

\input{tables/S_MCFAA_acc}

\textbf{Assumptions:}
For our analysis we make some assumptions: 1) All $\theta^*_i \in \mathbb{R}^d$ have unit norms and their minimum separation $\Delta> 0$ is close to $1$. 2) The number of clusters $K$ is small in comparison to the total number of clients $M$ and the dimension $d$. 3) Each client of cluster $k$ uses the same number $n_k$ of i.i.d. data points independently chosen at each round. 4) $\Delta / \sigma \gg 1$, i.e. there is either low noise, or high noise as long as the separation between cluster centers is high. In other words, the SNR (Signal to Noise Ratio) is not too small.

Note that, like \IFCA, we assume that the optimal models are separated by a value $\Delta$. 
However, unlike \IFCA, we do not impose any assumptions on the separation of the initial models.

For a given model \(\theta\), the \textbf{empirical loss} $\mathcal{L}_k^i(\theta)$ for a client \(i\) in cluster \(k\) is:
\[
\frac{1}{n_k} \sum_{j=1}^{n_k} \left( \langle \theta, x_j \rangle - y_j \right)^2 = \frac{1}{n_k} \sum_{j=1}^{n_k} \left( \langle \theta^*_k -\theta, x_j \rangle + \epsilon_j \right)^2
\]
For a given collection of models, $ \boldsymbol{\theta} =  [\theta_1 , \theta_2 , \cdots ]$, the square root loss vector (SLV) of a client $i$ of cluster $k$ is $ a_k^i(\boldsymbol{\theta}) = [F_k^i(\theta_1), F_k^i(\theta_2), \ldots ]$. 
Here $F_k^i(\theta) = \sqrt {\mathcal{L}_k^i(\theta)} \sim \frac{\alpha(\theta , \theta_k^*)}{\sqrt{n_k}} \chi(n_k)$, where $\alpha(\theta , \theta_k^*) = \sqrt{\|\theta - \theta_k^*\|^2 + \sigma^2}$, 
and $\chi(n_k)$ denotes a standard chi random variable with $n_k$ degrees of freedom.
Let $c_k(\boldsymbol{\theta}) = \left[ \mathbb{E}[F_k^i(\theta_1)], \mathbb{E}[F_k^i(\theta_2)], \ldots  \right]$
denote the distribution mean of the SLVs for a cluster $k$. In the following $m_k$ denotes the number of clients in cluster $k$ and $n = \min_k n_k$.

We say a  collection of $K$ models $ \boldsymbol{\theta} =  [\theta_1 , \theta_2 , \cdots \theta_K]$ is $\Delta$-proximal if each $\theta_k$ is within a distance at most $\Delta / 4$ of its optimal counterpart $\theta_k^*$. For our analysis we assume \clove is initialized with a set $ \boldsymbol{\theta}$ of $d$ randomly drawn (hence nearly ortho-normal) models.
One of our key results is the following. 
\begin{theorem}
\label{theorem:proximity:main}
For both ortho-normal or $\Delta$-proximal models collections, k-means clustering of SLVs, with suitably initialized centers, recovers accurate clustering of the clients in one shot. This result holds with probability at least $ \xi = 1 - \delta - \frac{1}{ \mbox {polylog} (M)}$, for any error tolerance $\delta$.
\end{theorem}
We prove this by showing that, for such model collections, the mixture of distributions formed by the collection of SLVs from different clients satisfies the proximity condition 
 of~\cite{kumar2010clustering}.
Specifically, it is shown in Theorem~\ref{thm:minpoints:bound} that
when the number of data points across all clients is much larger than $d, K$, 
then for any client $i$, with probability at least $\xi$: 
\begin{multline*}
\forall~{j' \neq j},~\|a_j^i(\boldsymbol{\theta}) - c_{j'}(\boldsymbol{\theta}) \| - \|a_j^i(\boldsymbol{\theta}) - c_j(\boldsymbol{\theta}) \| 
\\ 
\ge \left( \frac{c'K}{m_i} + \frac{c'K}{m_j} \right) \|A(\boldsymbol{\theta}) - C(\boldsymbol{\theta}) \|
\end{multline*}
Here $c'$ is a large constant and client $i$ is assumed to belong to cluster $j$. Theorem~\ref{theorem:proximity:main} then follows directly from the result of~\cite{kumar2010clustering}.

The proof of Theorem~\ref{thm:minpoints:bound} is based on a sequence of intermediate results. In particular, Lemma~\ref{lemma:center:sep} and Lemma~\ref{lemma:center:sep:prox} demonstrate that, for the ortho-normal and $\Delta$-proximal model collections, the means of the SLV distributions across distinct clusters are sufficiently well-separated:

For any $i \ne j$, $\|c_i(\boldsymbol{\theta}) - c_j(\boldsymbol{\theta})\| \ge \frac{ \Delta}{c} $. Here, $c \approx 2$.

Furthermore, Lemma~\ref{lemma:dist:bound} shows that with high probability, SLVs of each cluster are concentrated around their means.
We use these results, combined with a result of~\cite{dasgupta2007spectral} to prove Lemma~\ref{lemma:matrix:norm} which establishes that the norm of the matrix $X(\boldsymbol{\theta})$ of ``centered SLVs'', whose $i$-th row is $a_{s(i)}^i(\boldsymbol{\theta}) - c_{s(i)}(\boldsymbol{\theta})$, is bounded. 
That is, for $\gamma = \sqrt {d \frac{1}{2n} \left( 9 + \sigma^2 \right) }$, with high probability: $\|X(\boldsymbol{\theta}) \| \le \gamma \sqrt{M} \mbox {polylog} (M)$. By combining these results, we are able to derive the
proximity condition of Theorem~\ref{thm:minpoints:bound}.

Next, we prove our second key result, that models constructed by \clove converge exponentially fast towards their optimal counterpart. Let $\boldsymbol{\theta}^i$ denote model collection at round $i$.
\begin{theorem}
\label{thm:convg:bound} 
After $T$ rounds, each model $\theta_k \in \boldsymbol{\theta}^T$ satisfies $\| \theta_k - \theta^*_k\| \le \frac{\Delta}{c_1^T}$, with high probability, for a constant $c_1$.
\end{theorem} 

In the first round, the models $\boldsymbol{\theta}^1$ for \clove are initialized to be ortho-normal, ensuring accurate cluster recovery by Theorem~\ref{theorem:proximity:main}. 
Under the assumption that for all clusters $k$, $n_km_k \left(\frac{\rho}{\rho + 1}\right)^2 \gg \mbox {polylog} (K)$, where $\rho = \Delta^2 / \sigma^2$ (which stems from Assumption 4: $n_k m_k$, total number of data points in cluster $k$, should normally be much larger than the number of clusters $K$ as long as $\rho$ -- which is proportional to the SNR (Signal to Noise Ratio) -- is not too small), Lemma~\ref{lem:bi:matching} shows that the min-cost bi-bipartite matching found by \clove yields an optimal client-to-model match with probability close to 1. Furthermore, Lemma~\ref{lemma:gradient:update} demonstrates that as a consequence of this optimal match, the updated $K$ models $\boldsymbol{\theta}^2$, obtained by applying gradient updates from clients to their assigned models, are $\Delta$-proximal with high probability, provided that the learning rate $\eta$ is suitably chosen.

In subsequent rounds, the $\Delta$-proximality of the models $\boldsymbol{\theta}^2$ enables the application of a similar argument, augmented by Corollary~\ref{cor:smaller}. With high probability, this implies that the updated $K$ models $\boldsymbol{\theta}^3$ not only remain $\Delta$-proximal, but also have a distance to their optimal counterparts that is a constant fraction $c_1$ times smaller, resulting from the gradient updates from their assigned clients. Thus, by induction, the proof of Theorem~\ref{thm:convg:bound} follows, provided $\delta$ is chosen such that $\delta T \ll 1$.

%% file: tables/S_MCFAA_acc.tex
\begin{table*}[ht!]
\centering
\caption{Supervised test accuracy results for MNIST, CIFAR-10, FMNIST, Amazon and AG News}
\label{tab:S_MCFAA_acc}
\fontsize{9pt}{9pt}\selectfont
\resizebox{\textwidth}{!}{%
\begin{tabular}{@{}l|clll|clll|cl@{}}
\toprule
\multicolumn{1}{c|}{Algorithm} & \begin{tabular}[c]{@{}c@{}}Data\\ mixing\end{tabular} & \multicolumn{1}{c}{MNIST} & \multicolumn{1}{c}{CIFAR-10} & \multicolumn{1}{c|}{FMNIST}
& \begin{tabular}[c]{@{}c@{}}Data\\ mixing\end{tabular} & \multicolumn{1}{c}{MNIST} & \multicolumn{1}{c}{CIFAR-10} & \multicolumn{1}{c|}{FMNIST}  
& \begin{tabular}[c]{@{}c@{}}Data\\ mixing\end{tabular} & \begin{tabular}[c]{@{}c@{}}Amazon / \\ AG News\end{tabular} \\ \midrule
FedAvg  & \multirow{12}{*}{\rotatebox[origin=c]{90}{Label skew 1}}	& $87.4 \pm 0.9$ 		& $34.0 \pm 3.6$ 		& $64.7 \pm 1.2$ 
    & \multirow{12}{*}{\rotatebox[origin=c]{90}{Concept shift}}	& $52.4 \pm 0.3$ 	& $\mathbf{63.1 \pm 1.7}$ 	& $49.1 \pm 0.1$   
    & \multirow{12}{*}{\rotatebox[origin=c]{90}{None (Amazon)}}  & $\mathbf{99.0 \pm 0.1}$ \\
Local-only 	&	& $99.5 \pm 0.1$ 		& $85.4 \pm 1.6$ 		& $98.8 \pm 0.1$ 
            & 	& $94.8 \pm 0.3$ 		& $45.1 \pm 0.4$ 		& $78.3 \pm 0.4$       
            &   & $81.2 \pm 0.4$ \\ \cmidrule(l){3-5}  \cmidrule(l){7-9} \cmidrule(l){11-11}
Per-FedAvg 	&	& $97.8 \pm 0.2$ 		& $75.0 \pm 0.7$ 		& $97.0 \pm 0.1$ 
            & 	& $80.2 \pm 1.0$ 		& $34.9 \pm 0.7$ 		& $70.4 \pm 1.0$      
            &   & $87.7 \pm 0.4$ \\
FedProto 	&	& $99.2 \pm 0.0$ 		& $83.0 \pm 0.3$ 		& $98.6 \pm 0.0$ 
            & 	& $95.0 \pm 0.4$ 		& $40.8 \pm 0.3$ 		& $78.4 \pm 0.3$       
            &   & $81.3 \pm 0.1$ \\
FedALA 	    &	& $98.9 \pm 0.1$ 		& $80.0 \pm 0.6$ 		& $97.8 \pm 0.0$ 
            & 	& $96.7 \pm 0.2$ 		& $42.0 \pm 1.0$ 		& $81.1 \pm 0.3$       
            &   & $88.3 \pm 0.3$ \\
FedPAC 	    &	& $94.4 \pm 0.7$ 		& $78.3 \pm 0.4$ 		& $91.7 \pm 1.2$ 
            & 	& $90.7 \pm 1.3$ 		& $39.0 \pm 1.0$ 		& $73.8 \pm 0.6$       
            &   & $88.2 \pm 0.2$ \\ \cmidrule(l){3-5}  \cmidrule(l){7-9} \cmidrule(l){11-11} 
CFL-S 	    &	& $82.5 \pm 21.8$ 		& $87.4 \pm 0.4$ 		& $65.8 \pm 22.4$ 
            & 	& $97.5 \pm 0.2$ 		& $35.7 \pm 3.2$ 		& $83.7 \pm 0.5$       
            &   & $87.7 \pm 0.8$ \\
FeSEM 	    &   & $73.2 \pm 0.8$ 		& $11.2 \pm 1.6$ 		& $41.8 \pm 0.7$ 
            & 	& $46.2 \pm 0.1$ 		& $12.7 \pm 0.5$ 		& $39.7 \pm 0.6$       
            &   & $50.9 \pm 0.2$ \\
FlexCFL 	&	& $99.4 \pm 0.0$ 		& $73.2 \pm 0.3$ 		& $99.0 \pm 0.1$ 
            & 	& $97.4 \pm 0.1$ 		& $17.8 \pm 2.1$ 		& $\mathbf{86.1 \pm 0.2}$  
            &   & $84.1 \pm 0.8$ \\
PACFL 		&   & $99.1 \pm 0.1$ 		& $79.1 \pm 1.0$ 		& $98.7 \pm 0.1$ 
            & 	& $93.3 \pm 0.1$ 		& $26.8 \pm 1.2$ 		& $72.6 \pm 1.2$       
            &   & $82.1 \pm 0.3$ \\
IFCA 		&   & $99.1 \pm 0.5$ 		& $85.7 \pm 6.5$ 		& $97.8 \pm 1.9$ 
            &   & $80.0 \pm 0.2$ 		& $52.7 \pm 4.7$ 		& $67.0 \pm 6.2$       
            &   & $85.5 \pm 0.5$ \\
CLoVE 		&   & $\mathbf{99.8 \pm 0.0}$		& $\mathbf{90.5 \pm 0.1}$ 		& $\mathbf{99.1 \pm 0.0}$ 
            & 	& $\mathbf{97.6 \pm 0.1}$ 		& $58.4 \pm 0.6$ 		& $84.3 \pm 0.2$      
            &   & $86.8 \pm 0.4$ \\

\midrule
           
FedAvg  & \multirow{12}{*}{\rotatebox[origin=c]{90}{Label skew 2}}	& $76.4 \pm 0.3$ 	& $47.7 \pm 4.0$ 	& $55.2 \pm 1.6$ 
        & \multirow{12}{*}{\rotatebox[origin=c]{90}{Feature skew}}	& $90.0 \pm 0.6$ 	& $52.4 \pm 2.5$	& $76.5 \pm 0.3$ 
        & \multirow{12}{*}{\rotatebox[origin=c]{90}{Label skew 2 (AG News)}}    & $40.2 \pm 0.1$     \\
Local-only 	&   & $94.2 \pm 0.2$ 		& $55.9 \pm 1.8$ 		& $86.1 \pm 0.4$ 
            & 	& $93.0 \pm 0.1$ 		& $38.5 \pm 1.6$ 		& $75.8 \pm 0.2$ 
            &   & $51.4 \pm 1.1$ \\  \cmidrule(l){3-5}  \cmidrule(l){7-9}           
Per-FedAvg 	&   & $94.9 \pm 0.6$ 		& $48.2 \pm 0.9$ 		& $78.2 \pm 1.0$ 
            &   & $87.8 \pm 0.5$ 		& $22.8 \pm 0.1$ 		& $66.7 \pm 0.4$ 
            &   & $48.0 \pm 1.1$ \\
FedProto 	&   & $94.4 \pm 0.2$ 		& $56.8 \pm 0.8$ 		& $82.1 \pm 0.9$ 
            &   & $92.3 \pm 0.3$ 		& $34.1 \pm 0.2$ 		& $74.0 \pm 0.3$ 
            &   & $18.9 \pm 1.5$ \\
FedALA 		&   & $95.1 \pm 0.3$ 		& $50.1 \pm 1.6$ 		& $81.6 \pm 0.7$ 
            & 	& $90.2 \pm 0.6$ 		& $22.3 \pm 0.4$ 		& $68.1 \pm 0.5$ 
            &   & $49.6 \pm 0.1$ \\
FedPAC 		&   & $91.9 \pm 0.5$ 		& $40.7 \pm 5.2$ 		& $77.6 \pm 1.2$ 
            & 	& $87.6 \pm 0.4$ 		& $24.6 \pm 0.2$ 		& $67.4 \pm 0.2$ 
            &   & $42.4 \pm 2.9$ \\  \cmidrule(l){3-5}  \cmidrule(l){7-9} 
CFL-S 		&   & $\mathbf{97.3 \pm 0.3}$ 		& $62.1 \pm 0.4$ 		& $87.2 \pm 1.2$ 
            & 	& $95.1 \pm 0.2$ 		& $\mathbf{61.6 \pm 5.1}$ 		& $77.8 \pm 3.5$ 
            &   & $48.4 \pm 2.7$ \\
FeSEM 		&   & $89.1 \pm 1.6$ 		& $15.3 \pm 2.5$ 		& $51.4 \pm 0.2$ 
            & 	& $79.5 \pm 0.4$ 		& $10.7 \pm 0.5$ 		& $67.6 \pm 0.5$ 
            &   & $19.1 \pm 1.5$ \\
FlexCFL 	&   & $95.6 \pm 0.3$ 		& $43.6 \pm 0.6$ 		& $88.7 \pm 0.2$ 
            & 	& $96.7 \pm 0.1$ 		& $12.2 \pm 0.4$ 		& $\mathbf{85.2 \pm 0.2}$ 
            &   & $\mathbf{60.2 \pm 5.5}$ \\
PACFL 		&   & $92.0 \pm 1.1$ 		& $40.4 \pm 1.0$ 		& $81.1 \pm 0.5$
            & 	& $89.3 \pm 0.1$ 		& $17.1 \pm 0.6$ 		& $70.6 \pm 0.6$ 
            &   & $51.1 \pm 1.9$ \\
IFCA 		&   & $96.8 \pm 0.2$ 		& $63.4 \pm 3.4$ 		& $89.4 \pm 1.3$ 
            & 	& $95.9 \pm 2.0$ 		& $50.8 \pm 7.0$ 		& $83.5 \pm 0.3$ 
            &   & $51.3 \pm 3.3$ \\
CLoVE 		&   & $97.1 \pm 0.3$ 		& $\mathbf{66.8 \pm 2.3}$ 		& $\mathbf{90.1 \pm 0.3}$ 
            & 	& $\mathbf{97.7 \pm 0.1}$ 		& $57.1 \pm 1.8$ 		& $85.1 \pm 0.2$ 
            &   & $55.1 \pm 0.2$ \\ 

\bottomrule

\end{tabular}%
}
\end{table*}

%% file: sections/5-perf-eval.tex
\section{Performance Evaluation}
\label{sec:performance-evaluation}

We now show how our algorithms compare to the state of the art in both supervised and unsupervised settings.
In the unsupervised setting in particular, which is understudied in the context of CFL, our evaluation is, to the best of our knowledge, the first extensive one.
The datasets, models and data distributions we use follow the ones used by the state of the art \citep{vahidian_efficient_2023_PACFL, duan_flexible_2022_FlexCFL, xu_personalized_2023_FedPAC, zhang_FedALA_2023, IFCA_paper_2020}.
The code for \clove and our experiments is publicly available and can be found here: \url{https://github.com/Nokia-Bell-Labs/Loss-Vector-based-Clustered-Federated-Learning}.

\subsection{Experimental Setup}

\paragraph{Datasets and Models} 
We evaluate our method on three types of tasks -- image classification, text classification, and image reconstruction -- using six widely-used datasets: MNIST, Fashion-MNIST (FMNIST), CIFAR-10, FEMNIST, Amazon Reviews, and AG News. 
For the classification tasks, we use three different convolutional neural networks (CNNs): a shared CNN model for MNIST, FMNIST and FEMNIST, a deeper CNN for CIFAR-10, a TextCNN for AG News, and an MLP-based model (AmazonMLP) for Amazon Reviews. 
For the unsupervised image reconstruction tasks on MNIST and FMNIST, we use a simple autoencoder, while a convolutional autoencoder is employed for CIFAR-10. 
Unless stated otherwise, each model is initialized independently using PyTorch's default random initialization.
Further details about these datasets and model architectures can be found in Appendix~\ref{app:hyperparameters-and-datasets}.

\input{tables/S_MCF_ARI_1}

\paragraph{Dataset Partitioning} 
As in the state of the art, we extensively cover many high variance and high overlap data heterogeneity scenarios through random data partitioning, without replacement, with various types of label and feature skews, as well as concept shifts. 
For label skew, we consider many non-overlapping and overlapping label distributions. 
We present only some of the results here (rest are in Appendices~\ref{app:additional-label-skews} and~\ref{app:additional-text-experiments}).
In case \textit{Label skew 1}, clients within each cluster receive data from only two unique classes, with no label overlap between clients in different clusters. 
Conversely, in case \textit{Label skew 2}, each client within each cluster receives samples from $U$ classes, with at least $V$ labels shared between the labels assigned to any two clusters. 
For the AG News dataset, we use $U=2$ and $V=1$, while for all image datasets we use $U=4$ and $V=2$. 
As is standard practice in the FL literature, for \textit{Label skew 1} and \textit{2} (and \textit{3} in Appendix~\ref{app:additional-label-skews}), the data of a class is distributed amongst the clients that are assigned to that class by sampling from a Dirichlet distribution with parameter $\alpha=0.5$.
For instance, suppose the clients are divided into three clusters, and a particular class contains $L$ data points.
First, we draw three proportions $p_1$, $p_2$, $p_3$ from a Dirichlet distribution with concentration parameter $\alpha = 0.5$ (by construction, $p_1 + p_2 + p_3 = 1$). 
Then we randomly select $Lp_1$ points of the class for the clients in cluster 1, $Lp_2$ points for the clients in cluster 2, and the remaining $Lp_3$ points for the clients in cluster 3. 
The selected points are assigned randomly among the clients within each respective cluster.

To simulate feature distribution shifts, we apply image rotations (0°, 90°, 180°, 270°) to the MNIST, CIFAR-10 and FMNIST datasets. 
Each client within a cluster is assigned data with a specific rotation. 
Concept shift is achieved through label permutation: all labels are distributed across all clusters, but clients within each cluster receive data with two unique label swaps.
Both test and train data of a client have the same distribution.
No data mixing is employed for the Amazon Review dataset, i.e. we use its original data classes.

\paragraph{Compared Methods} 
We compare the following baselines: \local, where each client trains its model locally; \fedavg, which learns a single global model; clustered federated learning methods, including \CFL \cite{Sattler_2021_CFL}, \IFCA \cite{IFCA_paper_2020}, \textit{FlexCFL} \cite{duan_flexible_2022_FlexCFL}, \textit{FeSEM} \cite{long_multicenter_2023_FeSEM}, and \textit{PACFL} \citep{vahidian_efficient_2023_PACFL}; and general PFL methods \textit{FedProto} \citep{tan_fedproto_2022}, \textit{Per-FedAvg} \citep{fallah_personalized_2020_Per-FedAvg}, \textit{FedALA} \citep{zhang_FedALA_2023}, and \textit{FedPAC}~\citep{xu_personalized_2023_FedPAC}.
The source code we used for these algorithms is linked in Table \ref{tab:github_links}.

\paragraph{Training Settings} 
We use an Adam \citep{Adam_paper} optimizer, a batch size of 64 and a learning rate $\gamma=10^{-3}$. 
The number of local training epochs is $\tau = 1$ and the number of global communication rounds is $T=100$. 
For supervised classification tasks we use cross-entropy loss, and for unsupervised tasks we use the MSE reconstruction loss. 
We vary the number of clients between 20 and 1000. 
In most cases, the training data size per client is set to 500. 
However, for the Amazon Reviews and AG News datasets, we use 1000 and 100 samples, respectively. 
Details on the number of clients and training data sizes for all experiments is provided in Appendix~\ref{app:hyperparameters-and-datasets}.

\paragraph{Result reporting} 
We report the average performance of the models assigned to the clients on their local test data. 
Specifically, we report test accuracies for supervised classification tasks and reconstruction losses for unsupervised tasks. 
In addition, we report the accuracy of client-to-cluster assignments as the Adjusted Rand Index (ARI) between the achieved clustering and the ground-truth client groupings. 
To account for randomness, we run each experiment for 3 values of a randomness seed and report the mean and standard deviation. 
In this section, we only report results for full participation, known number of clusters, and no early stopping of clustering.
The results for the other cases, as well as scaling experiments and ablation studies, are provided in Appendix~\ref{app:additional_experimental_results} and show that \clove performs well even in these settings.

\subsection{Numerical results}

The test accuracies for image classification tasks are reported in Table~\ref{tab:S_MCFAA_acc}. The results demonstrate that \clove performs consistently well across various data distributions, including label and feature skews and concept shifts. 
While some baseline algorithms exhibit strong performance in specific cases, \textbf{\clove is the only algorithm that consistently ranks among the top performers in almost all scenarios}.

\input{tables/S_MCF_T_1}

Table~\ref{tab:S_MCF_ARI_1} presents the clustering accuracies of different CFL algorithms for the image datasets under the same experimental setup. \clove achieves optimal performance in all cases. Although \textit{FlexCFL} also shows near-optimal performance, its accuracy declines when faced with concept shifts, especially on CIFAR-10, as it neglects label information in similarity estimation, as well as when faced with feature skews (we believe the latter can be attributed to \textit{FlexCFL}'s one-shot clustering phase, which faced with a challenging setting -- where each cluster consists of images from all categories with a fixed rotation -- does not produce good clusters). In contrast, loss-based approaches like \clove excel in this metric, as model losses can effectively account for various types of skews. Results for additional types of label skews (e.g. dominant class~\citep{xu_personalized_2023_FedPAC}) are included in Appendix~\ref{app:additional-label-skews}.

The test accuracies for the textual classification tasks are also reported in Table~\ref{tab:S_MCFAA_acc} (more metrics in Appendix~\ref{app:additional-text-experiments}). While \clove exhibits good performance here as well, the baseline \textit{FedAvg} performs especially well on the Amazon Review dataset.  One reason is that while the dataset is split by product categories, the underlying language and sentiment cues are still fairly similar across all 4 categories and can provide a good signal for classification. On AG News, \clove outperforms all baselines except \textit{FlexCFL}.

Table~\ref{tab:S_MCF_T_1} presents the clustering recovery speed for the image classification datasets under different client data distributions. The first 3 columns show the ARI reached within 10 rounds for each algorithm. \clove outperforms all baselines here. 
The next 3 columns show the first round at which an ARI of 0.9 or higher is achieved (dashes mean 0.9 ARI is never achieved). 
As can be seen, \textbf{\clove consistently reaches such high accuracy within 2-3 rounds, unlike any other baseline}.

\input{tables/U_MCF_acc}

\paragraph{Unsupervised Setting}
Unlike many of the baselines, \clove is equally applicable to unsupervised settings, as it relies solely on model losses to identify clients with similar data distributions. Another loss-based approach,  \IFCA, serves as a natural point of comparison in this context. As shown in Table~\ref{tab:unsupervised_final_ARI}, \clove exhibits consistently good performance compared to \IFCA and other baselines in unsupervised settings on image reconstruction tasks using autoencoders, further highlighting its versatility and effectiveness.
More experiments in an unsupervised setting can be found in Appendix~\ref{app:additional-unsupervised-results}.

\paragraph{Robustness to initialization}
Our results indicate that one key reason \clove outperforms related baselines such as \IFCA is its robustness to initialization. We evaluate this by varying two factors: the initialization of model parameters and the first round client-to-model assignment strategy.

\begin{figure}[ht]
    \centering
    \includegraphics[width=0.7\columnwidth]{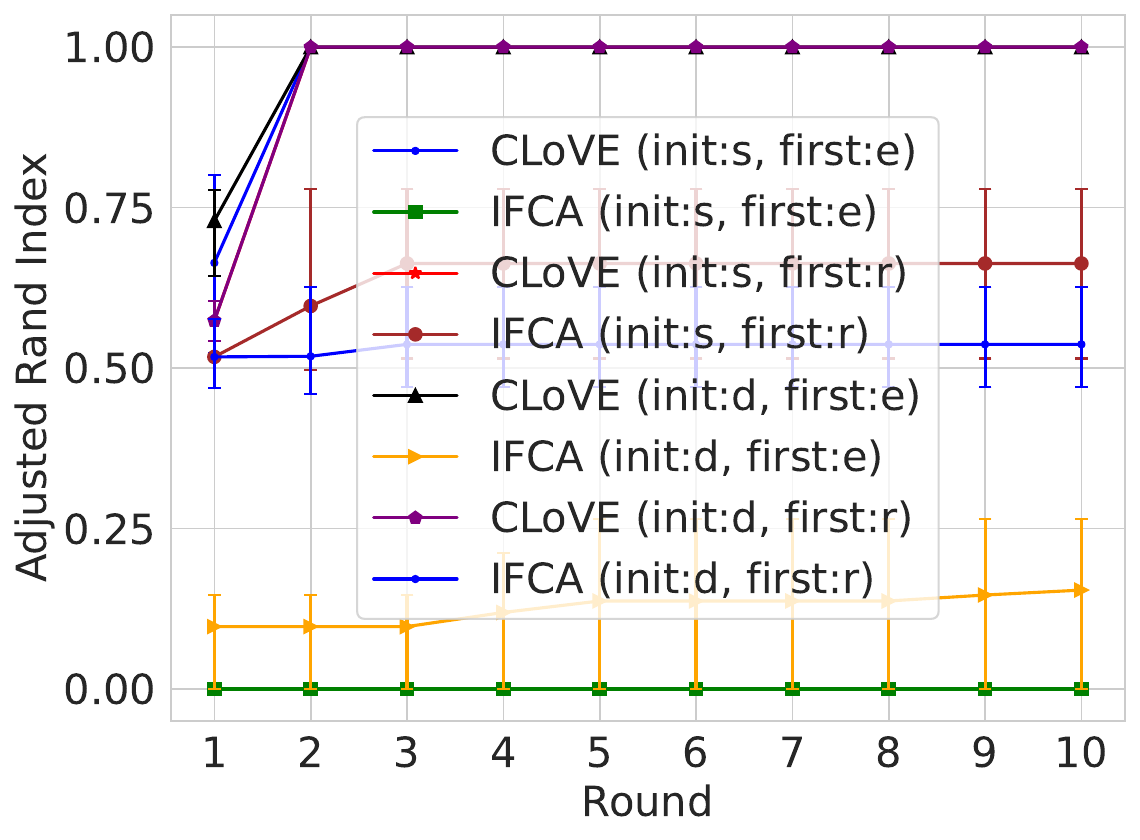}
    \caption{Clustering accuracy of \clove and \IFCA over time for the initialization experiments}
    \label{fig:initialization_experiments}
\end{figure}

For model initialization, we consider two settings: (1) all models begin with identical parameters (init:s -- same), and (2) each model is initialized independently using PyTorch's default random initialization (init:d -- different). 
For the first round client-to-model assignment, we also explore two approaches: (1) assigning clients to models at random (first:r -- random), and (2) assigning clients by clustering their loss vectors (first:e -- evaluation-based), followed by bipartite matching as in Alg. \ref{alg:CLoVE}. 
Fig. \ref{fig:initialization_experiments} compares \clove and \IFCA on unsupervised MNIST image reconstruction with 10 clusters, each containing 5 clients with 500 samples. 
The results show that \clove maintains high performance across different initialization methods, unlike \IFCA.

%% file: tables/S_MCF_ARI_1.tex
\begin{table*}[ht]
\centering
\caption{Supervised ARI results for MNIST, CIFAR-10 and FMNIST}
\label{tab:S_MCF_ARI_1}
\fontsize{9pt}{9pt}\selectfont
\begin{tabular}{@{}l|clll|clll@{}}
\toprule
\multicolumn{1}{c|}{Algorithm} & \begin{tabular}[c]{@{}c@{}}Data\\ mixing\end{tabular} & \multicolumn{1}{c}{MNIST} & \multicolumn{1}{c}{CIFAR-10} & \multicolumn{1}{c|}{FMNIST} 
& \begin{tabular}[c]{@{}c@{}}Data\\ mixing\end{tabular} & \multicolumn{1}{c}{MNIST} & \multicolumn{1}{c}{CIFAR-10} & \multicolumn{1}{c}{FMNIST}\\ \midrule

CFL-S     & \multirow{6}{*}{\rotatebox[origin=c]{90}{Label skew 1}}	& $0.71 \pm 0.20$ 	& $\mathbf{1.00 \pm 0.00}$	& $0.86 \pm 0.20$ 
        & \multirow{6}{*}{\rotatebox[origin=c]{90}{Concept shift}}    & $\mathbf{1.00 \pm 0.00}$ 		& $0.00 \pm 0.00$ 		& $\mathbf{1.00 \pm 0.00}$ \\
FeSEM 	&	& $0.18 \pm 0.00$ 		& $0.00 \pm 0.00$ 		& $0.00 \pm 0.00$ 
        &   & $0.00 \pm 0.00$ 		& $0.00 \pm 0.00$ 		& $0.00 \pm 0.00$ \\
FlexCFL &	& $\mathbf{1.00 \pm 0.00}$ 		& $\mathbf{1.00 \pm 0.00}$ 		& $\mathbf{1.00 \pm 0.00}$
        & 	& $\mathbf{1.00 \pm 0.00}$ 		& $0.22 \pm 0.24$ 		& $\mathbf{1.00 \pm 0.00}$ \\
PACFL 	&	& $0.39 \pm 0.07$ 		& $0.60 \pm 0.09$ 		& $\mathbf{1.00 \pm 0.00}$
        & 	& $0.00 \pm 0.02$ 		& $0.00 \pm 0.00$ 		& $0.00 \pm 0.00$ \\
IFCA 	&	& $0.83 \pm 0.12$ 		& $0.91 \pm 0.13$ 		& $0.92 \pm 0.12$
        & 	& $0.68 \pm 0.00$ 		& $0.76 \pm 0.17$ 		& $0.55 \pm 0.18$ \\
CLoVE 	&	& $\mathbf{1.00 \pm 0.00}$ 		& $\mathbf{1.00 \pm 0.00}$ 		& $\mathbf{1.00 \pm 0.00}$ 
        & 	& $\mathbf{1.00 \pm 0.00}$ 		& $\mathbf{1.00 \pm 0.00}$ 		& $\mathbf{1.00 \pm 0.00}$ \\ 
\midrule
CFL-S 	& \multirow{6}{*}{\rotatebox[origin=c]{90}{Label skew 2}}	& $0.00 \pm 0.00$ 		& $0.72 \pm 0.00$ 		& $0.57 \pm 0.00$ 
        & \multirow{6}{*}{\rotatebox[origin=c]{90}{Feature skew}}	& $0.48 \pm 0.00$ 		& $0.89 \pm 0.15$ 		& $0.16 \pm 0.23$ \\
FeSEM 	&	& $0.18 \pm 0.00$ 		& $0.00 \pm 0.00$ 		& $0.00 \pm 0.00$ 
        &   & $0.00 \pm 0.00$ 		& $0.00 \pm 0.00$ 		& $0.00 \pm 0.00$ \\
FlexCFL &	& $\mathbf{1.00 \pm 0.00}$ 		& $\mathbf{1.00 \pm 0.00}$ 		& $\mathbf{1.00 \pm 0.00}$ 
        & 	& $\mathbf{1.00 \pm 0.00}$ 		& $0.11 \pm 0.08$ 		& $\mathbf{1.00 \pm 0.00}$ \\
PACFL 	&	& $0.35 \pm 0.12$ 		& $0.32 \pm 0.11$ 		& $0.55 \pm 0.03$
        & 	& $0.00 \pm 0.00$ 		& $0.00 \pm 0.00$ 		& $0.68 \pm 0.11$ \\
IFCA 	&	& $0.83 \pm 0.12$ 		& $0.81 \pm 0.13$ 		& $0.92 \pm 0.12$ 
        & 	& $0.67 \pm 0.28$ 		& $0.71 \pm 0.22$ 		& $0.55 \pm 0.10$ \\
CLoVE 	&	& $\mathbf{1.00 \pm 0.00}$ 		& $\mathbf{1.00 \pm 0.00}$ 		& $\mathbf{1.00 \pm 0.00}$ 
        & 	& $\mathbf{1.00 \pm 0.00}$ 		& $\mathbf{1.00 \pm 0.00}$ 		& $\mathbf{1.00 \pm 0.00}$ \\
\bottomrule
\end{tabular}%
\end{table*}

%% file: tables/S_MCF_T_1.tex
\begin{table*}[t!]
\centering
\caption{Convergence behavior for MNIST, CIFAR-10 and FMNIST}
\label{tab:S_MCF_T_1}
\fontsize{9pt}{9pt}\selectfont
\begin{tabular}{@{}cllll|lll@{}}

\toprule
\multicolumn{1}{l}{\multirow{2}{*}[-0.075cm]{\begin{tabular}[c]{@{}c@{}}Data\\ mixing\end{tabular}}} & 
  \multicolumn{1}{c}{\multirow{2}{*}[-0.075cm]{Algorithm}} &
  \multicolumn{3}{c|}{ARI reached in 10 rounds} &
  \multicolumn{3}{c}{First round when ARI $\geq 0.9$}  \\ 
\cmidrule(l){3-8} 
&  &
  \multicolumn{1}{c}{MNIST} &
  \multicolumn{1}{c}{CIFAR-10} &
  \multicolumn{1}{c|}{FMNIST} &
  \multicolumn{1}{c}{MNIST} &
  \multicolumn{1}{c}{CIFAR-10} &
  \multicolumn{1}{c}{FMNIST} \\ 
  
\midrule
\multicolumn{1}{c}{\multirow{6}{*}{\rotatebox[origin=c]{90}{Label skew 1}}} &
	 CFL-S 		& $0.00 \pm 0.00$ 		& $0.00 \pm 0.00$ 		& $0.00 \pm 0.00$
            & --- 		& $41.3 \pm 2.1$ 		& --- \\
\multicolumn{1}{c}{} &
 FeSEM 		& $0.18 \pm 0.00$ 		& $0.00 \pm 0.00$ 		& $0.00 \pm 0.00$ 
		& --- 		& --- 		& --- \\
\multicolumn{1}{c}{} &
  FlexCFL 		& $\mathbf{1.00 \pm 0.00}$ 		& $\mathbf{1.00 \pm 0.00}$ 		& $\mathbf{1.00 \pm 0.00}$
		& $\mathbf{1.0 \pm 0.0}$ 		& $\mathbf{1.0 \pm 0.0}$ 		& $\mathbf{1.0 \pm 0.0}$ \\
\multicolumn{1}{c}{} &
  PACFL 		& $0.39 \pm 0.07$ 		& $0.60 \pm 0.09$ 		& $\mathbf{1.00 \pm 0.00}$
		& --- 		& --- 		& $\mathbf{1.0 \pm 0.0}$ \\
\multicolumn{1}{c}{} &
  IFCA 		& $0.83 \pm 0.12$ 		& $0.72 \pm 0.00$ 		& $0.83 \pm 0.12$
		& --- 		& --- 		& --- \\
\multicolumn{1}{c}{} &
  CLoVE 		& $\mathbf{1.00 \pm 0.00}$ 		& $\mathbf{1.00 \pm 0.00}$ 		& $\mathbf{1.00 \pm 0.00}$
		& $2.0 \pm 0.0$ 		& $2.0 \pm 0.0$ 		& $2.0 \pm 0.0$ \\

\midrule

\multicolumn{1}{c}{\multirow{6}{*}{\rotatebox[origin=c]{90}{Concept shift}}} &
CFL-S 		& $0.00 \pm 0.00$ 		& $0.00 \pm 0.00$ 		& $0.00 \pm 0.00$
		& $60.0 \pm 9.4$ 		& --- 		& $43.0 \pm 5.7$ \\
\multicolumn{1}{c}{} &
FeSEM 		& $0.00 \pm 0.00$ 		& $0.00 \pm 0.00$ 		& $0.00 \pm 0.00$
		& --- 		& --- 		& --- \\
\multicolumn{1}{c}{} &
FlexCFL 		& $\mathbf{1.00 \pm 0.00}$ 		& $0.22 \pm 0.24$ 		& $\mathbf{1.00 \pm 0.00}$
		& $\mathbf{1.0 \pm 0.0}$ 		& --- 		& $\mathbf{1.0 \pm 0.0}$ \\
\multicolumn{1}{c}{} &
PACFL 		& $0.00 \pm 0.02$ 		& $0.00 \pm 0.00$ 		& $0.00 \pm 0.00$
		& --- 		& --- 		& --- \\ 
\multicolumn{1}{c}{} &
IFCA 		& $0.68 \pm 0.00$ 		& $0.76 \pm 0.17$ 		& $0.55 \pm 0.18$ 
		& --- 		& --- 		& --- \\
\multicolumn{1}{c}{} &
CLoVE 		& $\mathbf{1.00 \pm 0.00}$ 		& $\mathbf{1.00 \pm 0.00}$ 		& $\mathbf{1.00 \pm 0.00}$
		& $2.0 \pm 0.0$ 		& $\mathbf{2.3 \pm 0.5}$ 		& $2.0 \pm 0.0$ \\

\bottomrule

\end{tabular}%
\end{table*}

%% file: tables/U_MCF_acc.tex
\begin{table}[H]
\centering
\caption{Unsupervised test loss results for MNIST, CIFAR-10 and FMNIST}
\label{tab:unsupervised_final_ARI}
\fontsize{9pt}{9pt}\selectfont
\resizebox{1\columnwidth}{!}{%
\begin{tabular}{@{}llll@{}}
\toprule
\multicolumn{1}{c}{Algorithm} & \multicolumn{1}{c}{MNIST} & \multicolumn{1}{c}{CIFAR-10} & \multicolumn{1}{c}{FMNIST} \\ \midrule
FedAvg &
  $0.032 \pm 0.001$ &
  $0.026 \pm 0.000$ &
  $0.030 \pm 0.000$ \\
Local-only &
  $0.019 \pm 0.000$ &
  $0.022 \pm 0.000$ &
  $0.170 \pm 0.000$ \\
IFCA &
  $0.024 \pm 0.005$ &
  $\mathbf{0.019 \pm 0.000}$ &
  $0.020 \pm 0.002$ \\
CLoVE &
  $\mathbf{0.014 \pm 0.000}$ &
  $0.021 \pm 0.000$ &
  $\mathbf{0.016 \pm 0.000}$ \\
\bottomrule
\end{tabular}%
}
\end{table}

%% file: sections/6-conclusion.tex
\section{Conclusion}
\label{sec:conclusion}

We introduced \clove, a simple loss vector embeddings-based framework for personalized clustered federated learning with low communication and computation overhead that avoids stringent model initialization assumptions and substantially outperforms the state of the art across a range of datasets and in a variety of both supervised and unsupervised settings.
Further discussion of the overheads and the privacy properties of \clove is provided in Appendix~\ref{app:discussion}.
In the future, we plan to further explore privacy, as well as fairness and adversarial behavior aspects.
Overall, \clove's design offers a promising and robust approach to scalable model personalization and clustering under heterogeneous data distributions.

%% file: sections/A1-proofs.tex
\section{Theoretical analysis}
\label{app:proofs}

\subsection{Background}
We consider a mixed linear regression problem~\citep{yi2014alternating} in a federated setting~\citep{IFCA_paper_journal_version_2022}. Client's data come from one of \(K\) different clusters, denoted as \(\mathcal{C}_1, \mathcal C_2, \dots, \mathcal C_K\). Cluster \(k\)'s optimal model  \ \(\theta^*_k \in \mathbb{R}^d\) is a vector of dimension \(d\).

{\bf Assumptions:}
We  assume all clients of a cluster \(\mathcal C_k\) use the same number $n_k$ of independently drawn feature-response pairs \((x_i, y_i)\) in each federated round. These feature-response pairs are  drawn from a distribution \(\mathcal{D}_k\), generated by the  model:
\[
y_i = \langle \theta^*_k, x_i \rangle + \epsilon_i,
\]
where \(x_i \sim \mathcal{N}(0, I_d)\) are the features, and \(\epsilon_i \sim \mathcal{N}(0, \sigma^2)\) represents additive noise. Both are independently drawn.
We assume $\sigma$ is very small. Specifically, $\sigma  \ll 1$.

We use notation $[K]$ for the set $\{1,2,...,K\}$. 

The loss function \(f(\theta)\)  is defined as the square of the error:
\[
f(\theta; x, y) = \left( \langle \theta, x \rangle - y \right)^2.
\]
The \textbf{population loss} for cluster \(k\) is:
\[
\mathcal{L}_k(\theta) = \mathbb{E}_{(x, y) \sim \mathcal{D}_k} \left[ \left( \langle \theta, x \rangle - y \right)^2 \right],
\]
Note that the optimal parameters \(\{\theta^*_k\}_{k=1}^{K}\) are the minimizers of the population losses \(\{\mathcal{L}_k\}_{k=1}^{K}\), i.e.,
\[
\theta^*_k = \arg \min_{\theta} \mathcal{L}_k(\theta) \quad \text{for each } k \in [K].
\]
We assume these optimal models have unit norm. That is $\|\theta^*_k\| = 1$ for all $k$. We also assume the optimal models are well-separated. That is, there exists a $\Delta > 0$ such that for any pair $i \ne j$:
\[
\|\theta^*_i - \theta^*_j\| \ge \Delta
\]
 For a given \(\theta\),  the \textbf{empirical loss} for client \(i\) in cluster \(k\) is:
\[
\mathcal{L}_k^i(\theta) = \frac{1}{n_k} \sum_{j=1}^{n_k} \left( \langle \theta, x_j \rangle - y_j \right)^2 
= \frac{1}{n_k} \sum_{j=1}^{n_k} \left( \langle  \theta^*_k -\theta, x_j \rangle + \epsilon_j \right)^2
\]
Let 
\[
\alpha(\theta , \theta_k^*) = \sqrt{\|\theta - \theta_k^*\|^2 + \sigma^2}.
\]
Note 
$\mathcal{L}_k^i(\theta) \sim \frac{\alpha(\theta , \theta_k^*) ^2}{n_k} \chi^2(n_k)$. That is, it is distributed as a scaled chi-squared random variable with $n_k$ degrees of freedom, with scaling factor $\frac{\alpha(\theta , \theta_k^*) ^2}{n_k}$.
In the following, we denote the square root of the empirical loss by $F_k^i(\theta)$. That is, $F_k^i(\theta) = \sqrt {\mathcal{L}_k^i(\theta)}$.  Note:
\[
F_k^i(\theta)  \sim \frac{\alpha(\theta , \theta_k^*)}{\sqrt{n_k}} \chi(n_k),
\]
where $\chi(n_k)$ denotes a chi-distributed random variable with $n_k$ degrees of freedom.

Using Sterling approximation, $\mathbb{E}[ \chi(n)] = \sqrt{n -1} \left[ 1 - \frac{1}{4n} + O\left(\frac{1}{n^2}\right) \right]$. Also,
$ \text{Var}[\chi(n)] = (n-1) - \mathbb{E}[ \chi(n)]^2 = \frac{n-1}{2n} \left[ 1 + O(\frac{1}{n}) \right]$. 
Therefore:
\[
\mathbb{E}\left[F_k^i(\theta) \right] = \alpha \left[ 1 - \frac{1}{4n_k} + O\left(\frac{1}{n_k^2}\right) \right]  \approx \alpha(\theta , \theta_k^*).
\]

\[
\text{Var}(F_k^i(\theta) ) = \frac{\alpha^2}{n_k} \left( \frac{ n_k - 1}{2n_k} + O\left(\frac{1}{n_k}\right) \right) \approx   \frac{\alpha(\theta , \theta_k^*)^2}{2n_k}.
\]

For $N$ models  $ \boldsymbol{\theta} = \begin{bmatrix} \theta_1 & \theta_2 & \cdots & \theta_N \end{bmatrix} $, the square root loss vector (SLV) of a client $i$ of cluster $k$   is $ a_k^i(\boldsymbol{\theta}) = [F_k^i(\theta_1), F_k^i(\theta_2), \ldots F_k^i(\theta_N)]$. 
Let $ c_k(\boldsymbol{\theta}) = [c^1_k, c^2_k, \ldots c^N_k] $ be the mean of $k$-th cluster's SLVs. Note that $c_k^j = \mathbb{E}[F_k^i(\theta_j)] = \alpha(\theta_j , \theta_k^*) = \sqrt{\|\theta_j - \theta_k^*\|^2 + \sigma^2} $.

All the model collections $ \boldsymbol{\theta}$ considered in this work are assumed to be one of two different types: a) $N = d$ ortho-normal models, such as $d$ models drawn randomly from a $d$-dimensional unit sphere, since such vectors are likely to be orthogonal, and b) $N = K$, $\Delta$-proximal models in which each $\theta_k$ is within a distance at most $\Delta / 4$ of its optimal counterpart $\theta_k^*$. 
Furthermore, the norm of any model in these model collections is required to be bounded. Specifically, $\|\theta \| \le 2$ for any model $\theta \in \boldsymbol{\theta}$.

\subsection{Results}

We  prove that the mean of different cluster SLVs are well-separated for our choice of model collections. We first show this for a $N=d$ ortho-normal $ \boldsymbol{\theta}$ model collection. \\
\begin{lemma}
\label{lemma:center:sep}
For ortho-normal models $ \boldsymbol{\theta} $, for any $i \ne j$, $\|c_i(\boldsymbol{\theta}) - c_j(\boldsymbol{\theta})\| \ge \dfrac{ \Delta}{c} $, where $c \approx 2$.
\end{lemma}

\begin{proof}
First, we   show the result for the following ortho-normal models:
\[
\theta_i = (0, 0, \dots, 1, \dots, 0) \quad \text{with the 1 in the $i$-th position.} 
\]
Note,
 \[
\left| \|\theta^*_i - \theta_k\|^2 - \|\theta^*_j - \theta_k\|^2 \right|  = 2\left| \langle (\theta^*_i - \theta^*_j), \theta_k\rangle\right| = 2\left| t_k \right|,
 \]
 where $t_k$ is the $k$-th component of $\theta^*_i - \theta^*_j$.
Thus 
\[
\left|(c^i_k)^2  -  (c^j_k)^2\right|=  \left| \|\theta^*_i - \theta_k\|^2 - \|\theta^*_j - \theta_k\|^2 \right|  = 2\left| t_k \right|.
 \]

Also, since $\forall_i, \|\theta^*_i\|=1$ and $\forall_k, \|\theta_k\|=1$, it follows for any $k,i$ that $c^i_k = \sqrt{\|\theta_k - \theta_i^*\|^2 + \sigma^2} \le \sqrt{4 + \sigma^2}$.
 Thus,
 \[
 \left|c^i_k  -  c^j_k\right| \ge \frac{\left| t_k \right|}{ \sqrt{4 + \sigma^2}}
 \]
 Thus, since $\sum_k t_k^2 = \|\theta^*_i - \theta^*_j \|^2$: 
 \[
\|c_i(\boldsymbol{\theta}) - c_j(\boldsymbol{\theta})\|^2 \ge \sum_k  \left( \frac{\left| t_k \right|}{ \sqrt{4 + \sigma^2}} \right)^2 \ge  \frac{\|\theta^*_i - \theta^*_j \|^2}{4  + \sigma^2}
 \]
 However, since $\forall {i \ne j}$, $\|\theta^*_i - \theta^*_j\| \geq \Delta$, it follows:
\[
\|c_i(\boldsymbol{\theta}) - c_j(\boldsymbol{\theta})\|^2 \ge \frac{ \Delta^2}{4  + \sigma^2}
\]
since, by assumption, $\sigma  \ll 1$. Thus, for $c \approx 2$,
\begin{equation}
\forall {i \ne j},  \|c_i(\boldsymbol{\theta}) - c_j(\boldsymbol{\theta})\| \ge \frac{ \Delta}{c }. 
\label{center:sep:slv}
\end{equation}
It is easy to see that the proof holds for any set \( \boldsymbol{\theta}  \) of  ortho-normal unit vectors. Specifically, since vectors  drawn randomly  from a  $d$-dimensional unit sphere are nearly orthogonal, for large $d$,  the result holds for them as well.
\end{proof}

We now prove an analogue of the Lemma~\ref{lemma:center:sep} for the $K$ models of $ \boldsymbol{\theta}$  that satisfy the proximity condition. \\
\begin{lemma}
\label{lemma:center:sep:prox}
Let $ \boldsymbol{\theta}$ be collection of  $K$ models that satisfy the $\Delta$-proximity condition. Then, for any $i \ne j$, $\|c_i(\boldsymbol{\theta}) - c_j(\boldsymbol{\theta})\| > \dfrac{ \Delta}{2}$. 
\end{lemma}

\begin{proof}
Recall,  $ c_k(\boldsymbol{\theta}) =  [c^1_k, c^2_k, \ldots c^N_k]$, where  $c_k^j = \mathbb{E}\left[F_k^i(\theta_j)\right] = \alpha(\theta_j , \theta_k^*) = \sqrt{\|\theta - \theta_k^*\|^2 + \sigma^2} $.

Since $\|\theta_i -\theta^*_i\| \le \frac{\Delta}{4}$  and since $\|\theta^*_i -\theta^*_j\| \ge \Delta$, by triangle inequality it follows that $\|\theta_i -\theta^*_j\| \ge \frac{3\Delta}{4}$.
Likewise, $\|\theta_j -\theta^*_j\| \le \frac{\Delta}{4}$ and $\|\theta_j -\theta^*_i\| \ge \frac{3\Delta}{4}$.
Thus, since $\sigma \ll \frac{\Delta}{4}$:
\begin{align*}
| c^i_j - c^i_i | &=
\left|\sqrt {\|\theta_i -\theta^*_j\|^2+ \sigma^2} - \sqrt {\|\theta_i -\theta^*_i\|^2 + \sigma^2} \right| \\ 
&\ge \left|  \|\theta_i -\theta^*_j\|^2 - \sqrt {\|\theta_i -\theta^*_i\|^2 + \left(\frac{\Delta}{4}\right)^2}\right| 
\ge \frac{3 - \sqrt 2}{4}\Delta
\end{align*} 
Likewise,
\[
| c^j_i - c^j_j | = \left|\sqrt {\|\theta_j -\theta^*_i\|^2+ \sigma^2} - \sqrt {\|\theta_j -\theta^*_j\|^2 + \sigma^2} \right|  
\ge \frac{3 - \sqrt 2}{4}\Delta
\]
Since:
\[
\|c_i(\boldsymbol{\theta}) - c_j(\boldsymbol{\theta})\|^2 \ge | c^i_j - c^i_i |^2 + | c^j_i - c^j_j |^2,
\]
it follows:
\[
\|c_i(\boldsymbol{\theta}) - c_j(\boldsymbol{\theta})\|^2 \ge 2 \left(\frac{3 - \sqrt 2}{4}\right)^2\Delta^2.
\]
Thus,
\[
\|c_i(\boldsymbol{\theta}) - c_j(\boldsymbol{\theta})\| >  \frac{ \Delta}{2} 
\]
\end{proof}

Any ortho-normal model, by definition, satisfies $\|\theta\| = 1 < 2$. Now we show the same for any $\Delta$-proximal model.\\

\begin{lemma}
$\|\theta\| < 2$, for any $\Delta$-proximal model $\theta$.
\end{lemma}
\begin{proof}
This follows from our assumption $\|\theta^*_k\| = 1$ for all $k\in [K]$ and since there exists a $k\in[K]$ such that $\|\theta -\theta^*_k\| \le \Delta / 4$, where $\Delta < 2$. Therefore, by triangle inequality, $\|\theta\| < 2$. 
\end{proof}

Since, for all models, $\|\theta\| < 2$, this implies $\|\theta -\theta^*_{k}\| \le 3$, for all $k \in [K]$. Thus, it  follows,  for all $k \in [K]$:
\begin{equation}
\label{eq:var:bound:theta}
\frac{1}{2n_k} \left( \|\theta - \theta_k^*\|^2+ \sigma^2 \right) \le \frac{1}{2n_k} \left( 9 + \sigma^2 \right)
\end{equation}
All our results below hold for any ortho-normal or  $\Delta$-proximal models collection $ \boldsymbol{\theta} $, unless noted otherwise.

The Lemma below establishes that SLVs of a cluster are concentrated around their means and this result holds with high probability. \\

\begin{lemma}
\label{lemma:dist:bound}
Let client $i$ be in cluster $s(i)$. 
Let $\mbox{Var}_j(a_{s(i)}^i)$ be the variance of the $j$-th entry of its SLV $a_{s(i)}^i(\boldsymbol{\theta})$. Then, for $t = \frac{9}{2}\log^2\frac{NM}{\delta}$ and for any error tolerance $\delta<1$:
\[
\label{eq:dist:bound}
\mathbb{P} \left[\forall i,  \mbox{  }   \|a_{s(i)}^i(\boldsymbol{\theta}) - c_{s(i)}(\boldsymbol{\theta})\|^2 \le t \max_j \mbox{Var}_j(a_{s(i)}^i) \right]  
\ge 1 - \delta.
\]
\end{lemma}

\begin{proof}

In what follows, we use $k$ to denote the cluster assignment $s(i)$ for client $i$. 
Let
\[
a_k^i(\boldsymbol{\theta}) = \left[F_k^i(\theta_1), F_k^i(\theta_2), \ldots F_k^i(\theta_N)\right],
\] 
and therefore
\[
c_k(\boldsymbol{\theta}) = \left[\mathbb{E}[F_k^i(\theta_1)], \mathbb{E}[F_k^i(\theta_2)], \ldots, \mathbb{E}[F_k^i(\theta_N)]\right].
\]

We start by focusing on the $j$-th component $ F_k^i(\theta_j)$ of  $a_k^i(\boldsymbol{\theta})$. For notational convenience, we will omit the index $j$ and refer to this component as $F_k^i(\theta)$ in the following proof.

Recall 
\[
\mathcal{L}_k^i(\theta) =\left(F_k^i(\theta)   \right)^2 = \frac{1}{n_k} \sum_{j=1}^{n_k} \left( \langle  \theta^*_k -\theta, x_j \rangle + \epsilon_j \right)^2.
\]
$\mathcal{L}_k^i(\theta) $ is scaled chi-squared random variable with $n_k$ degrees of freedom (i.e. $ \sim\frac{\alpha(\theta , \theta_k^*)^2}{n_k} \chi^2(n_k)$). Thus, its mean and variance  are:
\[
\mathbb{E} \left[\mathcal{L}_k^i(\theta)\right] = \alpha(\theta , \theta_k^*)^2 
\]
\[
\text{Var}(\mathcal{L}_k^i(\theta)) = \frac{2}{n_k} \alpha(\theta , \theta_k^*)^4
\]
Note
\[
\mathcal{L}_k^i(\theta) \sim \sum_j  \frac{1}{n_k}  \alpha(\theta , \theta_k^*)^2 y_j^2 
\]
$\mbox { where } y_j \sim \mathcal{N}(0, 1)$. Thus,
\[
\mathcal{L}_k^i(\theta) \sim \sum_j  \frac{ \mathbb{E}\left[\mathcal{L}_k^i(\theta)\right]}{n_k} y_j^2 
\]
Let $\gamma = \sqrt {\sum_j \left( \frac{ \mathbb{E}\left[\mathcal{L}_k^i(\theta)\right]}{n_k} \right)^2} =  \frac{1}{\sqrt{n_k}}\mathbb{E}\left[\mathcal{L}_k^i(\theta)\right]$. 
Let $\alpha =\frac{ \mathbb{E}[\mathcal{L}_k^i(\theta)]}{n_k} $.
We  apply Lemma 1, Section 4 of~\cite{laurent2000adaptive} to get:
\[
\mathbb{P} \left[ \mathcal{L}_k^i(\theta) - \mathbb{E}\left[\mathcal{L}_k^i(\theta)\right]  \ge  2 \alpha \sqrt {x} + 2\gamma x \right] \le e^{-x}
\]
Here 
\[
2 \alpha \sqrt {x} + 2\gamma x =  \mathbb{E}\left[\mathcal{L}_k^i(\theta)\right]\left(\frac{2 \sqrt {x} }{ n_k} + \frac{2 x }{\sqrt n_k} \right)
\]
Note that  $ \frac{3 x }{\sqrt n_k} > \frac{2 \sqrt {x} }{ n_k} + \frac{2 x }{\sqrt n_k} $ for $x > 1$ (specifically for any $x > \frac{4}{n_k}$).
Thus:
\[
\mathbb{P} \left[ \mathcal{L}_k^i(\theta) - \mathbb{E}\left[\mathcal{L}_k^i(\theta)\right]  \ge   \mathbb{E}\left[\mathcal{L}_k^i(\theta)\right]\frac{3 x }{\sqrt n_k}\right] \le e^{-x}
\]
With $3x = 2(1 + \beta)$:
\[
\mathbb{P} \left[ \mathcal{L}_k^i(\theta) \ge \mathbb{E}\left[\mathcal{L}_k^i(\theta)\right] \left(1 + 2(1 + \beta) \frac{1}{\sqrt{ n_k}} \right)\right]
\le e^{-\frac{2}{3}(1 + \beta)}
\]
Recall $\mathcal{L}_k^i(\theta) =\left(F_k^i(\theta)   \right)^2 $ and 
\[
 \mathbb{E}\left[\mathcal{L}_k^i(\theta)\right] =  \alpha(\theta , \theta_k^*)^2 = \mathbb{E}\left[F_k^i(\theta) \right]^2 
\]
Thus:
\[
\mathbb{P} \left[F_k^i(\theta)   ^2       \ge \mathbb{E}\left[F_k^i(\theta) \right]^2 \left(1+ 2(1 + \beta) \frac{1}{\sqrt{ n_k}} \right)\right]
\le e^{-\frac{2}{3}(1 + \beta)}
\]
Since
\[
 1+ 2(1 + \beta) \frac{1}{\sqrt{ n_k}}  \le \left(1+ (1 + \beta) \frac{1}{\sqrt{ n_k}} \right)^2,
\]
\[
\mathbb{P} \left[F_k^i(\theta) \ge \mathbb{E} \left[F_k^i(\theta)\right] \left(1+ (1 + \beta) \frac{1}{\sqrt{ n_k}} \right)\right]
\le e^{-\frac{2}{3}(1 + \beta)}
\]
or
\[
\mathbb{P} \left[ \left(F_k^i(\theta)   -    \mathbb{E}\left[F_k^i(\theta) \right] \right)^2    \ge  (1 + \beta)^2 \frac{\mathbb{E}\left[F_k^i(\theta) \right] ^2}{ n_k} \right] \\ \le e^{-\frac{2}{3}(1 + \beta)}
\]
Since
\[
\text{Var}\left(F_k^i(\theta) \right)  = \frac{\alpha(\theta , \theta_k^*)^2}{2n_k} =  \frac{\mathbb{E}\left[F_k^i(\theta) \right]^2}{2n_k}  ,
\]
\[
\mathbb{P}\left[ \left(F_k^i(\theta)   -  \mathbb{E}\left[F_k^i(\theta) \right] \right)^2    \ge  2(1 + \beta)^2 \text{Var}\left(F_k^i(\theta) \right)   \right] \\ \le e^{-\frac{2}{3}(1 + \beta)}
\]
We set $t = 2(1 + \beta)^2$. Then:
\[
\mathbb{P}\left[ \left(F_k^i(\theta)   -  \mathbb{E}\left[F_k^i(\theta) \right] \right)^2    \ge t \text{Var}\left(F_k^i(\theta) \right)   \right] \le e^{-\frac{\sqrt {2t}}{3}}
\]
Note that this result holds for a single component $ F_k^i(\theta_j)$ of  $A_i(\boldsymbol{\theta})$. By using union bound over all of its $N$ components:
\[
\mathbb{P}\left[ \sum_{j=1}^N \left(F_k^i(\theta_j)   -  \mathbb{E}\left[F_k^i(\theta_j) \right] \right)^2    \ge t \max_j \text{Var}\left(F_k^i(\theta_j) \right)   \right] \\ \le N e^{-\frac{\sqrt {2t}}{3}}
\]
or
\[
\mathbb{P}\left[ \left\|a_k^i(\boldsymbol{\theta}) - c_k(\boldsymbol{\theta})\right\|^2    \ge t \max_j \text{Var}\left(F_k^i(\theta_j) \right)   \right] \\  \le N e^{-\frac{\sqrt {2t}}{3}}.
\]
Since $k = s(i)$ and $\text{Var}(F_k^i(\theta_j) ) = \mbox{Var}_j(a_{s(i)}^i)$:
\[
\mathbb{P}\left[ \left\|a_{s(i)}^i(\boldsymbol{\theta}) - c_{s(i)}(\boldsymbol{\theta})\right\|^2    \ge t \max_j \ \mbox{Var}_j(a_{s(i)}^i)   \right] \\  \le N e^{-\frac{\sqrt {2t}}{3}}.
\]
Applying union bounds for all $M$ clients and by noticing that for $t = \frac{9}{2}\log^2\frac{NM}{\delta}$:
\[
M N e^{-\frac{\sqrt {2t}}{3}} =\delta
\]
we get
\[
\mathbb{P}\left[ \forall i,   \left\|a_{s(i)}^i(\boldsymbol{\theta}) - c_{s(i)}(\boldsymbol{\theta})\right\|^2    \ge t  \max_j \mbox{Var}_j(a_{s(i)}^i) \right] \\ \le \delta
\]
or
\[
\mathbb{P}\left[ \forall i,  \left\|a_{s(i)}^i(\boldsymbol{\theta}) - c_{s(i)}(\boldsymbol{\theta})\right\|^2   \le t  \max_j \mbox{Var}_j(a_{s(i)}^i)  \right]  \\ \ge 1- \delta.
\]
\end{proof}

For  models  \( \boldsymbol{\theta}  \), Let $A(\boldsymbol{\theta}) $ be the $M \times N$ matrix of the SLVs of all $M$ clients and let $C(\boldsymbol{\theta})$ be  the corresponding $M \times N$ matrix of SLV means. Thus, $X(\boldsymbol{\theta}) = A(\boldsymbol{\theta}) -C(\boldsymbol{\theta}) $ is the matrix of centered SLVs for all clients. 
Let $X_i(\boldsymbol{\theta}) = A_i(\boldsymbol{\theta})  - C_i(\boldsymbol{\theta}) $ be the $i$-th row of $A(\boldsymbol{\theta}) -C(\boldsymbol{\theta}) $, with $A_i(\boldsymbol{\theta}) $ being the SLV of the $i$-th client and $C_i(\boldsymbol{\theta}) $ the mean of that SLV.

We now establish a bound on the directional variance of the rows of $A(\boldsymbol{\theta}) -C(\boldsymbol{\theta}) $. First we show the following:\\

\begin{lemma}
\label{lemma:dir:var}
Let $X$ be an $N$-dimensional vector with components $x_i$ such that $\mathbb{E}[x_i] = 0$ for all $i$, and let $v$ be any $N$-dimensional vector. Then
\begin{equation}
   \mathbb{E}\left[ \langle X, v \rangle^2 \right] \leq \left( \sum_{i=1}^{N} |v_i| \sqrt{\text{Var}(x_i)} \right)^2.
\label{eq:dir:var}
\end{equation}

\end{lemma}

\begin{proof}

Note
\[
   \mathbb{E}\left[ \langle X, v \rangle^2 \right] = \sum_{i=1}^{N} \sum_{j=1}^{N} v_i v_j \mathbb{E}[x_i x_j] \\ \le  \sum_{i=1}^{N} \sum_{j=1}^{N} \left| v_i v_j \mathbb{E}[x_i x_j] \right|.
\]
Splitting the sum into diagonal and off-diagonal terms:
   \[
   \mathbb{E} \left[ \langle X, v \rangle^2 \right] \le \sum_{i=1}^{N} v_i^2 \mathbb{E}[x_i^2] + \sum_{i \neq j} \left|v_i v_j \mathbb{E}[x_i x_j]\right|.
   \]
Applying the Cauchy-Schwarz inequality  to the off-diagonal terms:
   \[
   \left|\mathbb{E}[x_i x_j]\right| \leq \sqrt{\mathbb{E}[x_i^2] \mathbb{E}[x_j^2]} \quad \text{for} \quad i \neq j.
   \]
Thus,
   \[
   \mathbb{E} \left[ \langle X, v \rangle^2 \right] \leq \sum_{i=1}^{N} v_i^2 \mathbb{E}[x_i^2] + \sum_{i \neq j} |v_i v_j| \sqrt{\mathbb{E}[x_i^2] \mathbb{E}[x_j^2]}.
   \]
Thus,
  \[
   \mathbb{E} \left[ \langle X, v \rangle^2 \right] \leq \left( \sum_{i=1}^{N} |v_i| \sqrt{\mathbb{E}[x_i^2]} \right)^2.
   \]
We substitute $\mathbb{E}\left[x_i^2\right] = \text{Var}(x_i)$ (since $\mathbb{E}[x_i] = 0$) to express the bound in terms of variances:
   \[
   \mathbb{E} \left[ \langle X, v \rangle^2 \right] \leq \left( \sum_{i=1}^{N} |v_i| \sqrt{\text{Var}(x_i)} \right)^2.
   \]
\end{proof}

\begin{lemma}
\label{lemma:dir:bound}
Let $X_i(\boldsymbol{\theta}) = A_i(\boldsymbol{\theta})  - C_i(\boldsymbol{\theta}) $ be the $i$-th row of $A(\boldsymbol{\theta}) -C(\boldsymbol{\theta}) $, with $A_i(\boldsymbol{\theta}) $ being the SLV of $i$-th client and $C_i(\boldsymbol{\theta}) $ the mean of that SLV. Let $i$-th client be in in cluster $k$.
\begin{equation}
\max_{v: \|v\|=1} \mathbb{E} \left[ \langle X_i(\boldsymbol{\theta}) , v \rangle ^2 \right] \le N \frac{1}{2n_k} \left( 9 + \sigma^2 \right)
\label{eq:dir:bound}
\end{equation}
\end{lemma}

\begin{proof}
Applying Equation~\ref{eq:dir:var} of Lemma~\ref{lemma:dir:var}, it follows:
\[
\max_{v: \|v\|=1} \mathbb{E} \left[ \langle X_i(\boldsymbol{\theta}) , v \rangle ^2 \right] \\ \le \max_{v: \|v\|=1} \left( \sum_{j=1}^{N} |v_j| \sqrt{\text{Var}(X_i(\theta_j))} \right)^2,
\]
where $X_i(\theta_j)$ is the $j$-th component of $X_i(\boldsymbol{\theta})$, which is the centered square root loss of client $i$ on model $\theta_j$. That is  $X_i(\theta_j) = A_i(\theta_j) - \mathbb{E}[A_i(\theta_j)]$. Therefore, $\text{Var}(X_i(\theta_j) ) = \text{Var}(A_i(\theta_j))$. Hence:
\[
\max_{v: \|v\|=1} \mathbb{E} \left[ \langle X_i(\boldsymbol{\theta}) , v \rangle ^2 \right]  \le  \max_{v: \|v\|=1} \left( \sum_{j=1}^{N} |v_j| \sqrt{\text{Var}(A_i(\theta_j))} \right)^2 
\]
Combining with Equation~\ref{eq:var:bound:theta}:
\[
\text{Var}(A_i(\theta_j))=  \frac{1}{2n_k} \left( \|\theta_j - \theta_k^*\|^2+ \sigma^2 \right) \le \frac{1}{2n_k} \left( 9 + \sigma^2 \right) .
\]
Thus 
\[
\max_{v: \|v\|=1} \mathbb{E} \left[ \langle  X_i(\boldsymbol{\theta}) , v \rangle ^2 \right] \le N \frac{1}{2n_k} \left( 9 + \sigma^2 \right)
\]
since maximum happens when all $v_j = \frac{1}{\sqrt N}$.
\end{proof}

Using a similar argument, it follows:
\begin{equation}
\mathbb{E} \left[ \|X_i(\boldsymbol{\theta}) \|^2 \right] \le N \frac{1}{2n_k} \left( 9 + \sigma^2 \right)
\label{eq:var:bound}
\end{equation}

We now establish a bound on the matrix operator norm $ \|X(\boldsymbol{\theta})  \| = \|A(\boldsymbol{\theta}) -C(\boldsymbol{\theta}) \|$. Let $n = \min_k n_k$ be the smallest number of data points any client has.
Let $\gamma = \sqrt {N \frac{1}{2n} \left( 9 + \sigma^2 \right) }$. Then we can show the following bound: \\
\begin{lemma}
\label{lemma:matrix:norm}
With  probability at least $1 - \frac{1}{ \mbox {polylog} (M)}$:
\begin{equation}
\label{eq:matrix:norm}
\|X(\boldsymbol{\theta})  \| \le \gamma \sqrt{M}  \mbox {polylog} (M)
\end{equation}
\end{lemma}

\begin{proof}
With abuse of notation, we will use $X_i$ to refer to $i$-th row of $\mathbf{X}(\boldsymbol{\theta})$ and use $X_{ij}$ to denote the $j$-th entry of this row.
Since $\text{Var} (\|X_i\|) \le E[\|  X_i \|^2]$, then by applying 
Equation~\ref{eq:var:bound} and by definition of $\gamma$, it follows $\text{Var} (\|X_i\|) \le E[\|  X_i \|^2] \le \gamma^2$.
Thus, by Chebyshev's inequality:
\[
\mathbb{P}\left[\|  X_i \| \ge \gamma \sqrt{M} \mbox {polylog} (M) \right] \le \frac{\gamma^2}{M \gamma^2\mbox {polylog} (M)}  
\]
Thus by union bound
\[
\mathbb{P}\left[\max_{i=1, \dots, M} \|  X_i \| \ge \gamma \sqrt{M} \mbox {polylog} (M) \right] \le \frac{1}{ \mbox {polylog} (M)}
\]
Thus with probability at least $1 - \frac{1}{ \mbox {polylog} (M)}$:
\[
\max_{i=1, \dots, M} \|  X_i \| < \gamma \sqrt{M} \mbox {polylog} (M )
\]
Now  for any unit column vector $v$  
\[
v^TE\left[ \mathbf{X}(\boldsymbol{\theta})^T\mathbf{X}(\boldsymbol{\theta}) \right] v  = \sum_{i=1}^M E\left[v^TX_i^T  X_iv \right]  \\ = \sum_{i=1}^M E\left[ \|X_i v\|^2 \right] =  \sum_{i=1}^M E\left[\langle X_i  , v \rangle ^2 \right]
\]
Thus from Equation~\ref{eq:dir:bound} of Lemma~\ref{lemma:dir:bound} and by definition of $\gamma$, it follows:
\[
\max_{v: \|v\|=1}  v^TE\left[ \mathbf{X}(\boldsymbol{\theta})^T\mathbf{X}(\boldsymbol{\theta}) \right] v   \le \gamma^2 M
\]
Since $E\left[ \mathbf{X}(\boldsymbol{\theta})^T\mathbf{X}(\boldsymbol{\theta}) \right]$ is symmetric, its spectral norm equals its largest eigenvalue, which is what is bounded above.
Thus 
\[
\left\|E\left[ \mathbf{X}(\boldsymbol{\theta})^T\mathbf{X}(\boldsymbol{\theta}) \right]\right\| \le  \gamma^2 M
\]
We now use the following fact (Fact 6.1) from~\cite{kumar2010clustering} that follows from the result of~\cite{dasgupta2007spectral}.  \\

\begin{lemma}
Let $Y$ be a $n \times d$ matrix, $ n \ge d$, whose rows are chosen independently. Let $Y_i$ denote its $i$-th row.
Let there exist $\gamma$  such that $\max_i \|Y_i\|  \le \gamma \sqrt {n}$ and $\| E\left [Y^T Y\right] \| \le \gamma^2 n$. Then $\|Y\|  \le \gamma \sqrt {n} \mbox {polylog} (n)$.
\end{lemma}
Thus, it follows that with probability at least $1 - \dfrac{1}{ \mbox {polylog} (M)}$:
\[
\|\mathbf{X}(\boldsymbol{\theta})\| \le \gamma \sqrt{M}  \mbox {polylog} (M)
\]
\end{proof}

We now establish one of our main results, that the SLVs satisfy the proximity condition of~\cite{kumar2010clustering}, thereby proving that k-means with suitably initialized centers recovers the accurate clustering of the SLVs (i.e. it recovers an accurate clustering of the clients).

Recall that for client $i$ of cluster $k$, its SLV is $ a_k^i(\boldsymbol{\theta}) = \left[F_k^i(\theta_1), F_k^i(\theta_2), \ldots F_k^i(\theta_N)\right]$. Also $c_k(\boldsymbol{\theta}) = \left[c^k_1, c^k_2, \ldots c^k_N\right], $ is the mean of $k$-th clusters SLVs, where  $c_k^j = \mathbb{E}\left[F_k^i(\theta_j)\right]$. Recall $X(\boldsymbol{\theta}) = A(\boldsymbol{\theta}) -C(\boldsymbol{\theta}) $ is the matrix of centered SLVs for all clients. Also, $X_i(\boldsymbol{\theta}) = A_i(\boldsymbol{\theta})  - C_i(\boldsymbol{\theta}) $ denotes the $i$-th row of $A(\boldsymbol{\theta}) -C(\boldsymbol{\theta}) $. Let $m_k$ denote the number of clients in cluster $k$.
\begin{theorem}
\label{thm:minpoints:bound}
When $nM$, which is a lower bound on the the number of data points across all clients,  is  much larger in comparison to $d, K$,  and specifically when
\begin{equation}
\label{eq:minpoints:bound}
d K^4  \mbox {  polylog} (M) =  o(nM),
\end{equation}
then for all ortho-normal or proximal models  $ \boldsymbol{\theta} $, the following proximity condition of~\cite{kumar2010clustering} holds  for the SLV of any client $i$, with probability at least $ 1 -  \delta -  \frac{1}{ \mbox {polylog} (M)}$, for any error tolerance $\delta$ :
\[
\forall~j' \neq j,~\left\|a_j^i(\boldsymbol{\theta})  - c_{j'}(\boldsymbol{\theta}) \right\| - \left\|a_j^i(\boldsymbol{\theta})  - c_j(\boldsymbol{\theta}) \right\| \\ \ge \left( \frac{c'K}{m_i} +  \frac{c'K}{m_j} \right) \left\|A(\boldsymbol{\theta})  - C(\boldsymbol{\theta}) \right\|
\]
for some large constant $c'$. Here it is assumed that client $i$ belongs to cluster $j$.
\end{theorem}

\begin{proof}

Since $  \mbox{Var}(A_i(\theta_j)) = \frac{1}{2n_k} \left( \|\theta_i - \theta_k^*\|^2 + \sigma^2 \right)$, it follows that
$\forall~i,j,~\mbox{Var}(A_i(\theta_j)) \le \frac{1}{2n} \left( 9 + \sigma^2 \right) = \gamma^2 / N$, where $n =  \min_k n_k$.
Combining this with Equation~\ref{eq:dist:bound} established in Lemma~\ref{lemma:dist:bound}, it follows with high probability (at least $1 - \delta$):
\[
\forall~i,~ \|A_i(\boldsymbol{\theta})  - C_i(\boldsymbol{\theta}) \|^2 \le \frac{9}{2}\log^2\frac{NM}{\delta} \max_j \mbox{Var}(A_i(\theta_j)) 
\\ \le \frac{9}{2}\log^2\frac{NM}{\delta} \frac{\gamma^2}{ N}
,
\]
where $A_i(\boldsymbol{\theta})$ and $ C_i(\boldsymbol{\theta})$  are the $i$-th rows of matrices $A(\boldsymbol{\theta}) $ and  $C(\boldsymbol{\theta})$ respectively. 

Without loss of generality, let the SLV of client $i$ belonging to cluster $j$ be in the $i$-th row of the matrix $A(\boldsymbol{\theta})$.
Thus, it follows:
\[
 \left\|a_j^i(\boldsymbol{\theta})  - c_j(\boldsymbol{\theta}) \right\|  =  \|A_i(\boldsymbol{\theta})  - C_i(\boldsymbol{\theta}) \|
\\ \le  \frac{3}{\sqrt 2} \frac{1}{\sqrt N} \log\frac{NM}{\delta} \gamma.
\]

Let
\[
\zeta = \frac{3}{\sqrt 2} \frac{1}{\sqrt N} \log\frac{NM}{\delta} \gamma
\]

Since models in $ \boldsymbol{\theta} $ collection are ortho-normal or   $\Delta$-proximal, from Lemma~\ref{lemma:center:sep} and Lemma~\ref{lemma:center:sep:prox} it follows that for any $j' \ne j$, $\|c_{j'}(\boldsymbol{\theta}) - c_j(\boldsymbol{\theta})\| \ge \Delta / c $, where $c \approx 2$.

Thus, by triangle inequality, for any $j' \neq j$:
\begin{equation}
\label{eq:proximity:sep}
\left\|a_j^i(\boldsymbol{\theta})  - c_{j'}(\boldsymbol{\theta}) \right\| - \left\|a_j^i(\boldsymbol{\theta})  - c_j(\boldsymbol{\theta}) \right\| \\ \ge \frac{ \Delta}{c } - 2 \zeta 
\end{equation}

We now show that:
\[
\frac{ \Delta}{c } - 2 \zeta  \ge \left( \frac{c'K}{m_i} +  \frac{c'K}{m_j} \right) \gamma \sqrt{M}  \mbox {polylog} (M)
\]

By  assumption, $N \le d \ll M$. Thus, in 
\[
 2 \zeta  + \left( \frac{c'K}{m_i} +  \frac{c'K}{m_j} \right) \gamma \sqrt{M}  \mbox {polylog} (M) \\ = \frac{3}{\sqrt 2} \frac{1}{\sqrt N} \log\frac{NM}{\delta} \gamma +  \left( \frac{c'K}{m_i} +  \frac{c'K}{m_j} \right) \gamma \sqrt{M}  \mbox {polylog} (M) ,
\]
the second term dominates.
Also, we make the assumption that number of clients in each cluster is  not too far from the average number of clients per cluster $M / K$. 
That is, there  exists a small constant $c''$ such that the number of clients $m_k$ in any cluster $k$ is at least $\frac{1}{c''}\frac{M}{K}$.
Hence:
\[
 2 \zeta  + \left( \frac{c'K}{m_i} +  \frac{c'K}{m_j} \right) \gamma \sqrt{M}  \mbox {polylog} (M) \\ = O\left( \gamma \frac{K^2}{\sqrt M }  \mbox {polylog} (M)\right) 
\]
Since $\sigma \ll 1$ is very small 
and $K \le N \le d$,  it follows that
$\gamma \le \sqrt {\frac{2d}{n} }  $. 
Thus:
\[
 2 \zeta  + \left( \frac{c'K}{m_i} +  \frac{c'K}{m_j} \right) \gamma \sqrt{M}  \mbox {polylog} (M) \\ =  O\left (\frac{\sqrt d}{\sqrt n}  \frac{K^2}{\sqrt M }  \mbox {polylog} (M)  \right)
\]

Since by assumption $\Delta$ is close to $1$ and
 \[
d K^4  \mbox {  polylog} (M) =  o(nM), 
\]
it follows that
$\frac{\Delta}{c }$ (where $c \approx 2$) is much larger than $O\left (\frac{\sqrt d}{\sqrt n}  \frac{K^2}{\sqrt M } \mbox {polylog} (M)  \right)$.

This establishes, with probability at least $1 - \delta$:
\[
\frac{ \Delta}{c }  \ge  2 \zeta + \left( \frac{c'K}{m_i} +  \frac{c'K}{m_j} \right) \gamma \sqrt{M}  \mbox {polylog} (M)
\]

Applying Equation~\ref{eq:matrix:norm} of Lemma~\ref{lemma:matrix:norm}, it follows  with  probability at least $ 1 -  \delta -  \dfrac{1}{ \mbox {polylog} (M)}$ that:
\[
\frac{ \Delta}{c} -  2 \zeta  \ge \left( \frac{c'K}{m_i} +  \frac{c'K}{m_j} \right) \|A(\boldsymbol{\theta}) -C(\boldsymbol{\theta}) \|
\]

Combining with Equation~\ref{eq:proximity:sep}, the proximity result follows. That is, with probability at least $ 1 -  \delta -  \dfrac{1}{ \mbox {polylog} (M)}$:
\[
\forall~j' \neq j,~\left\|a_j^i(\boldsymbol{\theta})  - c_{j'}(\boldsymbol{\theta}) \right\| - \left\|a_j^i(\boldsymbol{\theta})  - c_j(\boldsymbol{\theta}) \right\| \\ \ge \left( \frac{c'K}{m_i} +  \frac{c'K}{m_j} \right) \|A(\boldsymbol{\theta})  - C(\boldsymbol{\theta}) \|
\]
\end{proof}

This yields one of our main results (Theorem~\ref{theorem:proximity:main} in the main part of the paper): \\

\begin{theorem}
For ortho-normal or  $\Delta$-proximal models collection $\boldsymbol{\theta}$, with  probability at least  $ 1 -  \delta -  \dfrac{1}{ \mbox {polylog} (M)}$, k-means with suitably initialized centers recovers the accurate clustering of the SLVs (and hence an accurate clustering of the clients) in one shot.
\end{theorem}

\begin{proof}
Follows from Lemma~\ref{lemma:center:sep} and Theorem~\ref{thm:minpoints:bound} and by applying the result of~\cite{kumar2010clustering}.
\end{proof}

\begin{remark}
There have been further improvements to the proximity bounds of~\cite{kumar2010clustering} (e.g. \cite{awasthi2012improved}) that can be used to improve our bounds (Equation~\ref{eq:minpoints:bound}). However, since our goal in this work is  mainly to establish feasibility, we leave any improvements of those bounds for future work.
\end{remark}

\begin{remark}
In the proof above we used loss vectors of square roots of losses (SLV). This is because it allows us to recover the accurate clustering of SLVs in one round of k-means. For loss vectors (LV) formed by losses instead of square roots of losses, we are only able to  prove that the distance bound as established in Lemma \ref{lemma:dist:bound} holds for a large fraction of the rows $A(\boldsymbol{\theta}) $, which means that k-means can  recover accurate clustering of not all but still a large number of clients. This is still fine for our approach, since we do not need to apply updates from all clients to build the models for the next round. However, it does mean that the analysis gets more complex and we therefore leave that for future work. 
\end{remark}

In the following, let the set of clients in cluster $k$ be denoted by $Q_k$. Let $i\in Q_k$ be a client in cluster $k$.  Let $H_k^i$ denote the set of $n_k$ data points and additive noise values $(x_j, y_j, \epsilon_j)$ of client $i$. Its empirical loss is 
\[
\mathcal{L}_k^i \left(\theta; H_k^i \right) = \frac{1}{n_k} \sum_{(x_j, y_j, \epsilon_j) \in H_k^i} \left( \langle \theta, x_j \rangle - y_j \right)^2  \\ = \frac{1}{n_k} \sum_{(x_j, y_j, \epsilon_j) \in H_k^i} \left( \langle  \theta^*_k -\theta, x_j \rangle + \epsilon_j \right)^2
\]
Let $H_k$ be the $m_k n_k \times d$ matrix with rows being the features  of clients in cluster $k$:
\[
H_k = \left[ x_j : (x_j, y_j, \epsilon_j)  \in \bigcup_{j \in Q_k} H_k^j \right]
\]
Note that all rows of $H_k$ are drawn independently from $\mathcal{N}(0, I_d)$.  
Let $\xi_k$ be the $m_k n_k \times 1$ column vector of additive noise parameters of clients in cluster $k$:
\[
\xi_k = \left[ \epsilon_j : (x_j, y_j, \epsilon_j) \in \bigcup_{j \in Q_k} H_k^j  \right]
\]
Note that all entries of $\xi_k$ are drawn independently from $\mathcal{N}(0, \sigma^2)$.\\

\begin{lemma}
\label{lemma:gradient:update}
After applying  gradient updates of  clients belonging to a cluster $k$ to a  model $\theta, \|\theta\| \le 2$,  the new model $\theta_k$ satisfies $\|\theta^*_k - \theta_k\| \le \Delta / c$ with probability at least $1 - c_6 e^{-c_7 d}$, for some constants $c_6, c_7$. Furthermore, with probability at least $1 -  c_6 K e^{-c_7 d}$, $\|\theta^*_k - \theta_k\| \le \frac{\Delta}{c}$, for all clusters $k$. Here learning parameter $\eta$ is assumed to be at most $  \frac{1}{2} \left( 1 - \frac{\Delta}{3c } \right) $. 
\end{lemma}

\begin{proof}

Note that 
\[
\nabla\mathcal{L}_k^i \left(\theta; H_k^i \right) =  \frac{1}{n_k}  \sum_{(x_j, y_j) \in H_k^i}  \left( 2 x_j x_j^T ( \theta - \theta^*_k ) - 2x_j \epsilon_j \right) 
\]
Gradient updates from clients $Q_k$ of cluster $k$ on  model $\theta$ for learning rate $\eta$ yield model $\theta_k$:
\[
\theta_k = \theta - \frac{\eta}{m_k} \sum_{i \in Q_k} \nabla\mathcal{L}_k^i(\theta; H_k^i)
\]
Thus, as rows of matrix $H_k$ contain features of all clients in cluster $k$, and since entries of vector $\xi_k$ contain corresponding additive noise parameters of those clients:
\[
\theta_k = \theta - \frac{\eta}{m_k n_k} \left( 2 H_k^T H_k( \theta - \theta^*_k ) - 2H_k^T \xi_k \right) 
\]
Thus,
\[
\theta_k -\theta^*_k = \theta - \theta^*_k  - \frac{2\eta}{m_k n_k}  H_k^T H_k ( \theta - \theta^*_k ) + \frac{2\eta}{m_k n_k}  H_k^T \xi_k
\]
Or
\[
\theta_k -\theta^*_k = \theta - \theta^*_k - \frac{2\eta}{m_k n_k} \mathbb{E}[H_k^T H_k] ( \theta - \theta^*_k ) \\ + \frac{2\eta}{m_k n_k} \left(\mathbb{E}[H_k^T H_k] -  H_k^T H_k \right) ( \theta - \theta^*_k ) \\ + \frac{2\eta}{m_k n_k}  H_k^T \xi_k
\]
Note $ \mathbb{E}\left[H_k^T H_k\right] = m_k n_k I$. Therefore:
\[
\theta_k -\theta^*_k= \left(1 - 2\eta \right) ( \theta - \theta^*_k ) \\ + \frac{2\eta}{m_k n_k} \left(\mathbb{E}\left[H_k^T H_k\right] -  H_k^T H_k \right) ( \theta - \theta^*_k ) \\ + \frac{2\eta}{m_k n_k}  H_k^T \xi_k
\]
Taking norms:
\[
\left\|\theta_k -\theta^*_k\right\| \le |1 - 2\eta | \| \theta - \theta^*_k \| \\ + \frac{2\eta}{m_k n_k} \left\|\mathbb{E}\left[H_k^T H_k\right] -  H_k^T H_k \right\| \| \theta - \theta^*_k \| \\ + \frac{2\eta}{m_k n_k}  \left\|H_k^T \xi_k\right\|
\]
We now bound the various terms on the RHS.

Since $\|\theta\| \le 2$ and $\| \theta^*_k \| \le 1$:
\[
 \| \theta - \theta^*_k \|  \le 3.
\]
As shown in~\cite{IFCA_paper_journal_version_2022}, following the result of~\cite{wainwright2019high}, with probability at least $1 - 2e^{-\frac {d}{2}}$, the operator norm of the matrix $\mathbb{E}\left[H_k^T H_k\right] -  H_k^T H_k$ is bounded from above:
\[
\left\|\mathbb{E}\left[H_k^T H_k\right] -  H_k^T H_k \right\| \le 6 \sqrt {d m_k n_k}.
\]
By assumption $d \ll m_kn_k$. Therefore, 
\[
\left\|\mathbb{E}\left[H_k^T H_k\right] -  H_k^T H_k \right\|  = o(m_kn_k).
\]
and therefore 
$\frac{2\eta}{m_k n_k} \left\|\mathbb{E}[H_k^T H_k] -  H_k^T H_k \right\| \| \theta - \theta^*_k \| \le c_2 \Delta$, for some $c_2 \ll 1$.

As shown in~\cite{IFCA_paper_journal_version_2022}, with probability at least   $1 - c_4 e^{-c_5 d}$, for some constants $c_4, c_5$.
\[
\left\|H^T \xi_k \right\| \le c_1 \sigma \sqrt {d m_k n_k} = o(\sigma m_kn_k),
\] 
for some constant $c_1$. Since $\sigma \ll 1$, $\frac{2\eta}{m_k n_k}  \left\|H_k^T \xi_k \right\| \le c_3 \Delta$, for some $c_3 \ll 1$.

Combining all the bounds, with probability at least $1 - c_6 e^{-c_7 d}$, for some constants $c_6, c_7$:
\[
\label{eq:gradient:bound}
\|\theta_k -\theta^*_k\| \le 3|1 - 2\eta | + (c_2 + c_3) \Delta.
\]
For  $\eta \le  \frac{1}{2} \left( 1 - \frac{\Delta}{3c } \right) $, $3|1 - 2\eta |  \le \frac{\Delta}{c}$. Thus, with appropriate choice of $\eta$,  with probability at least $1 - c_6 e^{-c_7 d}$:
\[
\|\theta_k -\theta^*_k\| \le \frac{\Delta}{c} + (c_2 + c_3) \Delta < \frac{\Delta}{c}.
\]
Thus, by applying union bound, it follows that with probability at least $1 -  c_6 K e^{-c_7 d}$, $\|\theta^*_k - \theta_k\| \le \frac{\Delta}{c}$, for all clusters $k$.
\end{proof}

We set $c= 4$ and $\eta =  \frac{1}{2} \left( 1 - \frac{\Delta}{12} \right) $. Thus, $\|\theta^*_k - \theta_k\| \le \frac{\Delta}{4}$ for all clusters $k$ with probability at least $1 - c_6 K e^{-c_7 d}$.\\

\begin{corollary}
\label{cor:smaller}
After applying the gradient updates as in Lemma~\ref{lemma:gradient:update}, with probability at least $1 - c_6 K e^{-c_7 d}$, $\|\theta_k -\theta^*_k\|$ is a constant fraction smaller than $ \| \theta - \theta^*_k \|$ . The result  applies to all models $\theta$ whose norm is  less than $2$.
\end{corollary}
\begin{proof}

Note that Equation~\ref{eq:gradient:bound} can be restated as:
\[
\|\theta_k -\theta^*_k\| \le |1 - 2\eta | \| \theta - \theta^*_k \| + (c_2 + c_3) \Delta.
\]
For our choice of $\eta$,
\[
\|\theta_k -\theta^*_k\| \le \frac{\Delta}{3 c}\| \theta - \theta^*_k \| + (c_2 + c_3) \Delta.
\]
Since $\Delta$ is close to $1$ and since  $c \ge 4$, and since $(c_2 + c_3) \Delta$ is negligibly small, it follows that  $\|\theta_k -\theta^*_k\|$ is a constant fraction smaller than $ \| \theta - \theta^*_k \|$, as long as  $ \| \theta - \theta^*_k \|$ is much larger than $(c_2 + c_3) \Delta$.

\end{proof}

In the following, $s(i) \in [1,K]$ denotes the ground-truth cluster assignment for client $i$.
Let $Q$ be a clustering of the clients. That is, $Q = \left[ Q_1, Q_2, \ldots, Q_K \right]$, where $Q_k$ is the set of clients in the $k$-th cluster of $Q$. Let $G(Q; \boldsymbol{\theta}) = (U \cup V, E)$ be a fully-connected bi-partite graph.  Here, $U$ and $V$ are sets of nodes of size $K$ each, where $k$-th node of $U$ (or $V$) corresponds to the $k$-th cluster of $Q$. $E$ is the set of edges, with weight $w(k,j)$ of edge $(k, j)$ set to the total loss of clients in $Q_k$ on  model $\theta_j$ of  $ \boldsymbol{\theta} $. That is:
\[
w(k,j) = \sum_{ i \in Q_k} \mathcal{L}_{s(i)}^i(\theta; H_{s(i)}^i).
\]
\\
\begin{assumption}
$n_km_k (\frac{\rho}{\rho + 1})^2 \gg  \mbox {polylog} (K)  $,  for $\rho = \frac{\Delta^2}{\sigma^2}$.
In the following, $ \boldsymbol{\theta} = \begin{bmatrix} \theta_1 & \theta_2 & \cdots & \theta_K \end{bmatrix} $ refers to a collection of $K$ models that satisfy the $\Delta$-proximity condition 
  $\forall k \in [K], \|\theta_k -\theta^*_k\| \le \frac{\Delta}{c}$,  for $c \ge 4$.\\
\end{assumption}
\begin{lemma}
\label{lem:bi:matching}
Let $ \boldsymbol{\theta}$ be a collection of $K$ models that satisfy the $\Delta$-proximity condition. Let $Q$ be an accurate clustering of clients to clusters. Then,   the probability that min-cost perfect matching $M$ of $G(Q; \boldsymbol{\theta})$ is $M = \{(k,k): Q_k \in Q\}$ is at least $ 1 -  c_1 e ^{-c_2 \frac{1}{\log K^2} n_km_k (\frac{\rho}{\rho + 1})^2} $. 
\end{lemma}
\begin{proof}
Let clients of  a cluster $Q_k$ achieve lower loss on a model  $\theta_j$ than on model $\theta_k$, for $j \ne k$. Note that the probability of this is:
\[
\mathbb{P}\left[\sum_{ i \in Q_k} \mathcal{L}_{s(i)}^i\left(\theta_j; H_{s(i)}^i\right) < \sum_{ i \in Q_k} \mathcal{L}_{s(i)}^i\left(\theta_k; H_{s(i)}^i\right)\right]
\]
Since $\sum_{ i \in Q_k} \mathcal{L}_{s(i)}^i(\theta; H_{s(i)}^i) $ is scaled chi-squared random variable with $n_km_k$ degrees of freedom (i.e. $ \sim \frac{m_k}{n_k} (\|\theta - \theta_k^*\|^2 + \sigma^2) \chi^2(n_km_k)$), 
this probability is:
\[
\mathbb{P} \left[ \|\theta_j - \theta_k^*\|^2 + \sigma^2) \chi^2(n_km_k) \\ <  (\|\theta_k - \theta_k^*\|^2 + \sigma^2) \chi^2(n_km_k) \right]
\]

Since $\|\theta_k -\theta^*_k\| \le \frac{\Delta}{4} = \frac{\Delta}{2} - \frac{\Delta}{4}$  and since $\|\theta^*_k -\theta^*_j\| \ge \Delta$, by triangle inequality it follows that $\|\theta_k -\theta^*_j\| \ge \frac{3\Delta}{4} = \frac{\Delta}{2} - \frac{\Delta}{4}$.
Thus, from the result of \cite{IFCA_paper_journal_version_2022}, which is obtained by applying the concentration properties of chi-squared random variables \citep{wainwright2019high}, it follows that for some constants $c_1$ and $c_2$:
\[
\mathbb{P} \left[ (\|\theta_j - \theta_k^*\|^2 + \sigma^2) \chi^2(n_km_k) \\ <  (\|\theta_k - \theta_k^*\|^2 + \sigma^2) \chi^2(n_km_k) \right] \\ \le c_1 e^{-c_2n_km_k (\frac{\rho}{\rho + 1})^2}
\]
Hence:
\[
\mathbb{P} \left[ \sum_{ i \in Q_k} \mathcal{L}_{s(i)}^i\left(\theta_j; H_{s(i)}^i\right) \le  \sum_{ i \in Q_k} \mathcal{L}_{s(i)}^i\left(\theta_k; H_{s(i)}^i\right) \right] \\ \le c_1 e^{-c_2n_km_k \left(\frac{\rho}{\rho + 1}\right)^2}
\]
Thus, by union bound it follows:
\[
\mathbb{P} \left[ \exists~{k \ne j}~\text{s.t.}~\sum_{ i \in Q_k} \mathcal{L}_{s(i)}^i\left(\theta_j; H_{s(i)}^i\right) \le \sum_{ i \in Q_k} \mathcal{L}_{s(i)}^i\left(\theta_k; H_{s(i)}^i\right)\right] \\ \le K^2 c_1 e^{-c_2n_km_k \left(\frac{\rho}{\rho + 1}\right)^2}
\]
Thus:
\begin{align*}
\mathbb{P} \left[\forall~{k \ne j}, \sum_{ i \in Q_k} \mathcal{L}_{s(i)}^i \left(\theta_k; H_{s(i)}^i \right) \right.
    &< \left. \sum_{ i \in Q_k} \mathcal{L}_{s(i)}^i\left(\theta_j; H_{s(i)}^i\right)\right] \\ 
    & \ge 1 - K^2 c_1 e^{-c_2n_km_k \left(\frac{\rho}{\rho + 1}\right)^2} \\ 
    & \ge  1 -  c_1 e^{-c_2 \frac{1}{\log K^2} n_km_k \left(\frac{\rho}{\rho + 1}\right)^2} 
\end{align*} 
Thus, with  probability close to $1$, clients in  cluster $Q_k$, achieve lowest loss on model  $\theta_k$,   for all $k$.

\end{proof}

\subsection{Comparison with \IFCA}

We  demonstrate that the cluster recovery condition of \IFCA implies loss vector separation for \clove. We assume  a gap of at least $\delta > 0$   between the  loss a client achieves on its best model and its second best model.   The following Lemma shows that vectors of clients belonging to different clusters are at least $\sqrt2 \delta$ apart.\\
\begin{lemma}
\label{ifca:efc}
Let clients $i$ and $j$ achieve lowest loss on models $k_1$ and $k_2$ respectively ($k_1 \ne k_2$). Let there be at least a $\delta$ gap  between the lowest loss and the second lowest loss achieved by a client on any   model.  Then $\|L_i - L_j\| \ge \sqrt2 \delta$, where $L_i$ and $L_j$ are the loss vectors for clients $i$ and $j$.
\end{lemma}

\begin{proof}
Let $l(x,y)$ denote loss of client $x$ on model $y$. Then:
\[
\|L_i - L_j\|^2 \ge \left(l(i, k_1) - l(j, k_1)\right)^2 + \left(l(i, k_2) - l(j, k_2)\right)^2
\]
However, because of $\delta$ loss gap:
\[
l(i, k_2) \ge l(i, k_1) + \delta
\]
and 
\[
l(j, k_2)\le l(j, k_1) - \delta
\]
Thus:
\[
\|L_i - L_j\|^2 \ge (l(i, k_1) - l(j, k_1))^2 + (l(i, k_1) - l(j, k_1) + 2\delta)^2
\]
The right side is minimized when $ (l(i, k_1) - l(j, k_1))^2 = (l(i, k_1) - l(j, k_1) + 2\delta)^2$, or when $ l(i, k_1) - l(j, k_1) = -\delta$. Thus:
\[
\|L_i - L_j\|^2 \ge 2\delta^2
\]
Hence, the result follows.
\end{proof}

Note that in \IFCA the best model for each cluster is initialized to be close to its optimal counterpart. Consequently, $\delta$ can be much larger than $0$. Lemma~\ref{ifca:efc} therefore implies significant distance between the loss vectors of clients of different clusters.

%% file: sections/A2-discussion.tex
\section{Algorithm for selecting the number of clusters}
\label{app:alg-for-selecting-K}

\begin{algorithm}[ht!]
\fontsize{9pt}{9pt}\selectfont
    \caption{Selection of the number of clusters based on silhouette score}
    \label{alg:selection_of_K}

\begin{algorithmic}[1]
    \renewcommand{\algorithmicrequire}{\textbf{Input:}}    
    \REQUIRE Loss matrix $\mathcal{L}$, initial \# of models $K^{(0)}$
    \renewcommand{\algorithmicensure}{\textbf{Output:}}
    \ENSURE Best number of clusters $K_{best}$

    \FUNCTION{\text{\fontfamily{qcr}\selectfont selectBestK}($\mathcal{L}$, $K^{(0)}$)}
        \STATE $best\_score = -1$
        \FOR{$K=2, ..., K^{(0)}$}
            \STATE $Z_K$ = Clustering of the loss vectors (rows of $\mathcal{L})$ using a clustering algorithm, e.g. Agglomerative Clustering, into $K$ clusters
            \STATE $score = \text{\fontfamily{qcr}\selectfont silhouetteScore}(\mathcal{L}, Z_K)$
            \IF{$score > best\_score$}
                \STATE $best\_score = score$
                \STATE $K_{best} = K$
            \ENDIF
        \ENDFOR
        \STATE return $K_{best}$
    \ENDFUNCTION

\end{algorithmic}
\end{algorithm}

\section{Discussion}
\label{app:discussion}

\subsection{Communication and computation costs}

Compared with other PFL methods like \IFCA that exchange only model weights, \clove additionally sends the loss vectors. 
The size of the loss vectors is $\Theta(K)$ per client. 
This is small compared to the model weights themselves, which are $\Theta(K*R)$ per client (same as in \IFCA), where $R$ is the number of model parameters (or gradients). 
Also, in practice, $K$ is usually small.
So the overhead of communicating the loss vectors is not significant.

Furthermore, thanks to \clove's early stopping feature, all models need only to be sent to all clients for the rounds until stability is reached.
As shown in our experiments (see Appendix~\ref{app:early-stopping}), stability is achieved in a few rounds with high probability.
After that, only a single model needs to be sent to each client.
From then on, each cluster performs regular \fedavg, so the cost is similar to \fedavg, which is the best possible (other baselines also have a computation and communication cost at least as high as that of \fedavg).

Another technique for further reducing the size of the communicated models in the first few rounds is to leverage weight-sharing techniques from multi-task learning, as suggested also in \IFCA: the $K$ models can share a few initial layers that are meant to learn the common properties of the data, and the rest of the layers are meant to learn the distinct features of each cluster. 
This way, when the server sends the models to each client, it needs to only send the common part once alongside the distinct parts, thus achieving savings in communication costs.

Finally, one might argue that \clove might result in high peak memory usage, due to clients evaluating all the models before cluster stability is reached. 
This can be alleviated as follows: the $K$ models do not need to be evaluated by the client all at the same time. 
Instead, if memory is a constraint, the client can be requesting and evaluating each of the $K$ models sequentially one at a time, storing in memory only the loss produced by each model. 
This way, it can form the loss vector while only holding one model in memory at a time.

\subsection{Privacy}

Although privacy is not the focus of this paper, it is an important concern in FL and we now briefly discuss it.
Compared to standard FL algorithms such as \fedavg that share model weights/gradients, clients in \clove additionally share the average model losses. 
Importantly, these loss values are not the raw model outputs; they are computed based on the output and the (unknown to the server) input or labels, and also averaged for all datapoints of the client before being sent to the server. 
While this aggregate measure is generally not considered privacy-sensitive, its privacy implications remain underexplored. 
Further research is needed to assess whether, and under what conditions, such aggregate statistics could lead to unintended information leakage, not with regard to \clove in particular, but more generally. 
Moreover, several baselines (e.g. \cite{tan_fedproto_2022, xu_personalized_2023_FedPAC, vahidian_efficient_2023_PACFL}) similarly share 1-D vectors or summaries of client data (we only send losses, not summaries of data). 
This process is deemed irreversible by those baselines, and they even mention that further privacy-preserving techniques can be employed.

One may raise the concern that a malicious client could potentially try to take advantage of the shared model weights to attempt launching an attack with privacy implications. 
However, note that in \clove, just like in standard \fedavg-based learning, the models that the clients receive from the server are not models directly trained on individual clients' local data, but aggregated versions over the clients of each cluster. Because of this, even if a malicious client could launch e.g. a membership inference attack (to see if some data point has been used by a specific client during training), it could only infer that the data point is in the training set of some client of an entire cluster, but not learn the identity of the specific clients that used this data point.

Additionally, \clove could be enhanced with Differential Privacy (DP) to further strengthen the privacy guarantees. 
Of course, in that case, some privacy-vs-performance tradeoff would emerge. 
There are two aspects to this: one is the effect of the DP noise on the loss vectors, which could affect the clustering performance, and the other is the accuracy of the models themselves (which is not \clove-specific, but is also the case in regular DP-enhanced Federated Learning). 
By keeping the noise within some tolerance, these effects can be controlled to not hurt the overall performance.

\subsection{Comparison with FedGWC paper}

\cite{FedGWC} is a hierarchical approach like \cite{Sattler_2021_CFL}, and as such the number of rounds for clustering can be very large, as is also evident from the analytical guarantees that are only in the asymptotic regime as the number of rounds grows without bound; in contrast, \clove achieves correct clustering in 2-3 rounds, as shown both analytically and empirically. 
Moreover, in their approach, amount of state kept at server is quadratic in number of clients. 
With our approach, the amount of state at the server is not a function of the number of clients, but rather it is linear on the number of clusters, which is typically much smaller than number of clients. 
Additionally, FedGWC's clustering happens on the basis of losses computed on a single model. 
Specifically, the vectors used in Eq. 4 for computing entries of matrix $W$ are based on the losses (and rewards) of clients on the single model that is sent to the clients in that round. 
On the other hand, \clove performs clustering by comparing clients' losses across multiple models. 
This can be more effective, as losses of multiple models amplify the difference in data distributions among clients belonging to different clusters. 
This is why \clove is able to recover clusters so much faster.

%% file: sections/A3-hyperparameters-and-datasets.tex
\section{Details on datasets, models and compute used}
\label{app:hyperparameters-and-datasets}

\paragraph{Datasets} We use the MNIST dataset \citep{lecun_MNIST_1998}, containing handwritten digit images from 0 to 9 (10 classes; 60K points training, 10K testing), 
the CIFAR10 dataset \citep{CIFAR10_paper}, containing color images from 10 categories (airplane, bird, etc.) (50K points training, 10K testing),
and the Fashion-MNIST (FMNIST) image dataset \citep{FMNIST_paper}, containing images of  clothing items from 10 categories (T-shirt, dress, etc.) (60K points training, 10K testing).
Beyond image datasets, we also use two text datasets: 
Amazon Review Data \citep{AmazonReview_paper} (denoted by Amazon Review), containing Amazon reviews from 1996-2018 for products from 4 categories (2834 for books, 1199 for DVDs, 1883 for electronics, and 1755 for kitchen and houseware) and the consumer sentiment for each product based on the reviews (binary -- positive/negative), and
AG News \citep{AG_news_link, AG_news_paper}, containing news articles (30000 training and 1900 testing) from 4 classes (world, sports, business, and science).
Finally, we use the Federated EMNIST (FEMNIST) dataset \citep{caldas_leaf_2019}, an FL dataset with data points being handwritten letters or digits from a particular human writer.
All these datasets are also used by the baselines.

\paragraph{Models} For MNIST and FMNIST unsupervised image reconstruction, we used a fully connected autoencoder with Sigmoid output activation. 
For CIFAR-10 unsupervised image reconstruction, we used a Convolutional Autoencoder with a fully connected latent bottleneck and a Tanh output activation. 
For supervised MNIST, FMNIST and FEMNIST image classification we used a simple convolutional neural network (CNN) classifier designed for grayscale images. 
Each convolutional layer uses: 5×5 kernels, followed by ReLU and 2×2 max pooling with stride 2. Its output is a logit vector. 
For CIFAR-10 image classification, we used a deep convolutional neural network classifier, designed for 32×32 RGB images. 
It uses batch normalization for stable training and dropout in fully connected layers to reduce overfitting. 
For AG News we used a TextCNN with an embedding layer, and for Amazon News we used a simple multi-layer perceptron (MLP) model.

Note that we are the first to propose an algorithm that can cluster correctly \textit{using autoencoders in a fully unsupervised setting}. The datasets we use can all be clustered successfully using simple models, thanks to \clove's superior ability to separate the clients, so there is no need to employ heavier models for these tasks. (If a task requires a certain type of model, \clove should again work well, as it is only using the loss that the model outputs -- the model itself is used as a black box).

\paragraph{Experiment setup}

We provide the details on the number of clients and datapoints used for each of our experiments in Table \ref{tab:experiment_details}.

\input{tables/experiment_details}

\paragraph{GitHub links}

The GitHub links for the code for each of the baseline algorithms used in the paper are shown in Table \ref{tab:github_links}.

\input{tables/github_links}

Our own code for the self-implemented algorithms can be found at \url{https://github.com/Nokia-Bell-Labs/Loss-Vector-based-Clustered-Federated-Learning}.

\paragraph{Compute used}
\label{app:compute}

We train the neural networks and run all our evaluations on a server with 98 cores with Intel(R) Xeon(R) Gold 6248R CPU @ 3.00GHz, and 256GB RAM. 
For our experiments, we used Python 3.12 and PyTorch \citep{pytorch_2019}, version 2.7.0.

%% file: tables/experiment_details.tex
\begin{table}[ht]
\centering
\caption{Details on number of clients and datapoints (S: Supervised, U: Unsupervised)}
\label{tab:experiment_details}
\fontsize{9pt}{9pt}\selectfont
\begin{tabular}{@{}lccc|ccc@{}}
\toprule
  \multicolumn{1}{c}{\multirow{2}{*}[-0.075cm]{Experiment}} &
  \multicolumn{3}{c|}{Number of clients} &
  \multicolumn{3}{c}{Number of datapoints per client}  \\ 
\cmidrule(l){2-7} 
 &
  \multicolumn{1}{c}{MNIST} &
  \multicolumn{1}{c}{CIFAR-10} &
  \multicolumn{1}{c|}{FMNIST} &
  \multicolumn{1}{c}{MNIST} &
  \multicolumn{1}{c}{CIFAR-10} &
  \multicolumn{1}{c}{FMNIST} \\ 

\midrule
S: Label skew 1 	& 25 	& 15 	& 25 	& 500 	& 500 	& 500 \\
S: Label skew 2 	& 25 	& 15 	& 25 	& 500 	& 500 	& 500 \\
S: Label skew 3 	& 25 	& 25 	& 25 	& 500 	& 500 	& 500 \\
S: Label skew 4 	& 50 	& 30 	& 50 	& 500 	& 500 	& 500 \\
S: Feature skew 	& 40	& 20	& 40    & 500 	& 500 	& 500 \\
S: Concept shift 	& 20 	& 12 	& 20 	& 1000 	& 1000 	& 1000 \\

\midrule

S: Amazon Review	& \multicolumn{3}{c|}{20}   & \multicolumn{3}{c}{1000} \\
S: AG News 		    & \multicolumn{3}{c|}{30}   & \multicolumn{3}{c}{100} \\
S: Ablation (CIFAR-10) 		& \multicolumn{3}{c|}{25}   & \multicolumn{3}{c}{500} \\
U: Initialization (MNIST) 		& \multicolumn{3}{c|}{50}   & \multicolumn{3}{c}{500} \\
U: Linear 		    & \multicolumn{3}{c|}{25}   & \multicolumn{3}{c}{1000} \\
U: Everything else 	& \multicolumn{3}{c|}{50}   & \multicolumn{3}{c}{1000} \\

\bottomrule
\end{tabular}%
\end{table}

%% file: tables/github_links.tex
\begin{table}[ht]
\centering
\caption{GitHub links for the compared algorithms}
\label{tab:github_links}
\fontsize{9pt}{9pt}\selectfont
\begin{tabular}{ll}
\toprule
Algorithm       &       Source code \\
\midrule
Per-FedAvg      &       \url{https://github.com/TsingZ0/PFLlib} \\
FedProto        &       \url{https://github.com/TsingZ0/PFLlib} \\
FedALA          &       \url{https://github.com/TsingZ0/PFLlib} \\
FedPAC          &       \url{https://github.com/TsingZ0/PFLlib} \\
CFL-S           &       \url{https://github.com/felisat/clustered-federated-learning} \\
FeSEM           &       \url{https://github.com/morningD/FlexCFL} \\
FlexCFL         &       \url{https://github.com/morningD/FlexCFL} \\
PACFL           &       \url{https://github.com/MMorafah/PACFL} \\
FedAvg          &       self-implemented  \\
Local-only      &       self-implemented  \\
IFCA            &       self-implemented  \\
CLoVE           &       self-implemented  \\
\bottomrule
\end{tabular}%
\end{table}

%% file: sections/A4-more-experiments.tex
\section{Additional Experimental Results}
\label{app:additional_experimental_results}

\subsection{Additional supervised results}

\subsubsection{Mixed Linear Regression Experiments: Comparison with k-FED}
\label{app:mixed_linear_regression_experiments}

\begin{figure*}[ht]
\begin{center}
    \subfigure[Clustering accuracy of \clove over time (rounds) for different values of $\Delta$]{
        \includegraphics[width=0.4\linewidth]{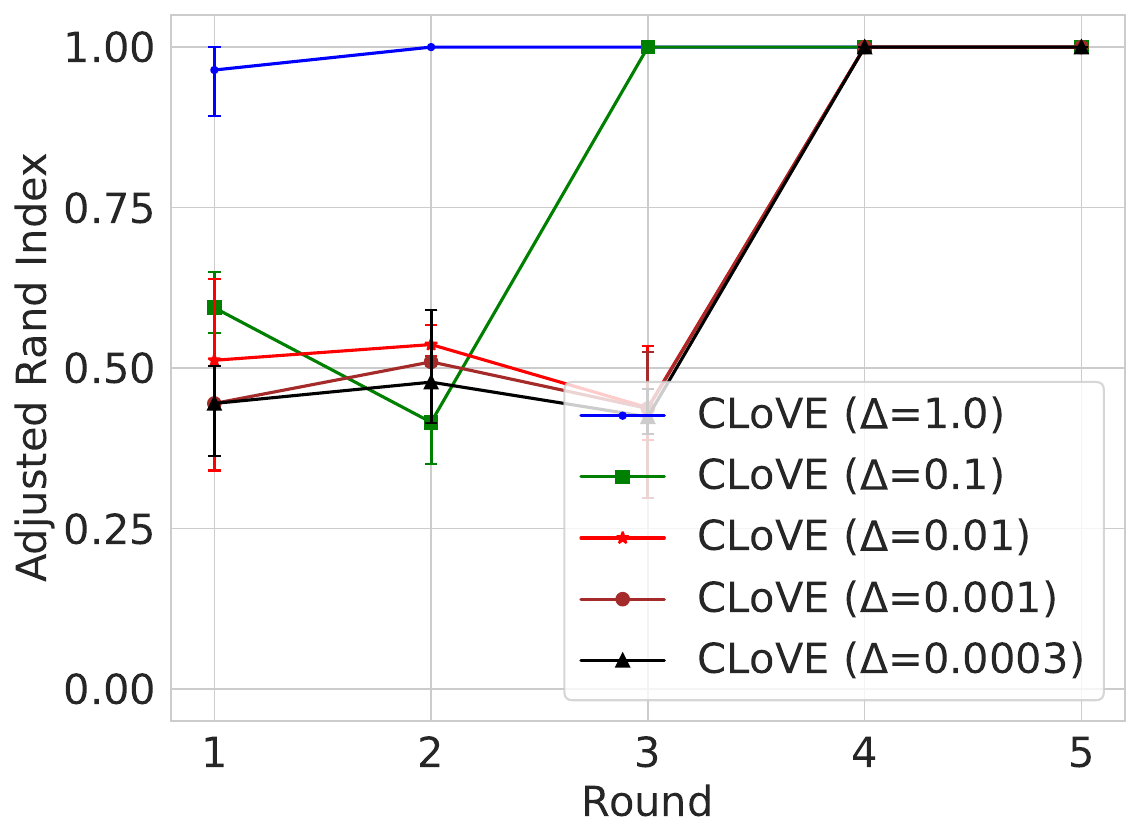}
        \label{fig:clustering_accuracy_LINEAR}
    }
    \hspace{0.5cm}
    \subfigure[Clustering accuracy of \kfed using standard k-means vs with \cite{awasthi2012improved}'s enhancements]{
        \includegraphics[width=0.4\linewidth]{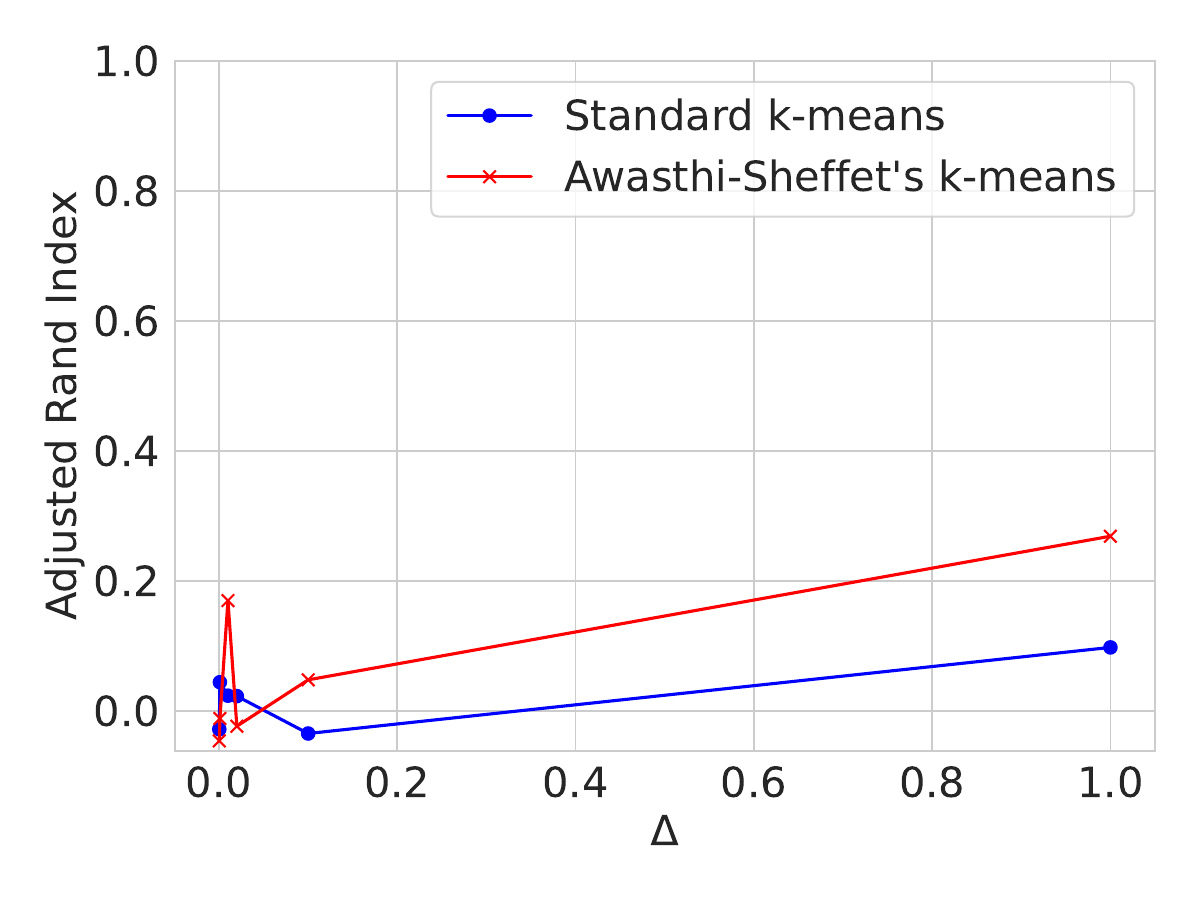}
        \label{fig:kfed}
    }
\end{center}
\caption{Linear regression experiment results}
\end{figure*}

We conduct a study to evaluate  the impact of closeness of the different clusters optimal model parameters on \clove's clustering performance.
We select  a supervised setting, where each cluster's data is generated by a linear model, as described in Section~\ref{sec:theory}. 
We randomly select optimal model parameters $\theta^*_k \in \mathbb{R}^d$ from a unit $d$-dimensional sphere, ensuring that the pairwise distance between them falls within the range $[\Delta, 5\Delta]$. 
Our experimental setup consists of $K=5$ clusters and $M=25$ clients; each client has $n=1000$ data points, and we use $\tau = 25$ local epochs.
Figure~\ref{fig:clustering_accuracy_LINEAR} illustrates the clustering accuracy of \clove over multiple rounds, with $\Delta$ varying from 0.0003 to 1.0.

As shown, \clove consistently recovers the correct cluster assignment. Moreover, as expected, the convergence rate of \clove slows down when the optimal model parameters are closer together (i.e. when $\Delta$ is smaller). In contrast, the performance of \kfed \citep{Dennis_2021} on the same data, even with Awasthi-Sheffet's k-means enhancements~\citep{awasthi2012improved}, is significantly worse (Fig.~\ref{fig:kfed}). 
This discrepancy can be attributed to the fact that the cluster centers (i.e. the mean of their data) are very close to the origin, highlighting the limitations of \kfed in such scenarios. 
Our results demonstrate that \clove outperforms \kfed, especially when the cluster centers are not well-separated.

Additionally, the mixed linear regression setting illustrates the key property behind \clove's design, i.e. that loss vectors are likely to be well-separated even though the underlying raw data may not be.
Here, feature-response pairs $(x_i, y_i)$ of clients in cluster $k$ are related as $y_i = \langle \theta^*_k, x_i \rangle + \epsilon_i$, with features $x_i \sim \mathcal{N}(0, I_d)$ and additive noise $\epsilon_i \sim \mathcal{N}(0, \sigma^2)$, both independently drawn. 
Note that, under this setup, even within a single client, some of the data points can be far apart, while points from different clients may be closer in distance, especially when $\Delta$ (or the pairwise distances between optimal model parameters $\theta^*_k$) is small. 
Hence, the datapoint clusters overlap heavily, and this is the reason methods such as \kfed that cluster data directly achieve lower than 0.3 ARI. 
In contrast, \clove clusters \textbf{client loss vectors} computed on models $[ \theta_1, \theta_2, ... ]$. On any one of these models $\theta$:

\begin{itemize}
    \item 
	For clients in the same true cluster $k$, the sample losses are approximately the same: they concentrate around the population mean loss $||\theta - \theta_k^*||^2$. (This is because loss computation involves averaging over many data points which reduces variance, and the additive noise contributes little to the loss).

    \item
    For clients from different clusters $k_1$ and $k_2$, sample losses can differ substantially, as their population means $||\theta - \theta_{k_1}^*||^2$ and $||\theta - \theta_{k_2}^*||^2$ are likely to be far apart, thus creating separation in their loss vectors. 
\end{itemize}

Thus, despite overlapping data clusters, the loss vectors remain well-separated, allowing \clove to achieve close to 1 ARI.


\subsubsection{Additional label skews}
\label{app:additional-label-skews}

Table~\ref{tab:S_MCF_acc_2} presents image classification test accuracy results for MNIST, FMNIST, CIFAR-10 datasets on additional types of label skews. 
The corresponding ARI accuracy results are presented in Table~\ref{tab:S_MCF_ARI_2} and the clustering convergence results in Table~\ref{tab:S_MCF_T_2}. 
These results include two cases {\it Label skew 3} and {\it Label skew 4}. 
The case {\it Label skew 3} is a data distribution in which there is significant overlap between labels assigned to clients of different clusters.
Specifically, there are 5 clusters with 5 clients each. 
Among the first 4 clusters, all clients  of a cluster get samples from all 10 classes except one. 
The missing class is uniquely selected for each of the 4 clusters. 
The clients of the 5th cluster get samples from all 10 classes. 
The data of a class is distributed amongst the clients that are assigned that class by sampling from a Dirichlet distribution with parameter $\alpha=0.5$, according to the procedure described in Section \ref{sec:performance-evaluation}.
The case {\it Label skew 4} is a data distribution in which $s\%$ of data (we use one third) for clients belonging to a cluster  is uniformly
sampled from all classes, and the remaining $(100 - s)\%$ comes from a  dominant class that is unique  for each cluster~\citep{xu_personalized_2023_FedPAC}. 
Each cluster has 5 clients each of which is assigned 500 samples. 

\input{tables/S_MCF_acc_2}

\input{tables/S_MCF_ARI_2}

\input{tables/S_MCF_T_2}

We observe that in some cases for \textit{Label skew 3} and \textit{Label skew 4}, \fedavg is achieving the highest test accuracy.
This can be attributed to the fact that under these cases, each client holds data from almost every data class, which makes the client distributions very similar. 
Thus, naturally, the single model that \fedavg trains is also a good model for the individual clients.
Moreover, the model of \fedavg is trained on the data of all clients, while the other algorithms only use the data of a single cluster to train each of their models.
This gives \fedavg an advantage, as it trained on much more data for the same number of rounds.

\subsubsection{Additional details for text classification}
\label{app:additional-text-experiments}

The ARI accuracy and speed of convergence results for the Amazon Review and AG News datasets are presented in Table~\ref{tab:S_AA_ARI} and Table~\ref{tab:S_AA_T}, respectively.

\input{tables/S_AA_ARI}

\input{tables/S_AA_T}

\newpage
\subsubsection{The case of unknown number of clusters $K$}
\label{app:unknown-K}

As mentioned in Section \ref{sec:our_algorithms}, \clove does not necessarily require the true number of clusters to be known in advance and given as an input, but it can converge to an appropriate number of clusters during its execution.
To demonstrate this feature of \clove, we perform the following experiment on the FEMNIST dataset.
We use $M=100$ clients and assign to each client all of the data from one of the human writers in FEMNIST. 
We use a CNN and random first-round weight initialization.
In this case, the number of true clusters (which human writers have a similar writing style) is unknown.
\clove searches over different possibilities for the number of clusters/models, achieving an accuracy of $68.2 \pm 1.3$.
This demonstrates the versatility of our algorithm in general settings where the number of true clusters is not given as input.

We also perform experiments with MNIST for all different skews in the unknown $K$ setting. 
Unlike FEMNIST previously where the true $K$ is undefined and hence no ARI computation is possible, in this case the true $K$ is well-defined through the data partitioning and used as ground truth to compute the ARI, but is unknown (not given as an input) to \clove.
The initial number of models $\hat{K}$ is set to 15, and all models are initialized with identical parameters.
The results for the round at which cluster stability is reached and for the ARI are shown in Table~\ref{tab:unknown_K}.

\input{tables/unknown_K}

We see that, in all cases, stability is reached in a small number of rounds (after which that there is no computational overhead from exchanging many models), with the reached value of clusters $K$ being the true one in most cases (ARI = 1).


\subsubsection{Early stopping of clustering}
\label{app:early-stopping}

As mentioned in Section \ref{sec:our_algorithms}, \clove has the ability to stop clustering (and the associated sending of all models by the server to each client and the inference by each client on all the models to generate the loss vectors) early once it stabilizes, so as to save in communication and computation costs.
In this section, we demonstrate the fact that \clove indeed can reach a stable clustering within a few rounds.
We run the same supervised classification experiments on CIFAR-10 for different modes of data mixing as before, but now allowing early stopping when the clustering stabilizes (we define stability as 3 rounds of no change).
In all experiments, \clove achieves an ARI of 1.0.
The results for the test accuracy and the round at which cluster stability is reached are shown in Table \ref{tab:S_C_ES}.
We observe that \clove is able to reach a stable client-to-cluster assignment within 4-7 rounds in all scenarios.

\input{tables/S_C_ES}

\subsubsection{Partial participation}
\label{app:partial-participation}

We now experiment with different levels of client participation (parameter $\rho$).
We use MNIST in a supervised setting with a CNN, $K=5$ clusters, $M=125$ clients (25 per cluster), $n=100$ datapoints/client, and random initialization of the model weights.
The results are shown in Table \ref{tab:S_M_PP}.
Note that for partial participation the ARI at each round is measured using only the clients that participated in that round and comparing their clustering with the ground truth clustering projected on those clients.
We see that \clove very quickly and consistently achieves perfect ARI and almost perfect accuracy even for very small (e.g. $10\%$) client participation rates.

\input{tables/S_M_PP}

We also perform an experiment for the extreme case where participation is so low that the number of participating clients is lower than the true number of clusters.
We examine \clove's behavior with early stopping after cluster stability (Fig. \ref{fig:pp_pathological_1}) and without early stopping (Fig. \ref{fig:pp_pathological_2}).
We use MNIST with $K = 10$ clusters, $m = 2$ clients per cluster and $\rho = 40\%$ participation rate (so only 8 clients participate at each round, which is lower than the true number of clusters, 10).
The ARI is evaluated only for clients that have participated at least once in the training process until that round.
In the first few rounds, the ARI is higher, then it drops, and then again rises to the optimal.
This is because in the beginning very few clients have participated until that point, so there are not many wrong cluster assignments.
As more clients join, initially the number of incorrect assignments increases, but then, as \clove enhances the assignments over time, the ARI increases once again.
As can be seen, after a few tens of rounds, \clove manages to almost reach optimal clustering, despite the very challenging nature of this setting.

\begin{figure}[ht]
\begin{center}
    \subfigure{
        \includegraphics[width=0.4\columnwidth]{figures/pp_pathological_1.pdf}
        \label{fig:pp_pathological_1}
    }
    \subfigure{
        \includegraphics[width=0.4\columnwidth]{figures/pp_pathological_2.pdf}
        \label{fig:pp_pathological_2}
    }
    \caption{ARI for very few participants -- fewer than the number of clusters -- with early stopping (left) and without early stopping (right).}
    \label{fig:heatmap}
\end{center}
\end{figure}

\subsection{Additional unsupervised results}
\label{app:additional-unsupervised-results}

For this section's experiments, we have used a batch size of 64.

\subsubsection{Benefit of personalization}

\begin{figure*}[ht]
    \subfigure{
        \includegraphics[width=0.285\linewidth]{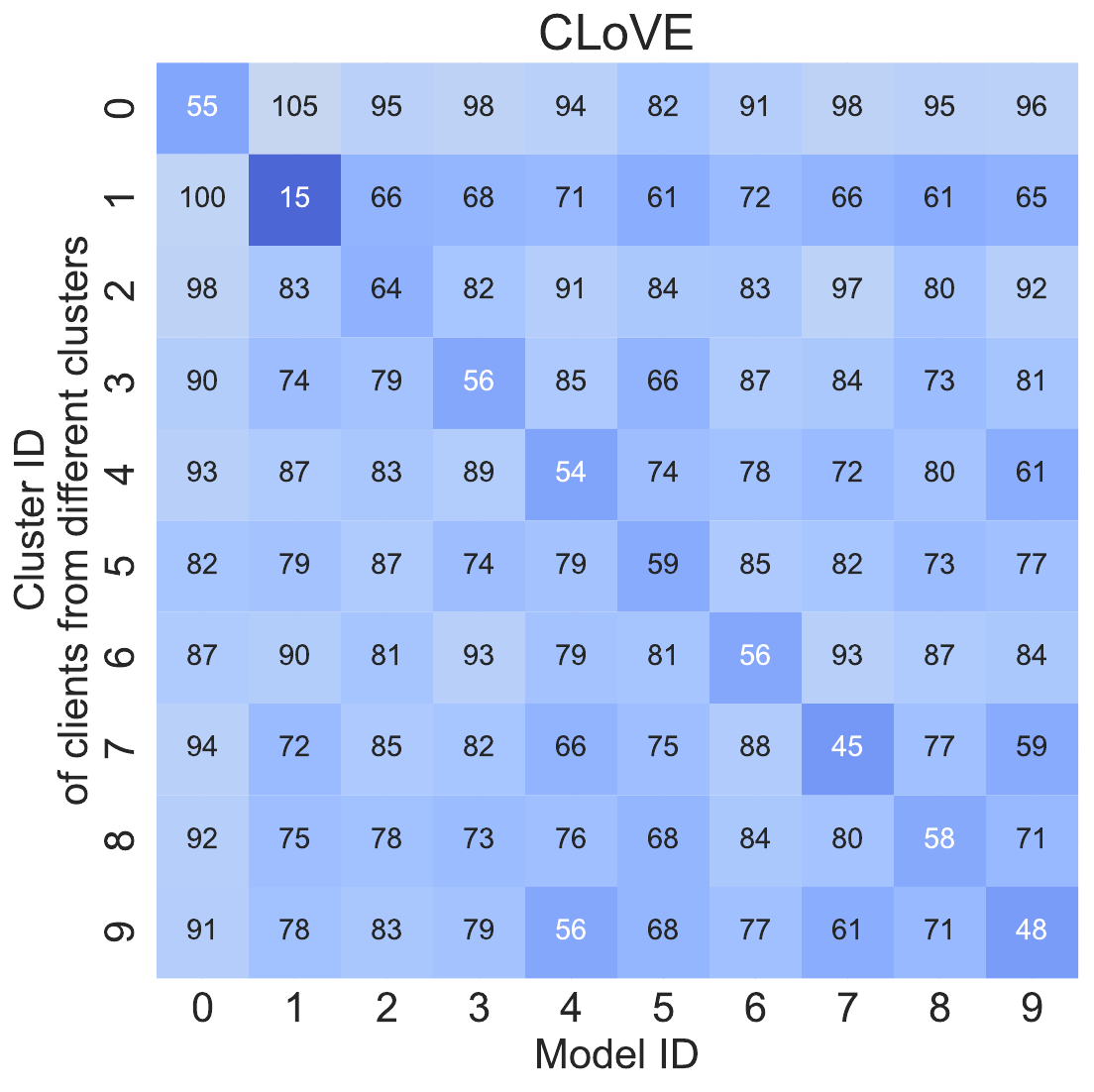}
    }
    \hfill
    \subfigure{
        \includegraphics[width=0.26\linewidth]{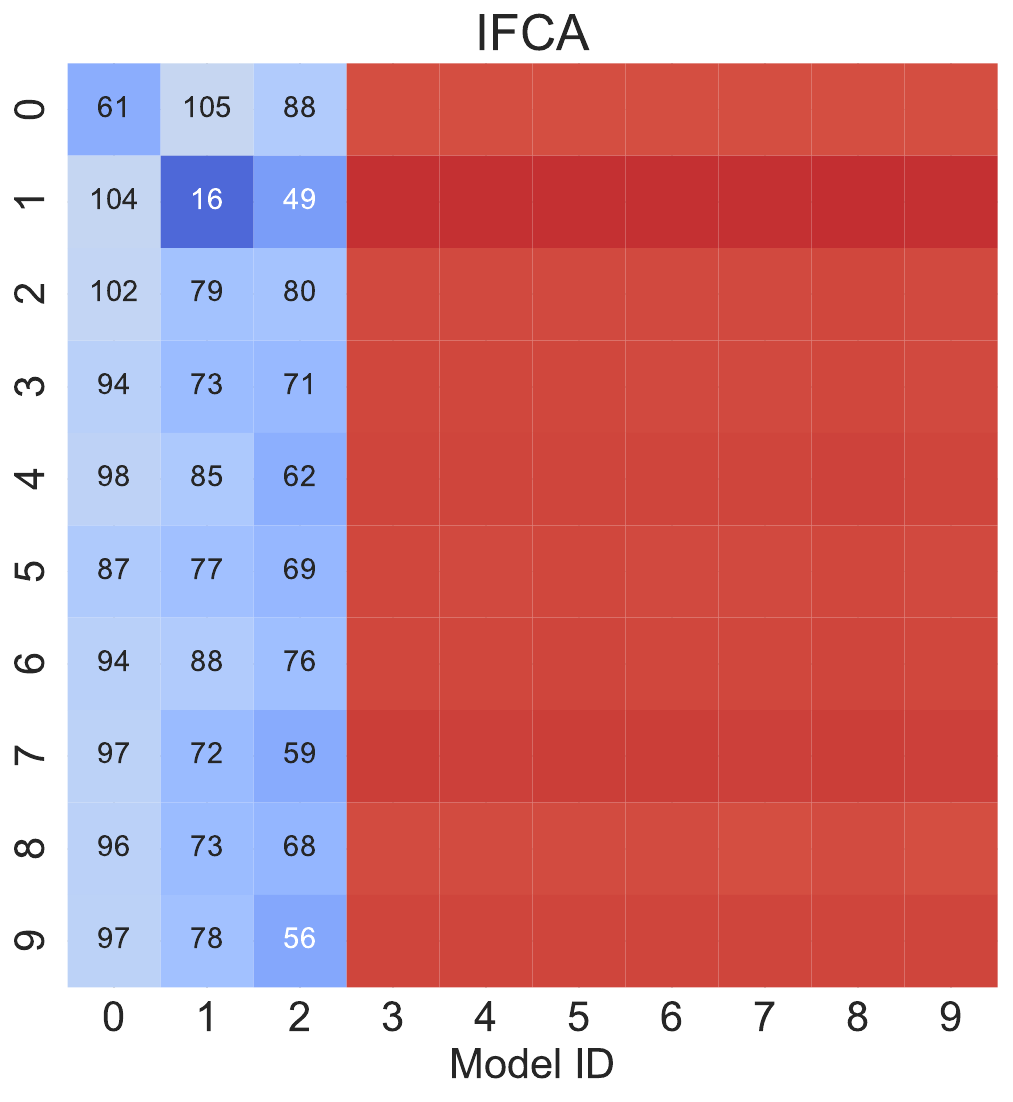}
    }
    \hfill
    \subfigure{
        \includegraphics[width=0.26\linewidth]{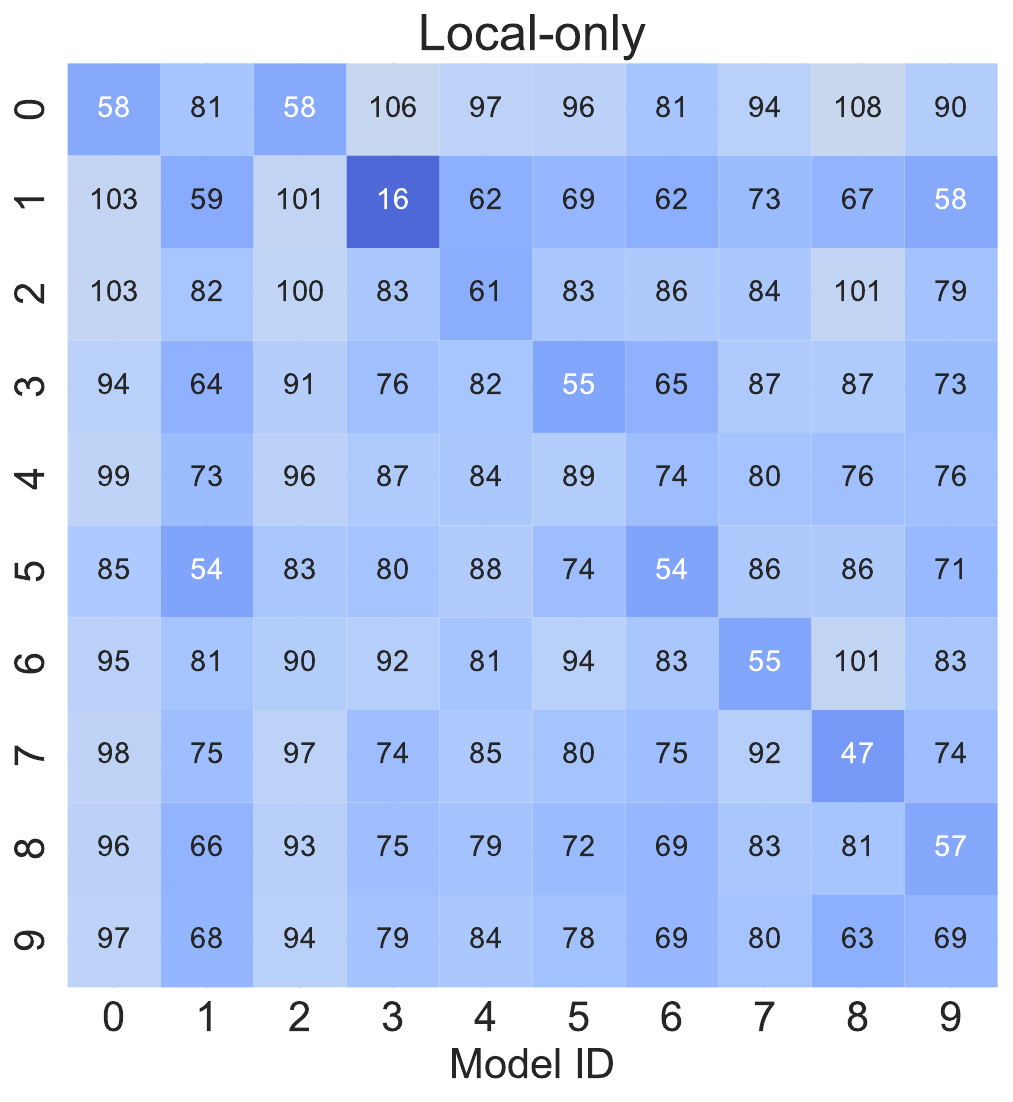}
    }
    \hfill
    \subfigure{
        \includegraphics[width=0.123\linewidth]{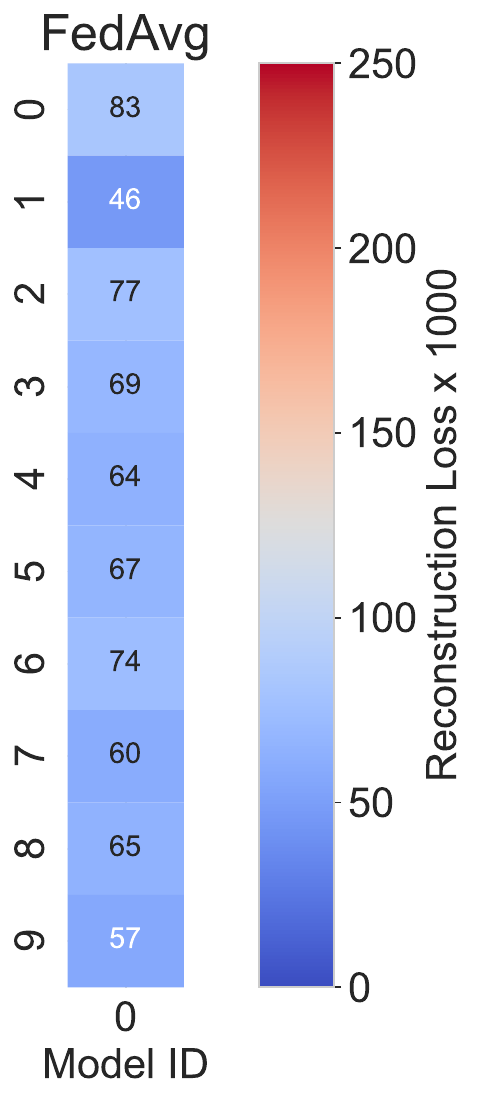}
    }
    \caption{Heatmaps of reconstruction loss of MNIST digits by models clustered and trained by each algorithm}
    \label{fig:loss_heatmaps_MNIST}
\end{figure*}

We now examine the reconstruction loss performance of the trained models and show the benefit of our personalized models over traditional FL models.
For MNIST and $K=10$, $M=50$ and $n=250$, we run 4 different algorithms: \clove, \IFCA, \fedavg, \local.
For this experiment, we consider that each of the 50 clients has a testset of the same data class of the client's trainset, and we feed its testset through all models (10 models for \clove and \IFCA, 1 for \fedavg and 50 for \local), creating the loss matrices shown as heatmaps in Fig. \ref{fig:loss_heatmaps_MNIST} (run for a single seed value).
The vertical axis represents clients, one selected from each true cluster, and the horizontal represents the different models.
Clients and models are labeled so that the diagonal represents the true client to model/cluster assignment.
Numbers on the heatmap express the reconstruction loss (multiplied by a constant for the sake of presentation) for each client testset and model combination.
Very high values are hidden and the corresponding cells are colored red.
For the local case, we only show 10 of the 50 models, the ones that correspond to the selected clients.

We observe that in all cases, the reconstruction loss of each client's assigned model by \clove (the diagonal) is the smallest in the client's row (i.e. the smallest among all models), which confirms that the models \clove assigned to each client are actually trained to reconstruct the client's data (MNIST digit) well, and not reconstruct well the other digits.
On the contrary, \IFCA gets stuck and assigns all clients to 3 clusters (models) -- the first 3 columns.
The rest of the models remain practically untouched (untrained), hence their high (red) loss values on all clients' data.
The 3 chosen models are not good for the reconstruction of MNIST digits,  because they do not achieve the lowest loss on their respective digits.

Comparing \clove and \fedavg, we see that for each row (client from each cluster), the loss achieved by \clove is lower than that achieved by \fedavg.
This demonstrates the \textbf{benefit of our personalized FL approach over traditional FL}: 
\fedavg uses a single model to train on local data of each client, and aggregates updated parameters only for this single model.
\clove, on the other hand, clusters the similar clients together and trains a separate model on the data of clients belonging to each cluster, thus effectively learning -- and tailoring the model to -- the cluster's particular data distribution.
Finally, comparing \clove with \local, we see that the diagonal losses achieved by \clove on the selected models for each client are lower than the corresponding diagonal losses by \local.
This shows the benefit of federated learning, as \clove models are trained on data from multiple clients from the same cluster, unlike \local where clients only train models based on their limited amount of private data.

\subsubsection{Model Averaging versus Gradient Averaging}
\begin{figure}[ht]
\begin{center}
\centerline{\includegraphics[width=0.4\columnwidth]{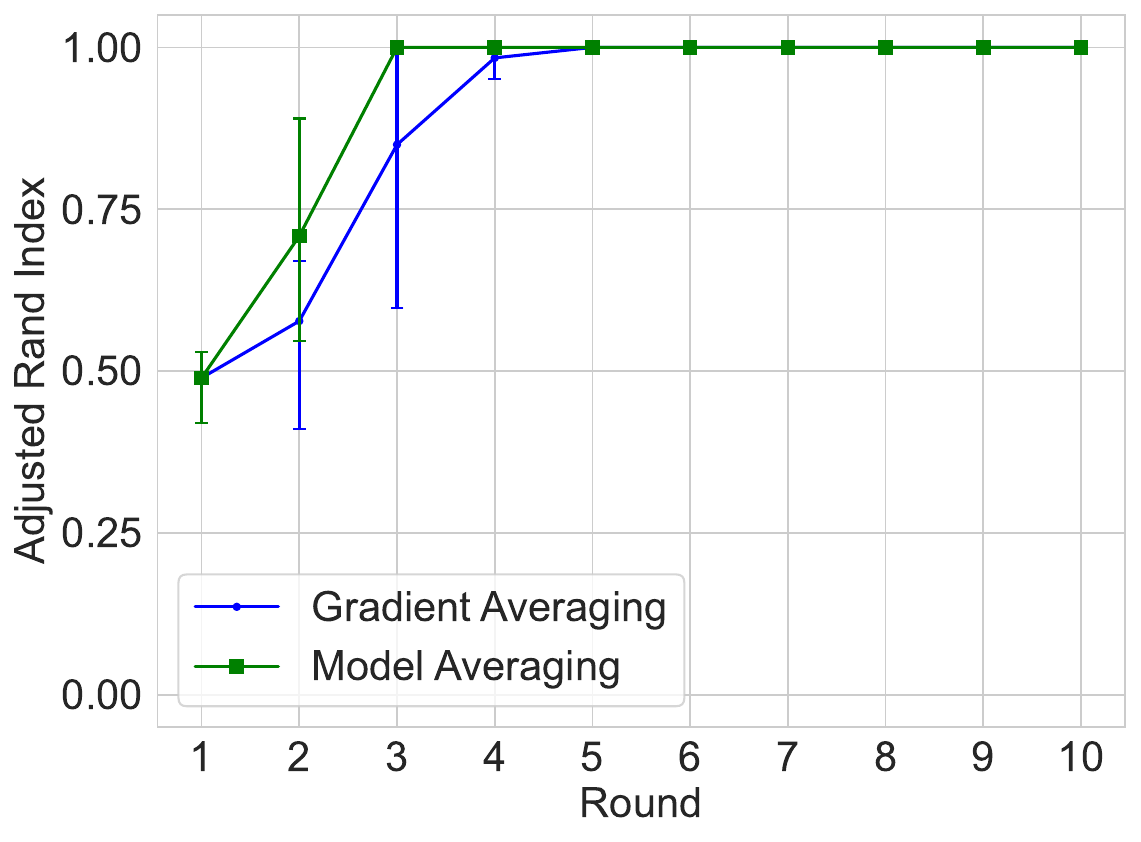}}
    \caption{Clustering accuracy (ARI) of \clove over time (rounds) for MNIST for model averaging vs gradient averaging.}
    \label{fig:MA_vs_GA}
\end{center}
\end{figure}

In this experiment, we compare the two variants of our \clove algorithm, i.e. \textit{model averaging} and \textit{gradient averaging}.
We run an MNIST experiment with $K=10$, $M=50$ and $n=100$, using both averaging methods. 
The results are shown in Fig. \ref{fig:MA_vs_GA}.
We observe that \clove is able to arrive at the correct clustering with both methods, with gradient averaging needing slightly more rounds to do so than model averaging.

\subsection{Scaling with the number of clients}

In this experiment, we examine the effect of increasing the number of clients. 
We use 10 clusters, and 100-200 datapoints per client.
The results for MNIST, CIFAR-10 and FMNIST in the supervised setting (using CNNs) are shown in Table \ref{tab:S_MCF_CS}, and for MNIST in the unsupervised setting (using autoencoders) are shown in Table \ref{tab:U_M_CS}.
In the supervised setting, we see that \clove's final ARI is always perfect, and both the ARI and the test accuracy of \clove consistently exceed those of \IFCA, no matter how big the number of clients is.
Similarly, in the unsupervised setting we observe that \IFCA consistently achieves a bad ARI, while \clove always manages to reach perfect clustering accuracy and scales successfully when the number of clients is large.

\input{tables/S_MCF_CS}

\input{tables/U_M_CS}

\subsection{Convergence dynamics}

\begin{figure}[h!]
\begin{center}
\centerline{\includegraphics[width=0.38\columnwidth]{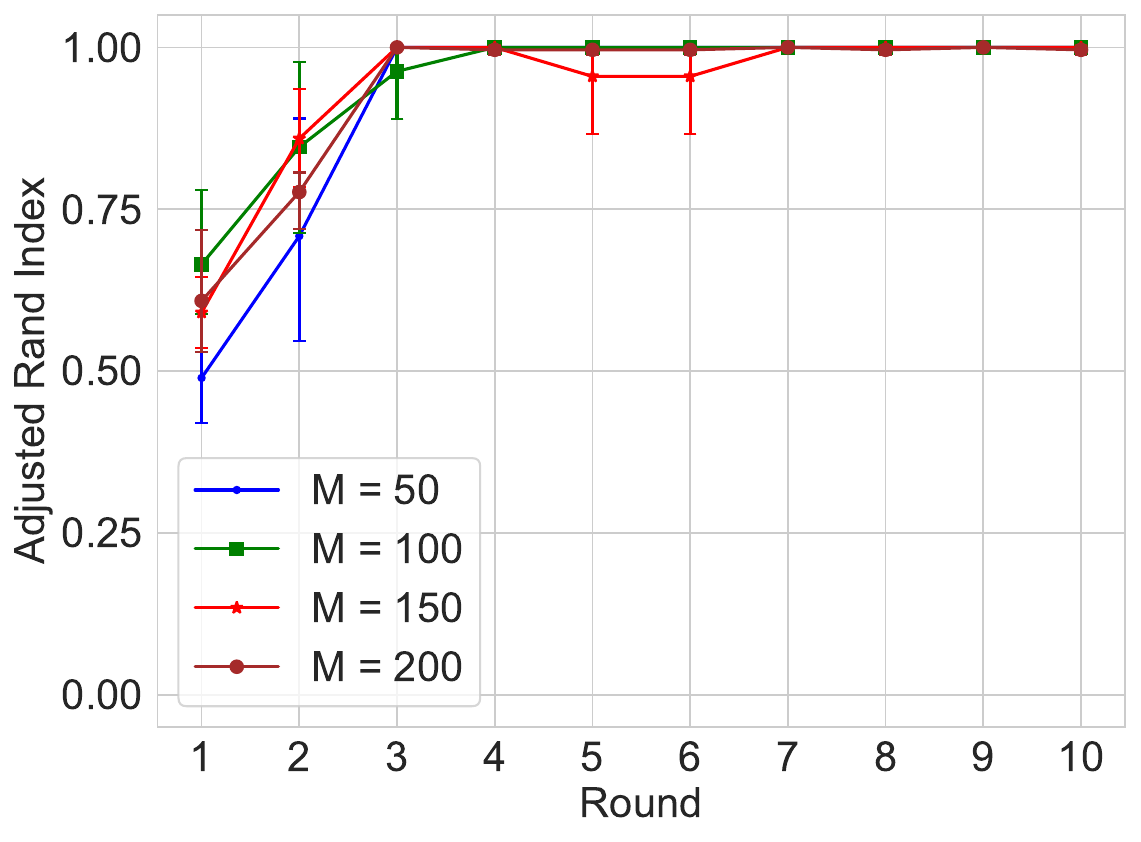}}
\caption{Clustering accuracy (ARI) of \clove over time (rounds) for MNIST for different numbers of clients}
\label{fig:convergence_dynamics}
\end{center}
\end{figure}

We now show the convergence dynamics of \clove for MNIST in the unsupervised setting.
We use $K=10$ clusters, and $n=100$ datapoints per client.
We vary the number of clients from 50 to 200.
The behavior of the ARI can be seen in Figure \ref{fig:convergence_dynamics}.
We observe that \clove reaches a perfect ARI in just a few rounds, for all values of the number of clients $M$.

\subsection{Ablation studies}
\label{app:ablation}

We analyze the impact of changing some of key components of \clove one at a time. 
This includes:

\begin {itemize}
    \item {\bf No Matching}: 
    Order the clusters by their smallest client ID and assign models to them in that order (i.e. cluster $k$ gets assigned model $k$).
    This is in lieu of using the bi-partite matching approach of Alg. \ref{alg:CLoVE}. 

    \item {\bf Agglomerative clustering}: Use a different method than k-means to cluster the loss vectors.
    
    \item {\bf Square root loss}: Cluster vectors of square roots of model losses instead of the losses themselves.
\end{itemize}

Our experiments are for image classification for the CIFAR-10 dataset using 25 clients that are partitioned uniformly among 5 clusters using a label skew with very large overlap. 
The clients of the first four cluster are assigned samples from every class except one class which is chosen to be different for each cluster. 
The clients of the fifth cluster are assigned samples from all classes. 
The partitioning of the data of each class among the clients that are assigned to that class is by sampling from a $\text{Dirichlet}(\alpha) $ distribution with $ \alpha = 0.5 $. 
The test accuracy achieved is shown in Table~\ref{tab:ablation_studies}.
We find that ``No Matching'' has the most impact, as it reduces the test accuracy by almost $4.3\% \pm 1\%$. 
Switching to Agglomerative clustering results in a slight improvement in the mean test accuracy, but the result is not statistically significant. 
Likewise, square root loss results in no significant change in performance. 
This shows that \clove is robust in its design.

\input{tables/ablation_studies}

Additionally, we perform experiments on MNIST, CIFAR-10 and FMNIST evaluating \clove's performance when using FINCH \cite{FINCH_paper} instead of k-means in the loss vector clustering step (line 12 of Alg.~\ref{alg:CLoVE}), using the same setup as before.
We study the most general case where the number of clusters $K$ is not known upfront and is determined by the clustering algorithms (FINCH, k-means + silhouette score as in Alg.~\ref{alg:selection_of_K}) themselves. 
Each experiment is run for 3 values of a randomness seed, and the mean and variance of the results are shown in Table~\ref{tab:FINCH}:

\input{tables/FINCH}

We see that, in all cases, the accuracy reached by k-means is higher than that reached by FINCH. 
Also, in most cases, the ARI reached by k-means is also higher. 
Furthermore, stability (measured by the round at which the clusters become stable) is reached faster under k-means than under FINCH.

%% file: tables/S_MCF_acc_2.tex
\begin{table}[h!]
\centering
\caption{Supervised test accuracy results for MNIST, CIFAR-10 and FMNIST}
\label{tab:S_MCF_acc_2}
\fontsize{9pt}{9pt}\selectfont
\begin{tabular}{@{}cllllll@{}}
\toprule
\begin{tabular}[c]{@{}c@{}}Data\\ mixing\end{tabular} & \multicolumn{1}{c}{Algorithm} & \multicolumn{1}{c}{MNIST} & \multicolumn{1}{c}{CIFAR-10} & \multicolumn{1}{c}{FMNIST} \\ \midrule
\multirow{12}{*}{\rotatebox[origin=c]{90}{Label skew 3}}
	 		& FedAvg 		& $\mathbf{99.0 \pm 0.0}$ 		& $\mathbf{84.1 \pm 0.5}$ 		& $\mathbf{88.8 \pm 0.3}$ \\
	 		& Local-only 		& $85.9 \pm 0.6$ 		& $41.1 \pm 1.5$ 		& $71.1 \pm 0.4$ \\ \cmidrule(l){2-5}
	 		& Per-FedAvg 		& $95.3 \pm 0.3$ 		& $41.6 \pm 0.7$ 		& $74.9 \pm 0.7$ \\
	 		& FedProto 		& $88.8 \pm 0.2$ 		& $39.9 \pm 0.4$ 		& $74.1 \pm 0.2$ \\
	 		& FedALA 		& $94.7 \pm 0.3$ 		& $40.0 \pm 0.7$ 		& $75.1 \pm 0.5$ \\
	 		& FedPAC 		& $95.6 \pm 0.2$ 		& $41.4 \pm 0.6$ 		& $75.5 \pm 0.8$ \\ \cmidrule(l){2-5}
	 		& CFL-S 		& $97.9 \pm 0.1$ 		& $61.2 \pm 0.6$ 		& $84.3 \pm 1.2$ \\
	 		& FeSEM 		& $94.0 \pm 0.2$ 		& $13.3 \pm 0.2$ 		& $72.2 \pm 1.9$ \\
	 		& FlexCFL 		& $92.2 \pm 0.3$ 		& $20.8 \pm 0.5$ 		& $79.7 \pm 0.5$ \\
	 		& PACFL 		& $82.8 \pm 0.9$ 		& $27.1 \pm 3.8$ 		& $68.1 \pm 2.1$ \\
	 		& IFCA 		& $94.2 \pm 0.2$ 		& $57.3 \pm 1.9$ 		& $80.6 \pm 1.0$ \\
	 		& CLoVE 		& $94.2 \pm 0.2$ 		& $56.9 \pm 0.5$ 		& $79.5 \pm 0.5$ \\ \midrule
\multirow{12}{*}{\rotatebox[origin=c]{90}{Label skew 4}}
	 		& FedAvg 		& $\mathbf{98.5 \pm 0.1}$ 		& $60.7 \pm 0.8$ 		& $85.9 \pm 0.2$ \\
	 		& Local-only 		& $94.8 \pm 0.2$ 		& $67.1 \pm 0.5$ 		& $85.5 \pm 0.1$ \\ \cmidrule(l){2-5}
	 		& Per-FedAvg 		& $95.6 \pm 0.1$ 		& $63.7 \pm 0.8$ 		& $83.7 \pm 0.1$ \\
	 		& FedProto 		& $93.3 \pm 0.2$ 		& $45.3 \pm 0.7$ 		& $76.5 \pm 0.3$ \\
	 		& FedALA 		& $96.8 \pm 0.1$ 		& $69.8 \pm 0.3$ 		& $87.3 \pm 0.1$ \\
	 		& FedPAC 		& $96.1 \pm 0.2$ 		& $66.3 \pm 0.5$ 		& $84.8 \pm 0.3$ \\ \cmidrule(l){2-5}
	 		& CFL-S 		& $97.8 \pm 0.1$ 		& $71.3 \pm 1.1$ 		& $88.2 \pm 0.4$ \\
	 		& FeSEM 		& $94.9 \pm 0.3$ 		& $11.9 \pm 1.5$ 		& $80.6 \pm 1.1$ \\
	 		& FlexCFL 		& $97.2 \pm 0.1$ 		& $66.8 \pm 0.0$ 		& $89.5 \pm 0.1$ \\
	 		& PACFL 		& $93.4 \pm 0.2$ 		& $67.4 \pm 0.3$ 		& $84.1 \pm 0.2$ \\
	 		& IFCA 		& $98.0 \pm 0.1$ 		& $64.3 \pm 0.9$ 		& $89.4 \pm 0.2$ \\
	 		& CLoVE 		& $97.8 \pm 0.2$ 		& $\mathbf{72.6 \pm 0.3}$ 		& $\mathbf{89.9 \pm 0.1}$ \\ 
            \bottomrule
\end{tabular}
\end{table}

%% file: tables/S_MCF_ARI_2.tex
\begin{table}[h!]
\centering
\caption{Supervised ARI results for MNIST, CIFAR-10 and FMNIST}
\label{tab:S_MCF_ARI_2}
\fontsize{9pt}{9pt}\selectfont
\begin{tabular}{@{}cllllll@{}}
\toprule
\begin{tabular}[c]{@{}c@{}}Data\\ mixing\end{tabular} & \multicolumn{1}{c}{Algorithm} & \multicolumn{1}{c}{MNIST} & \multicolumn{1}{c}{CIFAR-10} & \multicolumn{1}{c}{FMNIST} \\ \midrule
\multirow{6}{*}{\rotatebox[origin=c]{90}{Label skew 3}}
	 		& CFL-S 		& $0.00 \pm 0.00$ 		& $0.77 \pm 0.18$ 		& $0.00 \pm 0.00$ \\
	 		& FeSEM 		& $0.18 \pm 0.00$ 		& $0.00 \pm 0.00$ 		& $0.00 \pm 0.00$ \\  \cmidrule(l){2-5} 
	 		& FlexCFL 		& $\mathbf{1.00 \pm 0.00}$ 		& $\mathbf{1.00 \pm 0.00}$ 		& $\mathbf{1.00 \pm 0.00}$ \\
	 		& PACFL 		& $0.20 \pm 0.08$ 		& $0.00 \pm 0.00$ 		& $0.15 \pm 0.13$ \\
	 		& IFCA 		& $0.92 \pm 0.12$ 		& $0.69 \pm 0.08$ 		& $0.65 \pm 0.15$ \\
	 		& CLoVE 		& $\mathbf{1.00 \pm 0.00}$ 		& $\mathbf{1.00 \pm 0.00}$ 		& $\mathbf{1.00 \pm 0.00}$ \\ \midrule
\multirow{6}{*}{\rotatebox[origin=c]{90}{Label skew 4}}
	 		& CFL-S 		& $0.00 \pm 0.00$ 		& $0.61 \pm 0.03$ 		& $0.29 \pm 0.02$ \\
	 		& FeSEM 		& $0.12 \pm 0.00$ 		& $0.00 \pm 0.00$ 		& $0.11 \pm 0.02$ \\  \cmidrule(l){2-5} 
	 		& FlexCFL 		& $\mathbf{1.00 \pm 0.00}$ 		& $\mathbf{1.00 \pm 0.00}$ 		& $\mathbf{1.00 \pm 0.00}$ \\
	 		& PACFL 		& $0.53 \pm 0.06$ 		& $0.66 \pm 0.07$ 		& $0.92 \pm 0.02$ \\
	 		& IFCA 		& $0.55 \pm 0.03$ 		& $0.36 \pm 0.22$ 		& $0.50 \pm 0.05$ \\
	 		& CLoVE 		& $0.60 \pm 0.10$ 		& $\mathbf{1.00 \pm 0.00}$ 		& $0.95 \pm 0.07$ \\ 
            \bottomrule
\end{tabular}
\end{table}

%% file: tables/S_MCF_T_2.tex
\begin{table}[h!]
\centering
\caption{Convergence behavior for MNIST, CIFAR-10 and FMNIST}
\label{tab:S_MCF_T_2}
\fontsize{9pt}{9pt}\selectfont
\begin{tabular}{@{}cllll|lll@{}}
\toprule
\multicolumn{1}{l}{\multirow{2}{*}[-0.075cm]{\begin{tabular}[c]{@{}c@{}}Data\\ mixing\end{tabular}}} & 
  \multicolumn{1}{c}{\multirow{2}{*}[-0.075cm]{Algorithm}} &
  \multicolumn{3}{c|}{ARI reached in 10 rounds} &
  \multicolumn{3}{c}{First round when ARI $\geq 0.9$}  \\ 
\cmidrule(l){3-8} 
&  &
  \multicolumn{1}{c}{MNIST} &
  \multicolumn{1}{c}{CIFAR-10} &
  \multicolumn{1}{c|}{FMNIST} &
  \multicolumn{1}{c}{MNIST} &
  \multicolumn{1}{c}{CIFAR-10} &
  \multicolumn{1}{c}{FMNIST} \\ 

\midrule
\multicolumn{1}{c}{\multirow{6}{*}{\rotatebox[origin=c]{90}{Label skew 2}}} &
CFL-S 		& $0.00 \pm 0.00$ 		& $0.00 \pm 0.00$ 		& $0.00 \pm 0.00$
		& --- 		& --- 		& --- \\
\multicolumn{1}{c}{} &
FeSEM 		& $0.18 \pm 0.00$ 		& $0.00 \pm 0.00$ 		& $0.00 \pm 0.00$ 
		& --- 		& --- 		& --- \\
\multicolumn{1}{c}{} &
FlexCFL 		& $\mathbf{1.00 \pm 0.00}$ 		& $\mathbf{1.00 \pm 0.00}$ 		& $\mathbf{1.00 \pm 0.00}$ 
		& $\mathbf{1.0 \pm 0.0}$ 		& $\mathbf{1.0 \pm 0.0}$ 		& $\mathbf{1.0 \pm 0.0}$ \\
\multicolumn{1}{c}{} &
PACFL 		& $0.35 \pm 0.12$ 		& $0.32 \pm 0.11$ 		& $0.55 \pm 0.03$
		& --- 		& --- 		& --- \\
\multicolumn{1}{c}{} &
IFCA 		& $0.83 \pm 0.12$ 		& $0.72 \pm 0.00$ 		& $0.92 \pm 0.12$ 
	    & --- 		& --- 		& --- \\
\multicolumn{1}{c}{} &
CLoVE 		& $\mathbf{1.00 \pm 0.00}$ 		& $\mathbf{1.00 \pm 0.00}$ 		& $\mathbf{1.00 \pm 0.00}$
		& $2.0 \pm 0.0$ 		& $2.0 \pm 0.0$ 		& $2.0 \pm 0.0$ \\

\midrule
\multicolumn{1}{c}{\multirow{6}{*}{\rotatebox[origin=c]{90}{Label skew 3}}} &
CFL-S 		& $0.00 \pm 0.00$ 		& $0.00 \pm 0.00$ 		& $0.00 \pm 0.00$ 
		& --- 		& --- 		& --- \\
\multicolumn{1}{c}{} &
FeSEM 		& $0.18 \pm 0.00$ 		& $0.00 \pm 0.00$ 		& $0.00 \pm 0.00$ 
		& --- 		& --- 		& --- \\
\multicolumn{1}{c}{} &
FlexCFL 		& $\mathbf{1.00 \pm 0.00}$ 		& $\mathbf{1.00 \pm 0.00}$ 		& $\mathbf{1.00 \pm 0.00}$ 
		& $\mathbf{1.0 \pm 0.0}$ 		& $\mathbf{1.0 \pm 0.0}$ 		& $\mathbf{1.0 \pm 0.0}$ \\
\multicolumn{1}{c}{} &
PACFL 		& $0.20 \pm 0.08$ 		& $0.00 \pm 0.00$ 		& $0.15 \pm 0.13$ 
		& --- 		& --- 		& --- \\
\multicolumn{1}{c}{} &
IFCA 		& $0.92 \pm 0.12$ 		& $0.69 \pm 0.08$ 		& $0.65 \pm 0.15$ 
		& --- 		& --- 		& --- \\
\multicolumn{1}{c}{} &
CLoVE 		& $\mathbf{1.00 \pm 0.00}$ 		& $\mathbf{1.00 \pm 0.00}$ 		& $\mathbf{1.00 \pm 0.00}$
		& $2.0 \pm 0.0$ 		& $2.0 \pm 0.0$ 		& $2.0 \pm 0.0$ \\

\midrule
\multicolumn{1}{c}{\multirow{6}{*}{\rotatebox[origin=c]{90}{Label skew 4}}} &
CFL-S 		& $0.00 \pm 0.00$ 		& $0.00 \pm 0.00$ 		& $0.00 \pm 0.00$
		& --- 		& --- 		& --- \\
\multicolumn{1}{c}{} &
FeSEM 		& $0.12 \pm 0.00$ 		& $0.00 \pm 0.00$ 		& $0.11 \pm 0.02$
		& --- 		& --- 		& --- \\
\multicolumn{1}{c}{} &
FlexCFL 		& $\mathbf{1.00 \pm 0.00}$ 		& $\mathbf{1.00 \pm 0.00}$ 		& $\mathbf{1.00 \pm 0.00}$
		& $\mathbf{1.0 \pm 0.0}$ 		& $\mathbf{1.0 \pm 0.0}$ 		& $\mathbf{1.0 \pm 0.0}$ \\
\multicolumn{1}{c}{} &
PACFL 		& $0.53 \pm 0.06$ 		& $0.66 \pm 0.07$ 		& $0.92 \pm 0.02$
		& --- 		& --- 		& --- \\
\multicolumn{1}{c}{} &
IFCA 		& $0.54 \pm 0.03$ 		& $0.70 \pm 0.12$ 		& $0.50 \pm 0.05$
		& --- 		& --- 		& --- \\
\multicolumn{1}{c}{} &
CLoVE 		& $0.90 \pm 0.07$ 		& $\mathbf{1.00 \pm 0.00}$ 		& $\mathbf{1.00 \pm 0.00}$
		& $2.0 \pm 0.0$ 		& $2.0 \pm 0.0$ 		& $2.0 \pm 0.0$ \\

\midrule
\multicolumn{1}{c}{\multirow{6}{*}{\rotatebox[origin=c]{90}{Feature skew}}} &
CFL-S 		& $0.00 \pm 0.00$ 		& $0.00 \pm 0.00$ 		& $0.00 \pm 0.00$
		& --- 		& --- 		& --- \\
\multicolumn{1}{c}{} &
FeSEM 		& $0.00 \pm 0.00$ 		& $0.00 \pm 0.00$ 		& $0.00 \pm 0.00$
		& --- 		& --- 		& --- \\
\multicolumn{1}{c}{} &
FlexCFL 		& $\mathbf{1.00 \pm 0.00}$ 		& $0.11 \pm 0.08$ 		& $\mathbf{1.00 \pm 0.00}$
		& $\mathbf{1.0 \pm 0.0}$ 		& --- 		& $\mathbf{1.0 \pm 0.0}$ \\
\multicolumn{1}{c}{} &
PACFL 		& $0.00 \pm 0.00$ 		& $0.00 \pm 0.00$ 		& $0.68 \pm 0.11$
		& --- 		& --- 		& --- \\
\multicolumn{1}{c}{} &
IFCA 		& $0.67 \pm 0.28$ 		& $0.71 \pm 0.22$ 		& $0.55 \pm 0.10$
	    & --- 		& --- 		& --- \\
\multicolumn{1}{c}{} &
CLoVE 		& $\mathbf{1.00 \pm 0.00}$ 		& $\mathbf{1.00 \pm 0.00}$ 		& $\mathbf{1.00 \pm 0.00}$
		& $2.0 \pm 0.0$ 		& $\mathbf{4.3 \pm 0.9}$ 		& $2.0 \pm 0.0$ \\

\bottomrule
\end{tabular}%
\vspace{-0.3cm}
\end{table}

%% file: tables/S_AA_ARI.tex
\begin{table}[ht]
\centering
\caption{Supervised ARI results for Amazon Review and AG News}
\label{tab:S_AA_ARI}
\fontsize{9pt}{9pt}\selectfont
\begin{tabular}{@{}llllll@{}}
\toprule
\multicolumn{1}{c}{Algorithm} & \multicolumn{1}{c}{Amazon Review} & \multicolumn{1}{c}{AG News} \\ \midrule
	CFL-S 		& $0.07 \pm 0.08$ 	 & 	  $0.39 \pm 0.07$ \\
	FeSEM 		& $0.00 \pm 0.01$ 	 & 	  $0.14 \pm 0.00$ \\
	FlexCFL 	& $0.00 \pm 0.02$ 	 & 	  $0.02 \pm 0.00$ \\
	PACFL 		& $0.00 \pm 0.00$ 	 & 	  $0.00 \pm 0.00$ \\
	IFCA 		& $0.29 \pm 0.20$ 	 & 	  $0.64 \pm 0.12$ \\
	CLoVE 		& $\mathbf{0.73 \pm 0.24}$ 	 & 	  $\mathbf{0.94 \pm 0.09}$ \\
\bottomrule
\end{tabular}
\end{table}

%% file: tables/S_AA_T.tex
\begin{table}[h!]
\centering
\caption{Convergence behavior for the Amazon Review and AG News datasets}
\label{tab:S_AA_T}
\fontsize{9pt}{9pt}\selectfont
\begin{tabular}{@{}lll|ll@{}}
\toprule
  \multicolumn{1}{c}{\multirow{2}{*}[-0.075cm]{Algorithm}} &
  \multicolumn{2}{c|}{ARI reached in 10 rounds} &
  \multicolumn{2}{c}{First round when ARI $\geq 0.9$}  \\ 
\cmidrule(l){2-5} 
&
  \multicolumn{1}{c}{Amazon} &
  \multicolumn{1}{c|}{AG News} &
  \multicolumn{1}{c}{Amazon} &
  \multicolumn{1}{c}{AG News} \\ 

\midrule
CFL-S &     
  $0.00 \pm 0.00$  &
  \hskip 1.4em $0.00 \pm 0.00$ &
  --- &
  \hskip 0.8em --- \\
FeSEM &   
  $0.00 \pm 0.01$ &
  \hskip 1.4em $0.14 \pm 0.00$ &
  --- &
  \hskip 0.8em --- \\
FlexCFL &  
  $0.00 \pm 0.02$  &
  \hskip 1.4em $0.02 \pm 0.00$ &
  --- &
  \hskip 0.8em --- \\
PACFL &  
  $0.00 \pm 0.00$ 	 &
  \hskip 1.4em $0.00 \pm 0.00$ &
  --- &
  \hskip 0.8em --- \\
IFCA & 
  $0.29 \pm 0.20$  &
  \hskip 1.4em $0.64 \pm 0.12$ &
  --- &
  \hskip 0.8em ---\\
CLoVE & 
  $\mathbf{0.72 \pm 0.04}$ &
  \hskip 1.4em $\mathbf{0.94 \pm 0.09}$ &
  --- &
  \hskip 0.8em $\mathbf{2.0 \pm 0.0}$ \\

\bottomrule
\end{tabular}%
\end{table}

%% file: tables/unknown_K.tex
\begin{table}[ht]
\centering
\caption{Stability round and ARI results for MNIST in the unknown $K$ setting}
\label{tab:unknown_K}
\fontsize{9pt}{9pt}\selectfont
\begin{tabular}{@{}lcc@{}}
\toprule
\multicolumn{1}{c}{Data mixing} & \multicolumn{1}{c}{Stability round} & \multicolumn{1}{c}{ARI} \\ \midrule
	None 		        & $4$ 	 & 	  $1.00$ \\
	Label skew 1 		& $6$ 	 & 	  $1.00$ \\
	Label skew 2 	    & $4$ 	 & 	  $1.00$ \\
	Label skew 3 		& $4$ 	 & 	  $1.00$ \\
	Label skew 4 		& $6$ 	 & 	  $0.82$ \\
	Feature skew 		& $4$ 	 & 	  $1.00$ \\
	Concept shift 		& $4$ 	 & 	  $1.00$ \\
\bottomrule
\end{tabular}
\end{table}

%% file: tables/S_C_ES.tex
\begin{table}[ht]
\centering
\caption{Early stopping experiments (CIFAR-10)}
\label{tab:S_C_ES}
\fontsize{9pt}{9pt}\selectfont
\begin{tabular}{@{}ccc@{}}
\toprule
\multicolumn{1}{c}{Data mixing} & \multicolumn{1}{c}{Test accuracy} & \multicolumn{1}{c}{Round of stability} \\ \midrule
	Label skew 1 		& $90.7 \pm 0.2$ 	 & 	  $4.0 \pm 0.0$ \\
	Label skew 2 		& $67.2 \pm 1.5$ 	 & 	  $4.0 \pm 0.0$ \\
	Feature skew 		& $58.2 \pm 1.2$ 	 & 	  $6.3 \pm 1.2$ \\
	Concept shift 		& $59.4 \pm 0.4$ 	 & 	  $4.0 \pm 0.0$ \\
\bottomrule
\end{tabular}
\end{table}

%% file: tables/S_M_PP.tex
\begin{table}[ht]
\centering
\caption{\clove's performance under partial participation}
\label{tab:S_M_PP}
\fontsize{9pt}{9pt}\selectfont
\begin{tabular}{@{}ccccc@{}}
\toprule

\begin{tabular}[c]{@{}c@{}}Participation\\rate $\rho$\end{tabular} & Test accuracy & Final ARI & \begin{tabular}[c]{@{}c@{}}ARI reached\\in 10 rounds\end{tabular} & \begin{tabular}[c]{@{}c@{}}First round\\when ARI $\ge 0.9$\end{tabular} \\ \midrule

	1.0 		& $99.5 \pm 0.1$ 	 & 	  $1.00 \pm 0.00$     &   $1.00 \pm 0.00$      &   $2.0 \pm 0.0$    \\
	0.9 		& $99.6 \pm 0.1$ 	 & 	  $1.00 \pm 0.00$     &   $1.00 \pm 0.00$      &   $2.0 \pm 0.0$    \\
	0.75 		& $99.4 \pm 0.2$ 	 & 	  $1.00 \pm 0.00$     &   $1.00 \pm 0.00$      &   $2.0 \pm 0.0$    \\
	0.5 		& $99.6 \pm 0.1$ 	 & 	  $1.00 \pm 0.00$     &   $1.00 \pm 0.00$      &   $2.0 \pm 0.0$    \\
	0.25 		& $99.4 \pm 0.1$ 	 & 	  $1.00 \pm 0.00$     &   $1.00 \pm 0.00$      &   $2.0 \pm 0.0$    \\
	0.1 		& $99.6 \pm 0.2$ 	 & 	  $1.00 \pm 0.00$     &   $1.00 \pm 0.00$      &   $3.0 \pm 0.8$    \\
    
\bottomrule
\end{tabular}
\end{table}

%% file: tables/S_MCF_CS.tex
\begin{table}[ht]
\centering
\caption{Scaling with the number of clients -- Supervised setting}
\label{tab:S_MCF_CS}
\fontsize{9pt}{9pt}\selectfont
\begin{tabular}{@{}cclll|lll@{}}
\toprule
  \multicolumn{1}{c}{\multirow{2}{*}[-0.075cm]{Algorithm}} &
  \multicolumn{1}{c}{\multirow{2}{*}[-0.075cm]{\# of clients $M$}} &
  \multicolumn{3}{c|}{Final ARI} &
  \multicolumn{3}{c}{Test accuracy}  \\ 
\cmidrule(l){3-8} 
&  &
  \multicolumn{1}{c}{MNIST} &
  \multicolumn{1}{c}{CIFAR-10} &
  \multicolumn{1}{c|}{FMNIST} &
  \multicolumn{1}{c}{MNIST} &
  \multicolumn{1}{c}{CIFAR-10} &
  \multicolumn{1}{c}{FMNIST} \\ 
\midrule
\multirow{5}{*}{IFCA}
        & 50      & $0.92 \pm 0.11$ 		& $0.85 \pm 0.11$ 		& $0.74 \pm 0.22$ 
		& $99.2 \pm 0.8$ 		& $83.7 \pm 5.3$ 		& $96.8 \pm 2.3$ \\
        & 100     & $0.85 \pm 0.11$ 		& $0.65 \pm 0.25$ 		& $0.93 \pm 0.11$ 
		& $99.1 \pm 0.5$ 		& $72.7 \pm 12.7$ 		& $97.4 \pm 2.6$ \\
        & 250     & $0.87 \pm 0.18$ 		& $0.85 \pm 0.10$ 		& $0.85 \pm 0.10$ 
		& $99.4 \pm 0.7$ 		& $80.9 \pm 7.7$ 		& $98.6 \pm 0.5$ \\
        & 500     & $0.93 \pm 0.10$ 		& $0.85 \pm 0.10$ 		& $0.85 \pm 0.10$ 
		& $97.1 \pm 2.9$ 		& $77.3 \pm 5.8$ 		& $89.2 \pm 7.7$ \\
        & 1000    & $0.85 \pm 0.10$ 		& $0.93 \pm 0.10$ 		& $0.75 \pm 0.21$ 
		& $98.3 \pm 1.0$ 		& $78.0 \pm 7.2$ 		& $92.0 \pm 3.6$ \\
\midrule
\multirow{5}{*}{CLoVE}
        & 50      & $1.00 \pm 0.00$ 		& $1.00 \pm 0.00$ 		& $1.00 \pm 0.00$ 
		& $99.8 \pm 0.0$ 		& $90.8 \pm 0.2$ 		& $99.1 \pm 0.0$ \\
        & 100     & $1.00 \pm 0.00$ 		& $1.00 \pm 0.00$ 		& $1.00 \pm 0.00$ 
		& $99.8 \pm 0.0$ 		& $90.3 \pm 0.6$ 		& $99.3 \pm 0.0$ \\
        & 250     & $1.00 \pm 0.00$ 		& $1.00 \pm 0.00$ 		& $1.00 \pm 0.00$ 
		& $99.9 \pm 0.0$ 		& $91.2 \pm 1.5$ 		& $99.3 \pm 0.0$ \\
        & 500     & $1.00 \pm 0.00$ 		& $1.00 \pm 0.00$ 		& $1.00 \pm 0.00$ 
		& $99.6 \pm 0.1$ 		& $82.3 \pm 1.5$ 		& $95.3 \pm 4.4$ \\
        & 1000    & $1.00 \pm 0.00$ 		& $1.00 \pm 0.00$ 		& $1.00 \pm 0.00$ 
		& $99.6 \pm 0.1$ 		& $84.6 \pm 1.2$ 		& $96.5 \pm 3.0$ \\
\bottomrule
\end{tabular}%
\end{table}

%% file: tables/U_M_CS.tex
\begin{table}[ht]
\centering
\caption{Scaling with the number of clients -- Unsupervised setting}
\label{tab:U_M_CS}
\fontsize{9pt}{9pt}\selectfont
\begin{tabular}{@{}cccc@{}}
\toprule
Algorithm & \# of clients $M$ & Test loss & Final ARI  \\ \midrule

    \multirow{4}{*}{IFCA}
                & 100 	 & 	  $0.055 \pm 0.001$     &   $0.091 \pm 0.064$  \\
            	& 200 	 & 	  $0.055 \pm 0.001$     &   $0.079 \pm 0.060$  \\
            	& 500 	 & 	  $0.059 \pm 0.002$     &   $0.210 \pm 0.149$  \\
            	& 1000 	 & 	  $0.067 \pm 0.002$     &   $0.186 \pm 0.137$  \\
            	
    \midrule
    
    \multirow{4}{*}{CLoVE}
                & 100 	 & 	  $0.065 \pm 0.001$     &   $1.000 \pm 0.000$  \\
	            & 200 	 & 	  $0.063 \pm 0.000$     &   $1.000 \pm 0.000$  \\
                & 500 	 & 	  $0.065 \pm 0.000$     &   $1.000 \pm 0.000$  \\
         		& 1000 	 & 	  $0.073 \pm 0.000$     &   $1.000 \pm 0.000$  \\
    
\bottomrule
\end{tabular}
\end{table}

%% file: tables/ablation_studies.tex
\begin{table}[h!]
\centering
\caption{Ablation studies}
\label{tab:ablation_studies}
\fontsize{9pt}{9pt}\selectfont
\begin{tabular}{ll}
\toprule
       & Test accuracy  \\
\midrule
Original CLoVE              & $58.3 \pm 1.4$ \\
No matching                 & $55.8 \pm 0.6$ \\
Agglomerative clustering    & $59.3 \pm 1.5$    \\
Square root loss            & $58.1 \pm 3.3$    \\
\bottomrule
\end{tabular}%
\end{table}

%% file: tables/FINCH.tex
\begin{table}[h!]
\centering
\caption{FINCH vs k-means for MNIST, CIFAR-10, FMNIST}
\label{tab:FINCH}
\fontsize{9pt}{9pt}\selectfont
\begin{tabular}{@{}ccccc@{}}
\toprule
  \multicolumn{1}{c}{Dataset} &
  \multicolumn{1}{c}{Clustering algorithm} &
  \multicolumn{1}{c}{Test accuracy} &
  \multicolumn{1}{c}{ARI} &
  \multicolumn{1}{c}{Stability round} \\ 
\midrule

MNIST   & FINCH     & $92.2 \pm 0.5$    & $0.78 \pm 0.04$   & $8.7 \pm 2.5$ \\
MNIST   & k-means   & $94.4 \pm 1.2$    & $0.86 \pm 0.20$   & $4.7 \pm 0.6$ \\
CIFAR-10    & FINCH & $54.6 \pm 1.0$    & $0.81 \pm 0.08$   & $15.3 \pm 2.3$ \\
CIFAR-10    & k-means & $59.0 \pm 2.0$  & $0.78 \pm 0.31$   & $3.3 \pm 0.6$ \\
FMNIST  & FINCH     & $77.8 \pm 1.4$    & $0.83 \pm 0.13$   & $8.0 \pm 1.0$ \\
FMNIST  & k-means   & $79.3 \pm 0.3$    & $1.00 \pm 0.00$   & $4.7 \pm 1.2$ \\

\bottomrule
\end{tabular}%
\end{table}